\documentclass[runningheads]{llncs}
\usepackage[T1]{fontenc}
\usepackage{graphicx}
\usepackage{booktabs}
\usepackage[misc]{ifsym}
\newcommand{\corr}{(\Letter)}

\usepackage{amsmath,amssymb,amsfonts,bm}
\usepackage{multirow}

\begin{document}
\title{
  Dimensionality-induced information loss \\
  of outliers in deep neural networks} 
\titlerunning{Dimensionality-induced information loss of outliers in DNN}


\author{
  Kazuki~Uematsu \corr \inst{1}\orcidID{0000-0003-4794-7560} \and
  Kosuke~Haruki\inst{1} \and 
  Taiji~Suzuki\inst{2,3}\orcidID{0000-0003-3459-1016} \and
  Mitsuhiro~Kimura\inst{1} \and 
  Takahiro~Takimoto\inst{1} \and 
  Hideyuki~Nakagawa\inst{1} \orcidID{0000-0003-1443-5833}
}
\authorrunning{K. Uematsu et al.}
\institute{
  Corporate Research and Development Center, Toshiba Corporation, Kawasaki 212-8582, Japan
  \email{kazuki1.uematsu@toshiba.co.jp} \and
  The University of Tokyo, Tokyo 113-8656, Japan
  \email{taiji@mist.i.u-tokyo.ac.jp} \and
  Center for Advanced Intelligence Project, RIKEN, Tokyo 103-0027, Japan
}

\maketitle              

\begin{abstract}
  Out-of-distribution (OOD) detection is a critical issue for the stable and reliable operation of systems using a deep neural network (DNN).
  Although many OOD detection methods have been proposed, it remains unclear how the differences between in-distribution (ID) and OOD samples are generated by each processing step inside DNNs.
  We experimentally clarify this issue by investigating the layer dependence of feature representations from multiple perspectives.
  We find that intrinsic low dimensionalization of DNNs is essential for understanding how OOD samples become more distinct from ID samples as features propagate to deeper layers.
  Based on these observations, we provide a simple picture that consistently explains various properties of OOD samples.
  Specifically, low-dimensional weights eliminate most information from OOD samples, resulting in misclassifications due to excessive attention to dataset bias.
  In addition, we demonstrate the utility of dimensionality by proposing a dimensionality-aware OOD detection method based on alignment of features and weights, which consistently achieves high performance for various datasets with lower computational cost.
\end{abstract}

\section{Introduction}\label{sec:intro}

Deep neural networks (DNNs) have received much attention in recent years due to their remarkable versatility and performance.
DNNs are broadly applicable to real tasks including the long-term operation of DNN systems.
However, they suffer from data shifts caused by changes in the surrounding environment.
Shifted data degrade the performance of pretrained models and thus need to be detected as outliers.
This task is commonly referred to as out-of-distribution (OOD) detection, and also goes by other names including novelty detection and open-set recognition, although subtle differences exist in the terminology \cite{OOD-review1,OOD-review2}.
As the application of DNNs continues to expand, OOD detection becomes more critical for ensuring stable and reliable system operation.

Although there have been numerous proposals for precise detection of OOD samples \cite{OOD-review1,OOD-review2},
the mechanism by which OOD samples deviate from in-distribution (ID) samples remains unclear.
Several characteristics of OOD samples have been identified, including
comparatively smaller logit values and corresponding quantities \cite{OOD-baseline,OSR-familiarity},
deviation from the low-dimensional ID subspace \cite{OOD-Mahalanobis,OOD-Mahalanobis-PCA,OOD-feature-subspace,OOD-ViM},
a low-ranked feature vector for each OOD sample \cite{OOD-RankFeat},
and the data complexity-dependent change of layers suitable for OOD detection \cite{OOD-MOOD}.
However, these properties alone do not sufficiently reveal the relationship between observed behaviors and information processing components inside DNNs, 
including layerwise affine transformations using weight matrices and the activation functions.
This kind of a reductionistic perspective is valuable for clarifying the relevant processing that makes ID and OOD samples distinct in non-ideal tasks.
Non-ideal situations are taken to mean the cases where theoretical analysis is not suitable due to the complicated data distributions and network architectures, including that the probability distribution of input data or features is unknown and the model is not well-behaved.
Understanding the properties of OOD samples based on fundamental DNN components could lead to appropriate solutions for addressing the uncertainty inherent in DNNs.

It would also be valuable to clarify the behaviors of OOD samples for practical applications of OOD detection, particularly in cases with various constraints.
Typically, OOD samples are not available before they appear during the operational phase,
and introducing additional models for monitoring the DNN system is impractical because it increases the operating cost.
Furthermore, in some complex and large systems,
modification of the operating model can be challenging because it may impact downstream tasks beyond the current objective.
Additionally, computational time and memory usage can also be significant issues, particularly if real-time processing of a large amount of data is necessary with limited computational resources.
A simplified understanding of OOD samples could help identify the minimum relevant part in DNNs with all constraints satisfied.

In this work, we experimentally investigate how the difference between ID and OOD samples arises by focusing on individual feature transformations inside DNNs for non-ideal tasks.
To address this issue, we explore the layer dependence of feature representations and their relationship with weight matrices.
Our contributions are summarized as follows:
\begin{itemize}
\item Through a systematic analysis of various quantities given in Sec.~\ref{sec:relatedworks},
  we identify that the sharp change in dimensionality plays an essential role in making ID and OOD samples distinct.
\item We verify that this dimensionality change consistently explains not only the observed behaviors presented in Sec.~\ref{sec:results} but also those reported in previous studies as discussed in Sec.~\ref{sec:discussion}.
\end{itemize}
In addition to these, we demonstrate the utility of dimensionality by evaluating a dimensionality-aware OOD detection method in Sec.~\ref{sec:quantitative}.

\section{Problem setting and related work}
\label{sec:relatedworks}
We consider a straightforward and likely representative OOD detection problem where we disregard OOD samples during the training phase.
Let $\mathcal{D}_{\mathrm{ID}}$ and $\mathcal{D}_{\mathrm{OOD}}$ be the sets of ID and OOD samples.
For $\mathcal{D}_{\mathrm{ID}}$, we distinguish the training dataset $\mathcal{D}_{\mathrm{train}}$ and the test dataset $\mathcal{D}_{\mathrm{test}}$.
We examine a scenario where we exclusively utilize $\mathcal{D}_{\mathrm{train}}$ for model training and for hyper parameter tuning.
The training strategy is dedicated to generalization to $\mathcal{D}_{\mathrm{test}}$ without any additional considerations for $\mathcal{D}_{\mathrm{OOD}}$, 
which contrasts with some previous studies \cite{OOD-exposure,OOD-SSL,OOD-CSI}.
This situation enables us to simulate the occurrence of OOD samples and to investigate their intrinsic properties inside DNNs.

In the following subsections, we briefly introduce quantities to be evaluated in Sec.~\ref{sec:results}.
Details of the experimental setup are provided in Appendix~\ref{sec:app_related_detail}.

\subsection{Stable rank of the matrix}
\label{sec:related_srank}
The properties of a matrix, particularly its dimensionality, play an essential role in understanding the behavior of OOD samples.
Refs.~\cite{OOD-Mahalanobis-PCA,OOD-ViM} have reported the importance of the null space of the covariance matrix for OOD detection.
Furthermore, dimensionality of features is closely related to that of weight matrices.
Therefore, we investigate the stable rank $R_F$ of the matrix $A$:
\begin{align}
  R_F &= ||A||_F^2 / ||A||_2^2,
  \label{eq:stablerank}
\end{align}
where $||A||_F$ and $||A||_2$ are the Frobenius norm and the spectrum norm, respectively.
$R_F$ provides a numerical evaluation of the matrix rank robust against small singular values.

To fully investigate the layer dependence of propagation in the DNN,
we perform a similar analysis for the weight matrices $W$ of all layers including convolutional layers.
For convolutional layers, we focus on the local linear transformation represented by the weight matrix $W^{(l)}\in \mathbb{R}^{H^{(l+1)}\times F_x^{(l)} F_y^{(l)}H^{(l)}}$,
where $H^{(l)}$ is the number of channels and $F_x^{(l)}$ and $F_y^{(l)}$ are the kernel sizes at the $l$-th layer.
See Appendix~\ref{sec:app_conv} for more details.
Regarding the covariance matrix $\overline{\Sigma}$, we adopt average pooling of the feature following Ref.~\cite{OOD-Mahalanobis}, as outlined in Sec.~\ref{sec:related_feature}.
These simplifications allow us to diagonalize matrices and compute $R_F$ with practical computational cost.

\subsection{Feature-based detection}
\label{sec:related_feature}
Many studies have utilized feature representations inside DNNs for OOD detection to leverage their rich information \cite{OOD-review1,OOD-review2}.
One straightforward approach involves measuring the Mahalanobis distance from ID training samples \cite{OOD-Mahalanobis}
or considering various extensions \cite{OOD-MahalanobisNCD,OOD-Mahalanobis-Lp,OOD-relativeMahalanobis,OOD-Mahalanobis-PCA,MahalanobisAD,PaDim}.
The Mahalanobis distance $M^{(l)}$ is defined as follows:
\begin{align}
  M^{(l)} &= \min_c \sqrt{(x^{(l)}-\mu_c^{(l)})^{\mathsf T} (\overline{\Sigma}^{(l)})^{-1} (x^{(l)}-\mu_c^{(l)})},  \label{eq:mahalanobis}  \\
  \mu_c^{(l)} &= \frac{1}{N_{\mathrm{train}, c}} \sum_{x_{i,c}\in\mathcal{D}_{\mathrm{train}, c}} x_{i,c}^{(l)},\nonumber \\
  \overline{\Sigma}^{(l)} &= \frac{1}{K} \sum_c \frac{1}{N_{\mathrm{train}, c}}
  \sum_{x_{i,c}\in\mathcal{D}_{\mathrm{train},c}}  (x_{i,c}^{(l)}-\mu_c)(x_{i,c}^{(l)}-\mu_c)^{\mathsf T}. \label{eq:covariance}
\end{align}
Here, $x^{(l)}$ is the feature at the $l$-th layer generated by the input sample $x$,
$x_{i,c}^{(l)}$ is the $l$-th layer feature of the $i$-th training sample $x_{i,c}\in\mathcal{D}_{\mathrm{train}, c}$ with corresponding class label $y_i=c$,
$\mathcal{D}_{\mathrm{train}, c}$ is the training dataset with the class label $c$ out of $K$ classes in the ID dataset,
$N_{\mathrm{train}, c}=|\mathcal{D}_{\mathrm{train}, c}|$ is the number of training data with class label $c$, and
$\overline{\Sigma}^{(l)}$ is the tied covariance of the feature at the $l$-th layer, an approximation of the class-wise covariance \cite{OOD-Mahalanobis}.
By using $M^{(l)}$, we can classify a given sample $x$ as an ID (resp. OOD) sample if the value of $M^{(l)}$ is small (resp. large) for a certain layer $l$.
In the following, we refer to this detection method as ``feature-based detection''.

To reduce the computational cost, Ref.~\cite{OOD-Mahalanobis} employs pixel-averaged features as $x^{(l)}$, $\mu_c^{(l)}$, and $\overline{\Sigma}^{(l)}$.
After averaging the pixel values of the feature map, the Mahalanobis distance is computed for the feature vector with $H^{(l)}$ elements,
where $H^{(l)}$ represents the number of channels at the $l$-th layer.
In this paper, we adopt the pixel-averaged tied covariance for simplicity \cite{OOD-Mahalanobis}.
Furthermore, to avoid division by zero arising from singular covariance, we compute the Moore--Penrose inverse 
$\left(\overline{\Sigma}^{(l)}\right)^{-1} = \sum_k^d \lambda_k^{-1} v_k v_k^{\mathsf T}$,
where $\lambda_k$ denotes the $k$-th largest eigenvalue of $\overline{\Sigma}^{(l)}$ and $v_k$ is the corresponding eigenvector.
$d$ is taken to be the largest value satisfying $\lambda_d/\lambda_0 < \varepsilon$ with $\varepsilon=10^{-6}$.

\subsection{Projection-based detection}
\label{sec:related_projection}
Our main interest is how the deviation of OOD samples from ID samples arises during forward propagation, which consists of a series of linear transformations and activations.
One fundamental factor influencing feature transformation inside DNNs is alignment of weights and features as explored in Ref. \cite{OOD-NuSA}.
Let $W^{(l)}$ be the weight matrix at the $l$-th layer and
$W^{(l)}=L^{(l)}(Q^{(l)})^{\mathsf T}$ be its QR decomposition.
By projecting the feature at the $l$-th layer onto a subspace using the transformation $x_p^{(l)} = (Q^{(l)})^{\mathsf T} x^{(l)}$,
we obtain several OOD detection scores, including $||x_p^{(l)}|| / ||x^{(l)}|| $ and $||x_p^{(l)}||$.
Using these scores, we can classify a given sample $x$ as an OOD sample if the value of scores is large.
The contrapositive statement of the above is that features of ID samples aligns with weights,
which is justified by the noise stability of the trained network \cite{Arora-compression}.
In the following, we refer to this detection method as ``projection-based detection''.

Ref. \cite{OOD-NuSA} considered projections to only fully connected layers using QR decomposition.
This approach restricts access to the singular value associated with the importance of each projection vector in $Q^{(l)}$, as well as the full propagation properties inside the network including convolutional layers.

To overcome these limitations,
we propose a modified projection-based method that aim to identify the relevant subspace for propagation and to investigate the full layer dependence of propagation.
Our approach involves two key modifications to NuSA.
One is removing irrelevant singular vectors, thereby enhancing our awareness of dimensionality of weight matrices.
Another is representing the convolutional layer by the local linear transformation from $F_x^{(l)} F_y^{(l)}H^{(l)}$- to $H^{(l+1)}$-dimensional vector space.
We sometimes refer to the modified method as ``dimensionality-aware'' to clarify that this modification is aware of dimensionality.
See Appendices~\ref{sec:app_compress} and \ref{sec:app_conv} for more details.

\subsection{Similarity of DNN representations}
\label{sec:related_similarity}
The similarity between DNN features serves as a standard metric for investigating properties of DNNs \cite{similarityrevisit,similarity-cnn,similarity-vit,similarity-transfer,similarity-block}.
An extensive study \cite{similarityrevisit} revealed the effectiveness of centered kernel alignment (CKA) defined as follows \cite{CKA-2001,CKA-2012}:
\begin{align}
  \mathrm{CKA}(K_{\mathcal{D}}^{(l_1)}, K_{\mathcal{D}}^{(l_2)})
  &= \frac{\mathrm{tr}((K_{\mathcal{D}}^{(l_1)})^{\mathsf T}K_{\mathcal{D}}^{(l_2)})}{||K_{\mathcal{D}}^{(l_1)}||_F ||K_{\mathcal{D}}^{(l_2)}||_F}.
  \label{eq:cka}
\end{align}
Here, $l_1$ and $l_2$ represent the layer,
$K_{\mathcal{D}}^{(l)} = HK_{0, \mathcal{D}}^{(l)}H$ where $H=I-\frac{1}{N_{\mathcal{D}}}\sum_{i\in\mathcal{D}}(1,\cdots,1) (1,\cdots,1)^{\mathsf T}$ is the centering operator, and
$(K_{0, \mathcal{D}}^{(l)})_{ij} = (x_{i,\mathcal{D}}^{(l)})^{\mathsf T} x_{j,\mathcal{D}}^{(l)}$ where
$x_{i,\mathcal{D}}^{(l)}$ is the feature of the $i$-th sample of the dataset $\mathcal{D}$ at the $l$-th layer.
See Appendix~\ref{sec:app_similarity} for the detailed procedure and a comparison with other similarity scores.
By examining inter-layer CKA, we gain insights into the layerwise feature transformation inside the DNN.
We use 10,000 randomly selected samples to obtain $K_{0, \mathcal{D}}$ for each dataset.

\subsection{Noise sensitivity in the DNN}
\label{sec:related_sensitivity}
The projection to the weight matrix is effective to investigate the property of one-layer linear propagation.
However, we need a more sophisticated approach to characterize multi-layer nonlinear propagation.
In this context, noise sensitivity $\psi$ \cite{Arora-compression} proves useful:
\begin{align}
  \psi_{\eta}(x; M^{(l_1,l_2)})
  &= \frac{||M^{(l_1,l_2)}(x_{\eta}^{(l_1)})-M^{(l_1,l_2)}(x^{(l_1)})||^2}{||M^{(l_1,l_2)}(x^{(l_1)})||^2}.
  \label{eq:noise-sensitivity}
\end{align}
Here, $x^{(l)}$ is defined in the same manner as Eq.~\ref{eq:mahalanobis},
$x_{\eta}^{(l)}=x^{(l)}+\eta||x^{(l)}||$ represents the noise-injected feature, where $\eta$ is typically isotropic noise generated by elementwise Gaussian variables, and
$M^{(l_1,l_2)}(\cdot)$ denotes the function of the DNN forward propagation from the layer $l_1$ to $l_2$, that is, $M^{(l_1,l_2)}(x^{(l_1)}) = x^{(l_2)}$.
Note that in the linear case with the standard Gaussian noise $\eta$, the average of $\psi$ is identical to $||W||_F^2 ||x||^2 / ||Wx||^2$, which is quite similar to the score of projection-based detection in Sec.~\ref{sec:related_projection}.
To simplify matters, we neglect the average over $\eta$, assuming that it will not significantly impact the typical value of $\psi$.
We employ the median of samples in a dataset as the typical value of $\psi$.
We fix the norm of noise $\eta$ to 0.1 \cite{Arora-compression}.

\section{Results}
\label{sec:results}

In this section, we present the experimental results for the various quantities introduced in Sec.~\ref{sec:relatedworks}
to bring the picture in Fig.~\ref{fig:the-picture} into better focus.
Details of the experimental setup are provided in Appendix~\ref{sec:app_related_detail}.
Our findings are corroborated by aggregating results from multiple perspectives, emphasizing consistency rather than relying solely on individual properties.

\subsection{Overview of the experiments and a possible picture}
\label{sec:outline}
\begin{figure*}[t]
\centering
  \begin{tabular}{c}
    \includegraphics[width=0.9\hsize, bb=0.000000 0.000000 960.083207 296.185669]{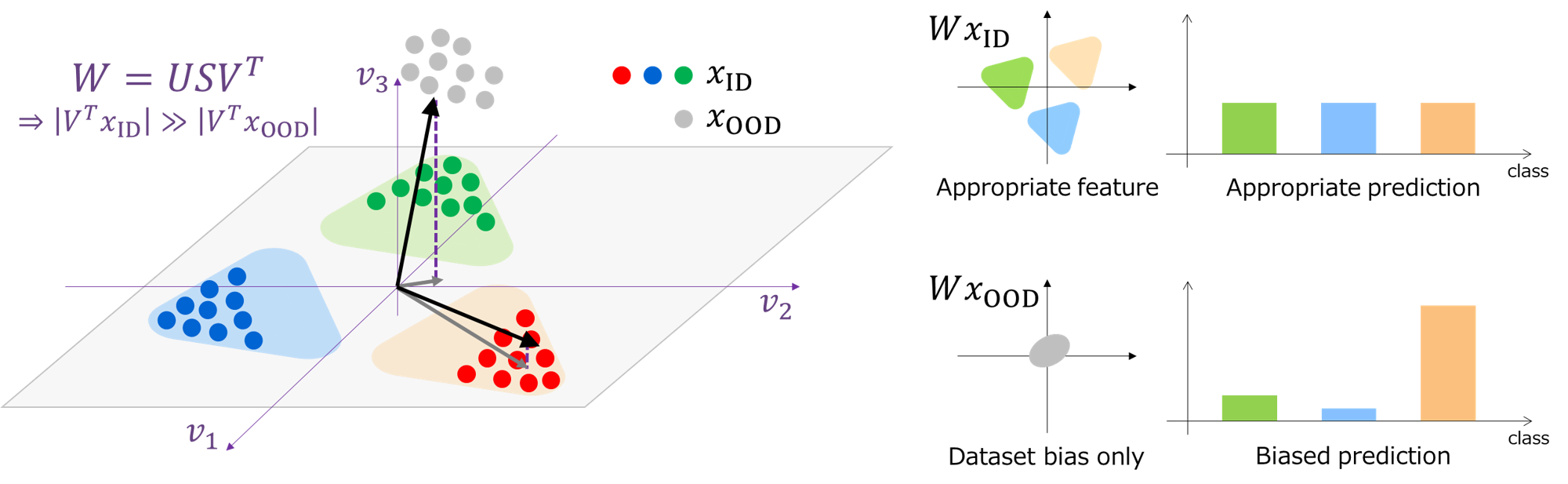}\\
  \end{tabular}
  \caption{
    Qualitative picture showing how OOD samples deviate from ID samples, and how OOD samples are classified.
    Low dimensionalization of weights arises from the significant difference in feature propagations between ID and OOD samples due to their alignment.
    The resulting features of OOD samples are dominated by dataset bias, the common characteristics in the dataset, leading to the biased prediction.
  }
\label{fig:the-picture}
\end{figure*}

The primary objective of our experiments is to establish the picture in Fig.~\ref{fig:the-picture}.
We achieve this by investigating how OOD samples deviate from ID samples as the features propagate to deeper layers.
To accomplish this, we evaluate the layer dependence of various quantities.

The most intriguing observation is the presence of a common layer that exhibits transition-like behaviors across various quantities:
a significant decrease in the stable ranks of covariances and weights,
stabilization of feature-based detection performance,
peaky structure of projection-based detection performance,
boundary of block-like CKA,
and a notable decrease in noise sensitivity.
Furthermore, although the location of this \textit{transition layer} varies depending on network architectures,
\footnote{The transition layer is typically located just after the deepest pooling layer except the global average pooling. 
The exception is ResNet-18 where the transition layer is a little deeper.
This may be due to insufficient low dimensionalization around the corresponding pooling layer.}
similar behaviors are universally observed independent of architectures (Appendix~\ref{sec:app_full-model}) and ID datasets, as long as the number of ID classes is comparably small (Appendix~\ref{sec:app_svhn-mnist}).
These consistent behaviors strongly suggest the existence of a fundamental origin shared across all scenarios.

An essential common element to all of them is low dimensionalization of weight matrices, as observed by the stable rank.
This naturally evokes the following propagation process for ID and OOD samples as illustrated in Fig.~\ref{fig:the-picture}.
For ID samples, low dimensionalization \textit{preserves} most of the characteristic information, including features relevant to classification in the case of classification tasks \cite{Arora-compression}.
However, when it comes to OOD samples, low dimensionalization \textit{eliminates} most of their characteristic information.
The resulting features of OOD samples are thus dominated by the \textit{dataset bias}, except the dataset-independent bias parameters in the model.
Here, the term ``dataset bias'' refers to common characteristics shared within a dataset, which can be characterized by the center of the feature distribution for certain dataset.
For instance, images in the SVHN or MNIST dataset all contain numerical digits.
When we train the model using the CIFAR dataset where digits are not considered crucial,
the digit-related features cannot propagate through the trained weight parameters.
Yet, due to the repeated appearance of digits in the dataset,
the center of features for OOD (SVHN and MNIST) samples deviates from the origin in the feature space.
Consequently, OOD samples exhibit dataset-dependent imbalanced classification.
This excessive attention to dataset bias stemming from low dimensionalization could be a key factor contributing to the overconfident prediction of OOD samples.

These results are closely related to but slightly different from manifold hypothesis \cite{ManifoldHypothesis} and neural collapse \cite{NeuralCollapse}.
The manifold hypothesis states that high-dimensional data lie on a low-dimensional (typically non-linear) manifold.
It is not obvious whether DNNs can transform the non-linear input manifold to the linear output distribution and whether this kind of phenomena is relevant to OOD detection.
Our results devoted to the linear method indicate that the DNN can embed input data into low-dimensional linear space, and that the low-dimensional embedding is essential for OOD detection compared with other effects.
Regarding neural collapse, it states that the weight matrix at the final layer forms a low-dimensional simplex equiangular tight frame.
Our analysis extends this notion beyond the final layer, revealing that several deep layers also exhibit low dimensionality
and that the change of dimensionality is sharp enough to occur at the transition layer.
Our layerwise analysis serves as an experimental validation of these aspects.

Detailed descriptions corresponding to each quantity are given in the following subsections along with specific results.
In Sec.~\ref{sec:stablerank}, we first verify the reduction of dimensionality through stable ranks.
Next, we evaluate the layer dependence of OOD detection performance using the area under receiver-operating characteristic curves (AUROCs) in Sec.~\ref{sec:performance} to demonstrate the relevance of dimensionality to OOD detection.
We further investigate multi-layer properties using CKA and noise sensitivity in Secs.~\ref{sec:cka} and \ref{sec:sensitivity}, respectively, to eliminate the possibility that the multi-layer processing is more important than single-layer dimensionality.
After these characterizations of propagation, we examine resulting classification in Sec.~\ref{sec:inference} to confirm the effect of dataset bias.
Finally, we showcase the utility of dimensionality through the dimensionality-aware projection-based detection method in Sec.~\ref{sec:quantitative}, a modification of NuSA \cite{OOD-NuSA}.
While the main text presents results for VGG-13 for simplicity,
additional verification can be found in Appendix~\ref{sec:app_full-model} for other model architectures, as well as in Appendix~\ref{sec:app_svhn-mnist} for other ID datasets.

\begin{figure}[t]
\centering
  \includegraphics[width=0.4\hsize, bb=0.000000 0.000000 441.000000 311.000000]{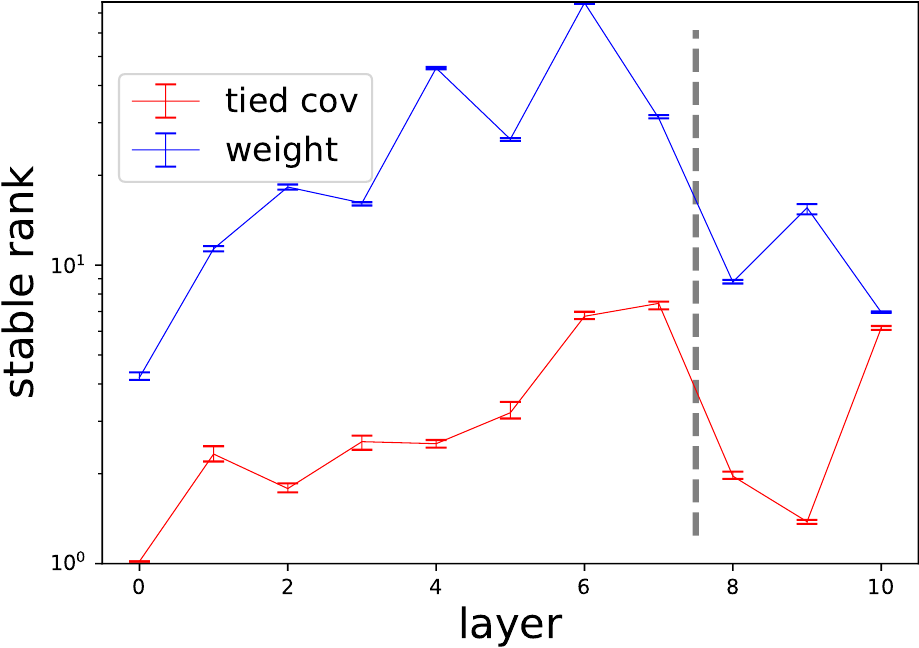}
  \caption{
    Layer dependence of the stable rank of the covariance matrix $\overline{\Sigma}$ and the weight matrix $W$ for the VGG-13 model.
    The dashed line indicates the transition layer.
    Low dimensionalization of features and weights occurs at almost the same layer.
    See Appendices \ref{sec:app_full-model} and \ref{sec:app_svhn-mnist} for further verification.
  }
  \label{fig:stableranks}
\end{figure}

\subsection{Observation of dimensionality via stable ranks}
\label{sec:stablerank}
We first confirm low-dimensional characteristics of weights and features using the stable rank $R_F$ defined by Eq.~\ref{eq:stablerank}.
In Fig.~\ref{fig:stableranks}, we show the layer dependence of $R_F$ of both weight matrices $W$ and covariance matrices $\overline{\Sigma}$.
The dashed line representing the transition layer is manually determined, and is located at the same position hereafter to verify the argument given in Sec.~\ref{sec:outline}.
We can see that $R_F$ exhibits exponential growth in shallower layers corresponding to the increasing number of channels in intermediate features and weights.
However, $R_F$ of both matrices suddenly and simultaneously fall to small values around the transition layer across various architectures and small-class ID datasets (See Appendices~\ref{sec:app_full-model} and \ref{sec:app_svhn-mnist}.).
These behaviors indicate the significance of low dimensionalization in DNNs, 
implying the importance of dimensionality in understanding properties of OOD samples.

\begin{figure}[t]
\centering
\begin{tabular}{ccc}
  \includegraphics[width=0.4\hsize, bb=0.000000 0.000000 453.000000 335.000000]{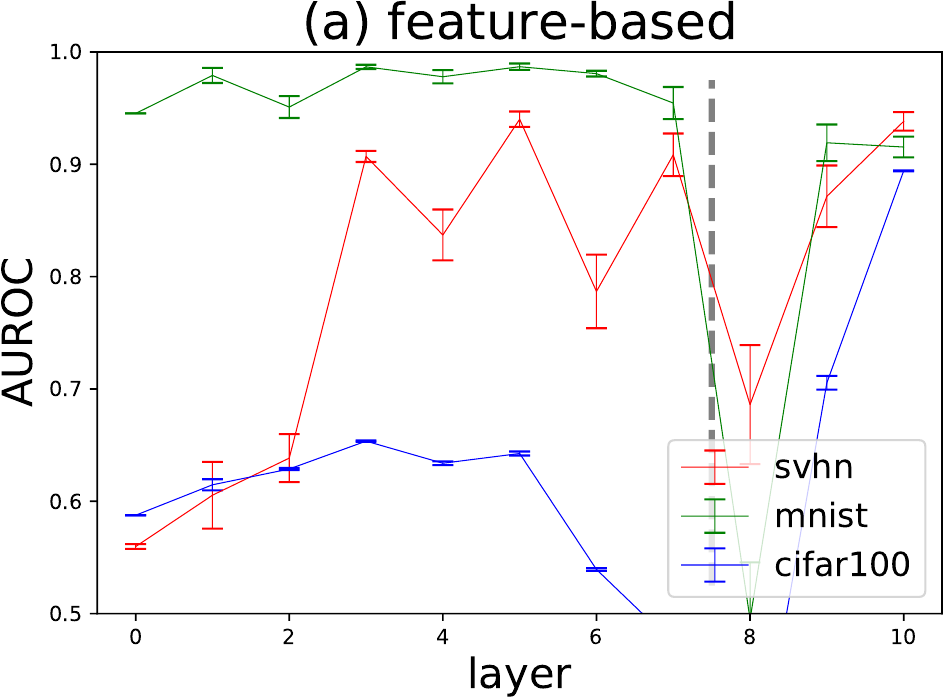} &
  \hspace{10mm} &
  \includegraphics[width=0.4\hsize, bb=0.000000 0.000000 453.000000 335.000000]{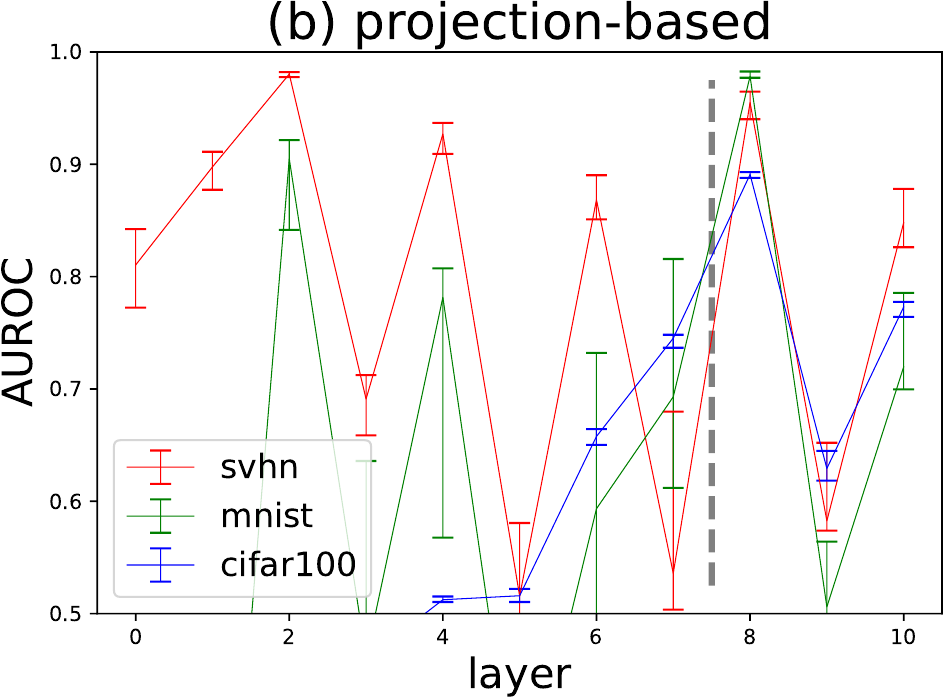}
\end{tabular}
  \caption{
    Layer dependence of the AUROC obtained through (a) Mahalanobis distance $M$ and (b) projected norm $||x_p||$ for the VGG-13 model.
    Different line colors represent the different OOD datasets evaluated.
    The dashed line indicates the transition layer.
    AUROCs are stabilized after transition independent of datasets for feature-based detection,
    while the projection-based discrimination between ID and OOD samples becomes clear just at the transition layer.
    See Appendices~\ref{sec:app_full-model} and \ref{sec:app_svhn-mnist} for further verification.
  }
  \label{fig:auroc}
\end{figure}

\subsection{Transition of OOD detection performance}
\label{sec:performance}
We proceed to explore the relationship between dimensionality and both feature-based and projection-based OOD detection performance using AUROCs.

Fig.~\ref{fig:auroc} (a), we present the layer dependence of AUROCs obtained through the Mahalanobis distance defined by Eq.~\ref{eq:mahalanobis}.
We can see that although the observed AUROCs are highly layer-dependent and dataset-dependent in shallower layers,
they consistently maintain high values independent of the datasets and architectures in deeper layers.
The stabilization of AUROCs is particularly significant for the close-to-ID OOD samples, CIFAR-100 samples represented by the blue curve in this case, almost independent of the network architectures. (See Appendix~\ref{sec:app_full-model}.)
This indicates that, as observed by Ref. \cite{OOD-Mahalanobis-PCA,OOD-ViM}, the features of OOD samples are distributed in the null space of ID covariance matrices within deeper layers.
The correlation between dimensionality and detection performance provides evidence for our picture.

A similar analysis can be conducted for the projection-based method, which quantifies the difference in alignment of ID and OOD samples.
Specifically, we utilize the norm of the projected feature $||x_{p}||$ as the OOD detection score and evaluate the layer dependence of AUROCs for some OOD datasets as depicted in Fig.~\ref{fig:auroc} (b).
Note that our projection-based method modifies NuSA to incorporate singular values and analyze the convolutional layer.
See Appendices~\ref{sec:app_conv} and \ref{sec:app_compress} for more details.
In Fig.~\ref{fig:auroc} (b), we observe a highly layer-dependent and dataset-dependent AUROC.
However, in the vicinity of the transition layer, the AUROC consistently exhibits high values for all datasets and models, albeit not the highest. 
This stable detection performance indicates effective elimination of null-space components of weight matrices due to low dimensionalization of weights.
This is also consistent with the stabilization of feature-based detection in subsequent layers.
We again mention that, although the transition layer varies depending on network architectures,
the same behavior consistently remains independent of network architectures.
(See Appendix~\ref{sec:app_full-model}.)
This universal behavior lends support to the picture outlined in Sec.~\ref{sec:outline}.

\begin{figure}[t]
\centering
\begin{tabular}{ccc}
  \includegraphics[width=0.4\hsize, bb=0.000000 0.000000 398.000000 335.000000]{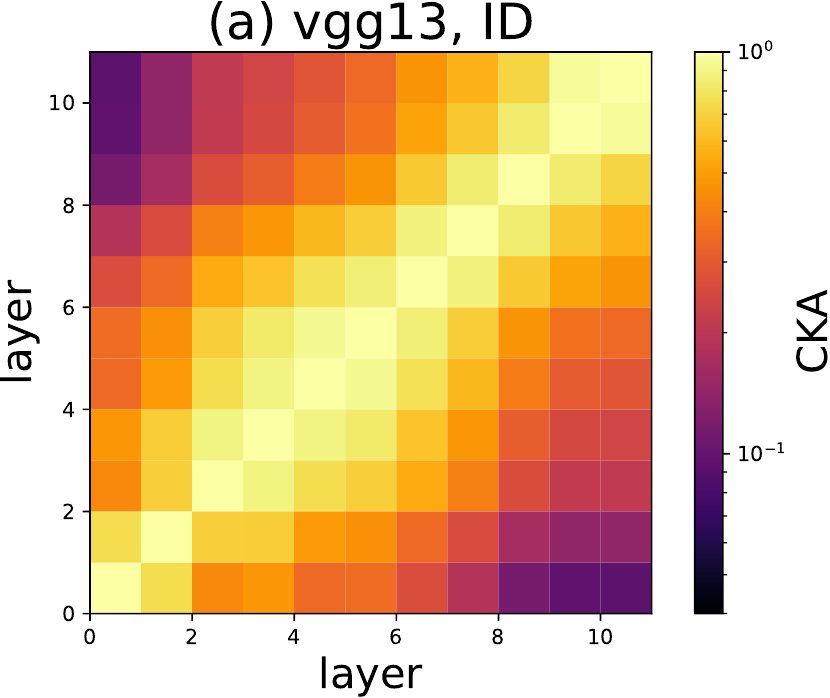} &
  \hspace{10mm} &
  \includegraphics[width=0.4\hsize, bb=0.000000 0.000000 398.000000 335.000000]{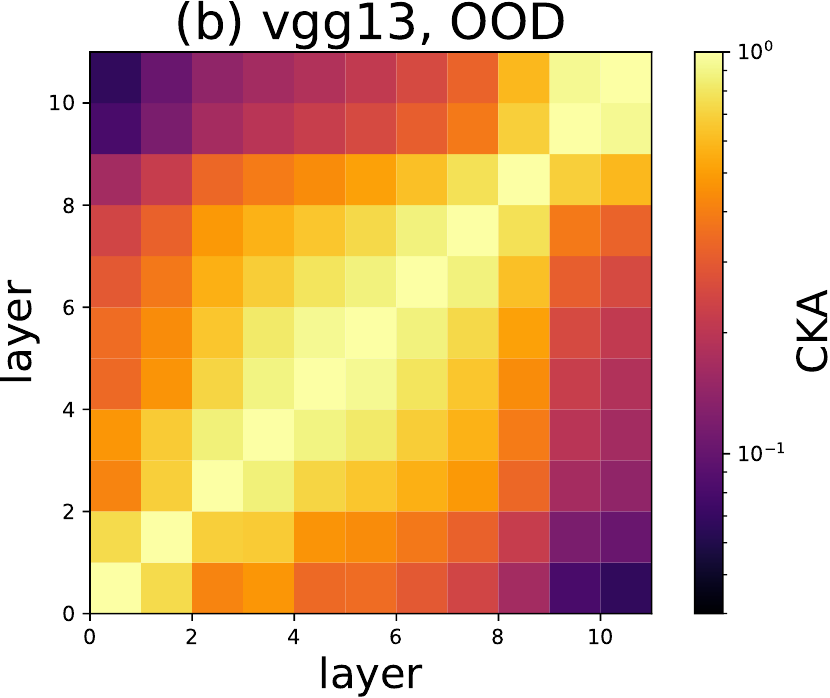}
\end{tabular}
  \caption{
    CKA of features in various layers
    for the VGG-13 model.
    (a) CKA of ID (CIFAR-10) samples.
    (b) CKA of OOD (CIFAR-100) samples.
    In both pannels, the horizontal and vertical axes represent layers, and the color bar represents CKA.
    Block-like saturations appear both for ID and OOD samples around the transition layer.
    See Appendices~\ref{sec:app_full-model} and \ref{sec:app_svhn-mnist} for further verification.
  }
  \label{fig:cka-layers}
\end{figure}

\subsection{Block structure of CKA}
\label{sec:cka}
We delve deeper into the relationship between dimensionality and CKA defined by Eq.~\ref{eq:cka}
 \cite{similarityrevisit,similarity-cnn,similarity-vit,similarity-transfer,similarity-block,CKA-2001,CKA-2012}. 
In Fig.~\ref{fig:cka-layers} we present CKA for ID and OOD samples, representing the interlayer similarity of features.
Not only in ID samples but also in OOD samples, CKA saturates in deeper layers particularly for deeper models.
In addition, the saturating layer corresponds to the transition layer.
This demonstrates that after low dimensionalization, the structure of features is frozen even if performing deeper layer processing.
This is consistent with the stable feature-based detection performance for deeper layers, and with the picture in Sec.~\ref{sec:outline}.

When we carefully compare interlayer CKA of ID samples with that of OOD samples,
Fig.~\ref{fig:cka-layers} reveals that the change of CKA around the transition layer in ID samples is more gradual than that in OOD samples.
This indicates that low dimensionalization around the transition layer discards most of the intrinsic information of OOD samples while retaining information of ID samples, which is expected to be a characteristic of OOD samples as described in Sec.~\ref{sec:outline}.
Furthermore, the smoothness of CKA around the transition layer emerges as the most significant difference between ID and OOD samples,
suggesting that dimensionality is the most important factor compared with other effects including multi-layer propagation.
See Appendix~\ref{sec:app_cka-penultimate} for further comparison.

\begin{figure}[t]
\centering
\begin{tabular}{ccc}
  \includegraphics[width=0.4\hsize, bb=0.000000 0.000000 452.000000 335.000000]{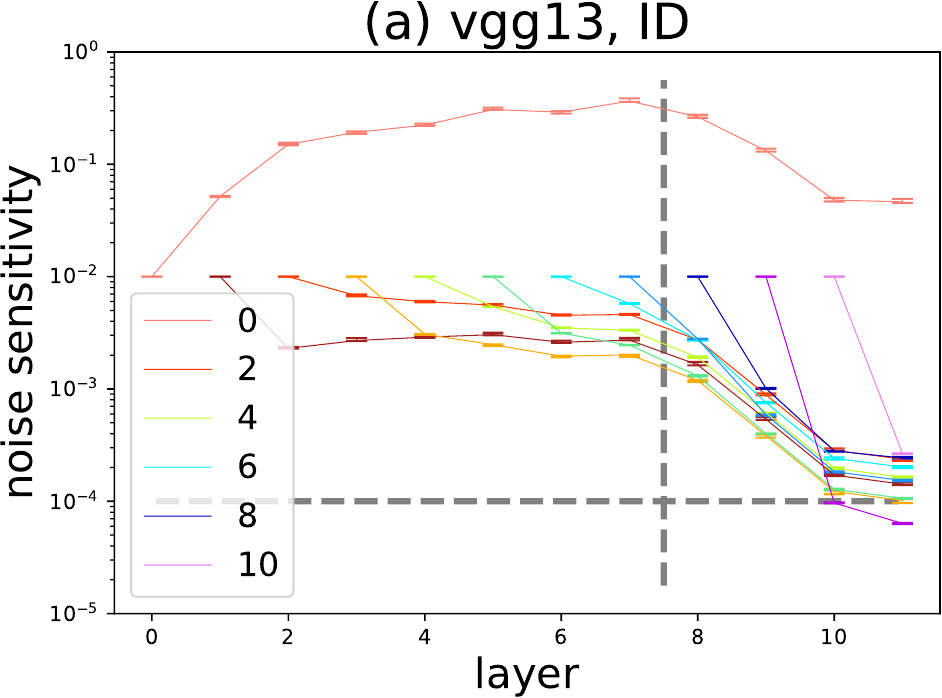} &
  \hspace{10mm} &
  \includegraphics[width=0.4\hsize, bb=0.000000 0.000000 452.000000 335.000000]{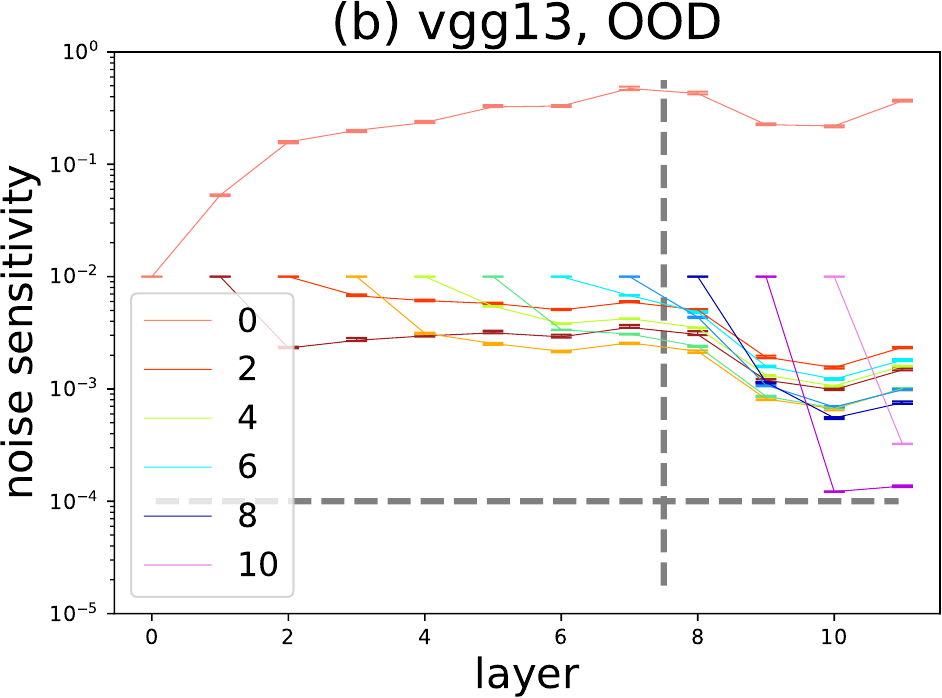}
\end{tabular}
  \caption{
    Noise sensitivity
    for the VGG-13 model.
    Left (a) and right (b) figures show the noise sensitivities
    of ID (CIFAR-10) samples and OOD (CIFAR-100) samples, respectively.
    In each figure, the horizontal axis represents the layer
    and the vertical axis represents corresponding noise sensitivity.
    Different colors indicate the input layers where noise is injected.
    The dashed vertical line indicates the transition layer.
    The dashed horizontal line is plotted to clarify the difference between ID and OOD samples.
    OOD samples are more sensitive to noise injection compared with ID samples.
    See Appendices~\ref{sec:app_full-model} and \ref{sec:app_svhn-mnist} for further verification.
  }
  \label{fig:sensitivity}
\end{figure}

\subsection{Instability of OOD samples to noise injection}
\label{sec:sensitivity}
To further investigate the multi-layer propagation property,
we evaluate noise sensitivity $\psi$ defined by Eq.~\ref{eq:noise-sensitivity} in Fig.~\ref{fig:sensitivity}.
We can see that the difference between ID and OOD samples is determined by
whether the noise-injected layer and the observed layer are deeper than the transition layer or not.
If both layers are shallower or deeper than the transition layer, the difference between ID and OOD samples is not substantial.
However, deep-layer features of ID samples exhibit much less sensitivity to shallow-layer noise compared with those of OOD samples.
The similar behavior of the former demonstrates that low-dimensional weights in deeper layers appropriately mitigate noise injected before them.
The distinct behavior of the latter suggests that shallower-layer noise to OOD samples impacts deep-layer features dominated by dataset bias, while that to ID samples is appropriately removed from deep-layer features specific to classification.
These observations provide further evidence for the picture in Sec.~\ref{sec:outline}.

One might consider to utilize the difference in noise sensitivity for OOD detection.
However, their performance falls short compared with that of the single-layer projection-based method,
although sensitivity-based detection outperforms the probability-based method.
See Appendix~\ref{sec:app_auroc-sensitivity} for more details.

\begin{figure}[t]
\centering
  \begin{tabular}{c}
    \includegraphics[width=0.4\hsize, bb=0.000000 0.000000 457.000000 314.000000]{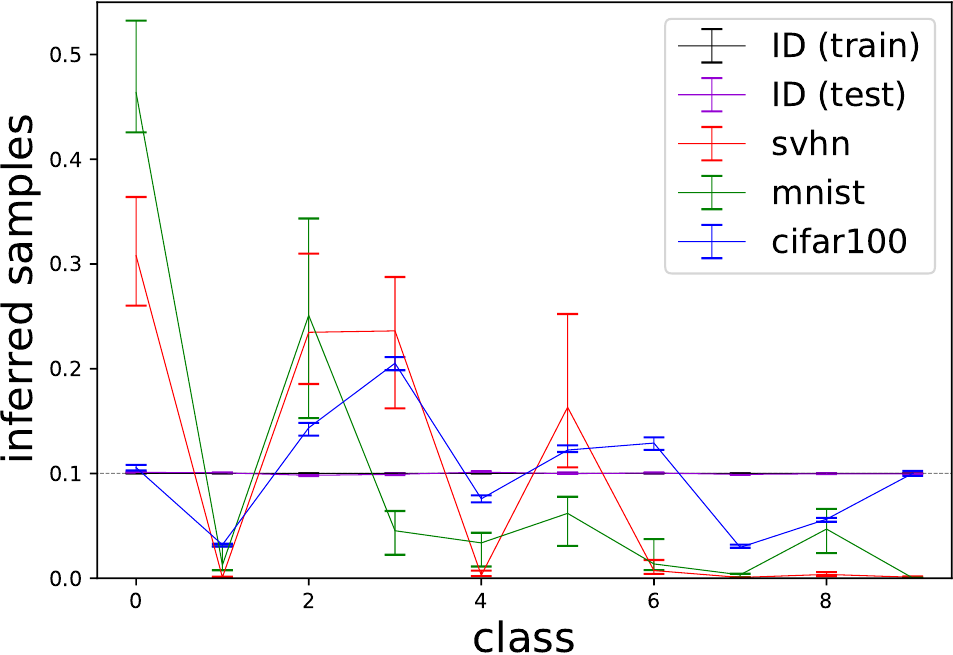}
  \end{tabular}
  \caption{
    Rates of samples predicted to belong to class 0, 1, $\cdots$, 9
    for the VGG-13 model.
    Different line colors represent different datasets.
    The prediction of OOD samples is highly imbalanced compared with that of ID (CIFAR-10) samples.
    See Appendices~\ref{sec:app_full-model} and \ref{sec:app_svhn-mnist} for further verification.
  }
  \label{fig:inference}
\end{figure}

\begin{table}[t]
  \caption{
    Summary of the coefficient of variation (ID: CIFAR-10).
    Values inside parentheses () represent the error.
    More precisely, the error is given by the average of deviations from the median to the first and third quartiles.
    For example, $1.32(5)$ means $1.32\pm 0.05$, $1.54(11)$ means $1.54\pm 0.11$.
  }
  \begin{center}
    \begin{tabular}{ccccc}
      \hline
      \multirow{2}{*}{\textbf{Dataset}}
      & \multicolumn{4}{c}{\textbf{Architectures}} \\
      & VGG13 & VGG16 & Res18 & Res34 \\
      \hline
      ID (test) & 0.01(0) & 0.01(0) & 0.01(0) & 0.01(0) \\
      CIFAR-100 & 0.52(1) & 0.52(1) & 0.51(1) & 0.50(2) \\
      SVHN & 1.32(5) & 1.33(7) & 1.45(13) & 1.43(9) \\
      MNIST & 1.54(11) & 1.58(9) & 1.45(12) & 1.40(16)\\
      \hline
    \end{tabular}
    \label{tab:infer_cv}
  \end{center}
\end{table}

\subsection{Dataset bias-induced imbalanced inference}
\label{sec:inference}
To demonstrate the impact of dataset bias on classification,
we examine the prediction of OOD samples for each class in Fig.~\ref{fig:inference}.
We can see that the prediction of ID (CIFAR-10) samples is well balanced due to the equivalent number of training samples for each class.
For OOD samples, however, the imbalanced prediction is significant and seems to be common for all network architectures.
This suggests that trained networks behave similarly regardless of specific architectures.
A more quantitative comparison is provided by the coefficient of variation in Tab.~\ref{tab:infer_cv}, corresponding to the standard deviation of the prediction rates to each class.
Compared with the coefficient of variation of ID (CIFAR-10) samples, those of OOD samples are extremely large even in the case of close-to-ID OOD (CIFAR-100) samples.
This dataset-dependent imbalanced prediction supports the residual bias of the dataset.

\begin{table}[t]
  \caption{
    Summary of OOD detection performance (AUROC) for VGG-13 trained on CIFAR-10.
    All results were reproduced in our experiments.
    Values inside parentheses () represent the error.
  }
  \begin{center}
    \begin{tabular}{cccc}
      \hline
      \multirow{2}{*}{\textbf{Detection score}}
      & \multicolumn{3}{c}{\textbf{OOD dataset}} \\
      & CIFAR-100 & SVHN & MNIST \\
      \hline
      Probability &
      0.886(1) & 0.944(5) & 0.927(7) \\
      Feature &
      0.898(1) & 0.948(4) & 0.936(5) \\
      Projection &
      0.70(5) & 0.62(7) & 0.47(16) \\
      Projection (ours) &
      \textbf{0.901(1)} & \textbf{0.960(4)} & \textbf{0.978(4)} \\
      \hline
    \end{tabular}
    \label{tab:comparison}
  \end{center}
\end{table}

\subsection{Quantitative comparison of OOD detection performance}
\label{sec:quantitative}
Finaly, we demonstrate the importance of dimensionality through a quantitative comparison of OOD detection performance.
Tab.~\ref{tab:comparison} summarizes the OOD detection performance detected by the ratio of norm $||x_{p,\varepsilon}^{(l)}|| / ||x^{(l)}||$ at the penultimate fully connected layer.
As discussed in Sec.~\ref{sec:related_projection}, our method is quite similar to NuSA in this case \cite{OOD-NuSA}.
However, NuSA employs full projections ($\varepsilon=0$), while we eliminate irrelevant singular vectors from the projection matrix based on corresponding singular values (typically $\varepsilon\sim 10^{-2}$).
See Appendix~\ref{sec:app_compress} for more details.
This minor modification markedly improves not only detection performance but also its stability.
Furthermore, our dimensionality-aware projection-based method is comparable to or better than both the probability-based method \cite{OOD-baseline} and the feature-based method \cite{OOD-Mahalanobis}.
In addition, our dimensionality-aware projection-based method is superior to the feature-based method in terms of computational cost.
This is mainly because the feature-based method requires a huge null space for accurate feature-based detection, which is in contrast to our method eliminating a null space.
See Appendix~\ref{sec:app_comprehensive} for comprehensive comparison.

\section{Discussion}
\label{sec:discussion}

As described in Sec.~\ref{sec:results},
dimensionality plays an essential role in the propagation process,
which naturally suggests a simple picture of feature propagations of ID and OOD samples.
We now discuss how various properties observed in previous studies are derived from the proposed picture.
We also argue the close connection with generalization of ID samples.


Our picture given in Sec.~\ref{sec:outline} can explain how observed properties of OOD samples in previous studies emerge from the information processing inside the DNN.
The low-dimensional weight matrices first eliminate most features of the OOD samples.
The features of OOD samples then aggregate around the origin with only dataset bias retained.
This aggregation persists untill the final layer, yielding the small logit values of OOD samples utilized in so many studies \cite{OOD-baseline,OOD-ODIN,OOD-energy,OSR-IDperform,OSR-familiarity,OOD-ReAct,OOD-DICE,OOD-ViM}.
Meanwhile, dataset bias appears in the null space of the subsequent weight matrices, providing the large deviation from ID samples distributed in the linear span of principal components of the weight matrix.
This orthogonality improves feature-based and projection-based detection \cite{OOD-Mahalanobis,OOD-Mahalanobis-PCA,OOD-feature-subspace,OOD-NuSA,OOD-ViM}.
Furthermore, residual dataset bias in each OOD sample contains poor information, leading to the concentration of each feature at the largest singular value \cite{OOD-RankFeat}.
Our picture provides insights into how these methods work well based on the low-dimensional property of DNNs, which could help understand the nature of OOD samples in DNNs.

The importance of dimensionality can also explain the positive correlation between ID generalization and OOD detection \cite{OSR-IDperform}, and the hardness of OOD detection in large-scale classification \cite{OOD-MOS}.
In the context of ID generalization error analysis, the importance of low dimensionalization has also been noted \cite{Arora-compression,Suzuki-compression,stableranknorm-optim}.
Combined with our results, a positive correlation between ID generalization and OOD detection is directly derived.
In addition, we observe poor OOD detection performance in cases with a large number of classes in ID samples due to insufficient low dimensionalization.
(See Appendix~\ref{sec:app_cifar100}.)
This is also consistent with the achievement of good ID generalization and OOD detection as long as the internal feature dimension is sufficiently larger than the output dimension.

The relationship between ID optimization and OOD detection is also likely to correlate with the data complexity dependence of the suitable detection layer \cite{OOD-MOOD}.
The optimization of the model using ID samples induces an overfitting-like behavior in ID samples with reference to their data complexity.
The loss of information is thus more significant for OOD samples with distinct data complexity from ID samples, leading to dataset bias-dominated features even at shallower layers.
This might be the reason why the feature-based detection method performs extremely well in shallower layers for far-from-ID OOD samples as in Fig.~\ref{fig:auroc} (a), and why CKA of far-from-ID OOD samples tends to be retained in shallower layers as in Appendix~\ref{sec:app_cka-penultimate}.
Furthermore, when we closely compare AUROCs of various layers with the data complexity provided by Ref.~\cite{OOD-MOOD}, rather than the data complexity itself, the difference in data complexity between ID and OOD samples determines the suitable layer for OOD detection.
In this sense, the propagation properties investigated in our study are expected to correlate with the data complexity-based method, although further studies are required to clarify this.

\section{Summary}
\label{sec:summary}
Based on various analyses investigating how ID and OOD samples propagate inside DNNs,
we demonstrated the importance of dimensionality for various properties of OOD samples.
These observations might serve as the baseline for understanding the nature of OOD samples in DNNs and improving OOD detection performance.

\begin{credits}
\subsubsection{\ackname}
TS was partially supported by JSPS KAKENHI (24K02905) and JST CREST (JPMJCR2015).


\subsubsection{\discintname}
The authors have no competing interests to declare that are relevant to the content of this article.

\end{credits}


\clearpage

\appendix

\renewcommand{\thefigure}{S\arabic{figure}}
\renewcommand{\thetable}{S\arabic{table}}

\begin{figure}[t]
\centering
  \begin{tabular}{cccc}
    \includegraphics[width=0.22\hsize, bb=0.000000 0.000000 460.800000 345.600000]{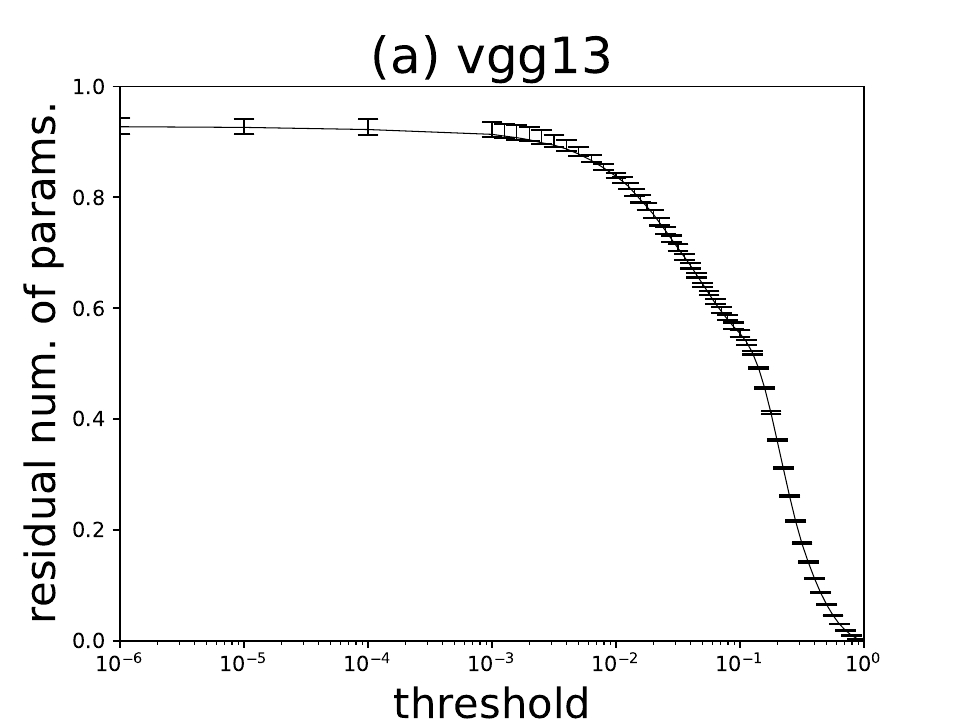}&
    \includegraphics[width=0.22\hsize, bb=0.000000 0.000000 460.800000 345.600000]{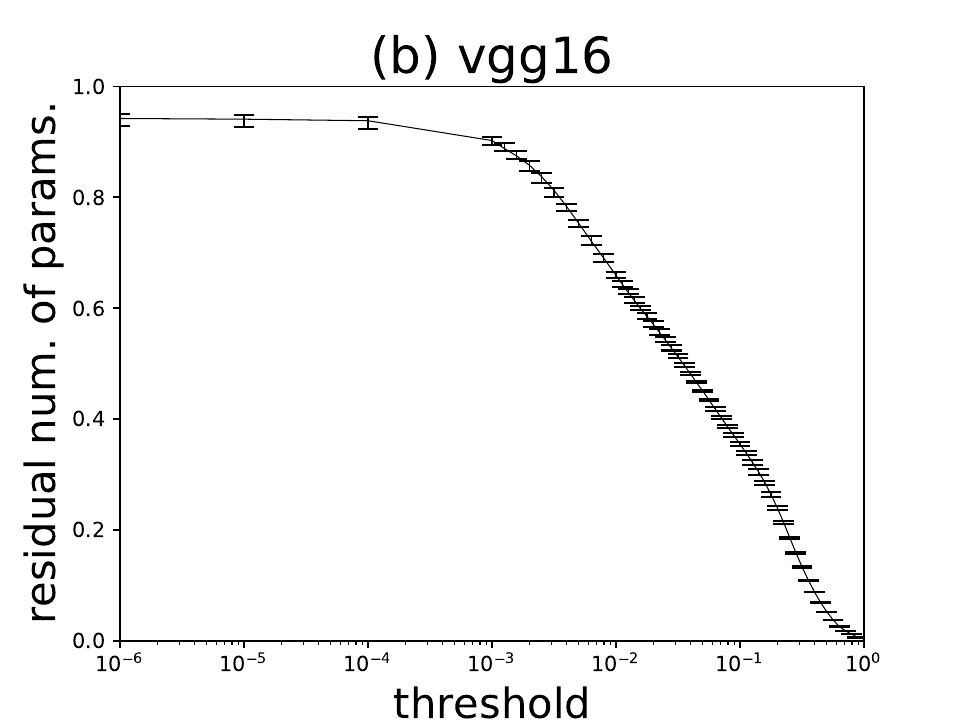}&
    \includegraphics[width=0.22\hsize, bb=0.000000 0.000000 460.800000 345.600000]{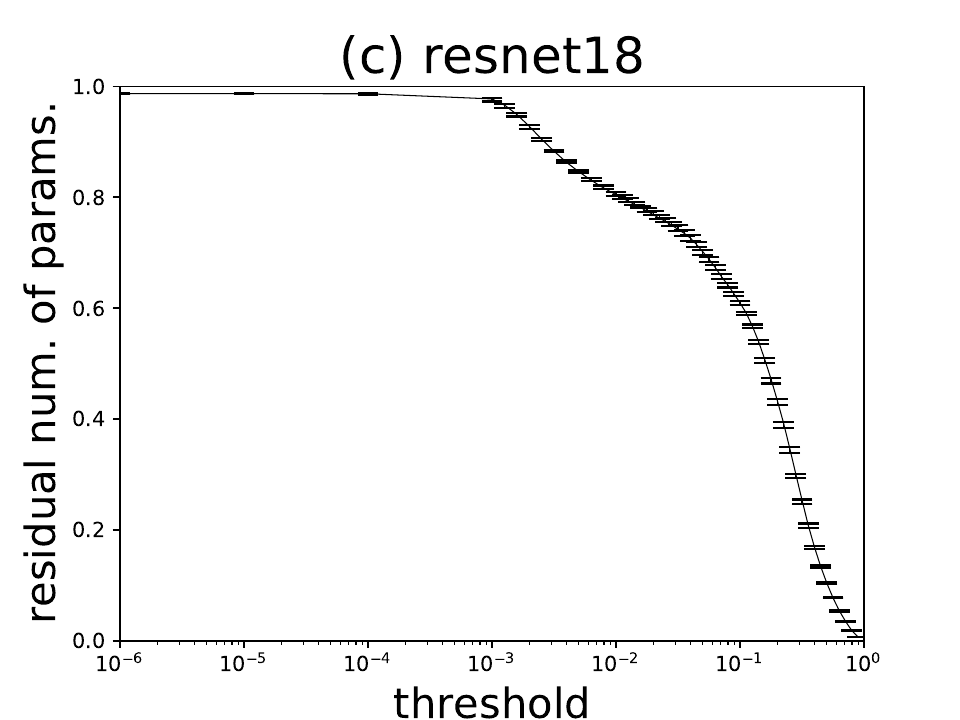}&
    \includegraphics[width=0.22\hsize, bb=0.000000 0.000000 460.800000 345.600000]{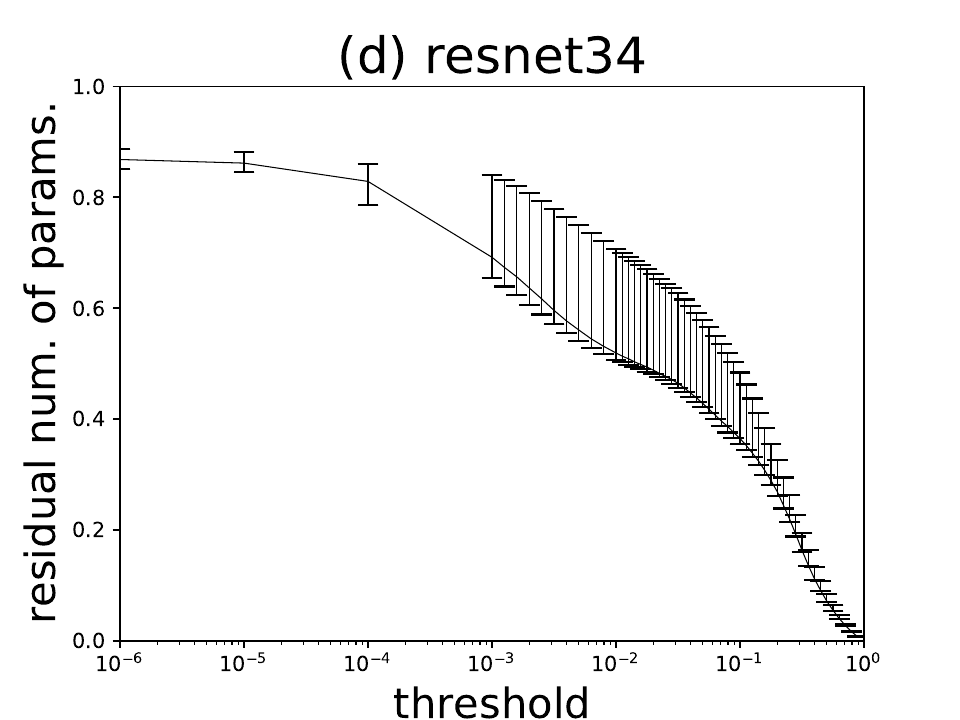}
  \end{tabular}
  \caption{
    Threshold $\varepsilon$ dependence of the residual number of parameters
    for the (a) VGG-13, (b) VGG-16, (c) ResNet-18, and (d) ResNet-34 models.
  }
  \label{fig:app_compress-nparams}
\end{figure}

\begin{figure}[t]
\centering
  \begin{tabular}{cccc}
    \includegraphics[width=0.22\hsize, bb=0.000000 0.000000 460.800000 345.600000]{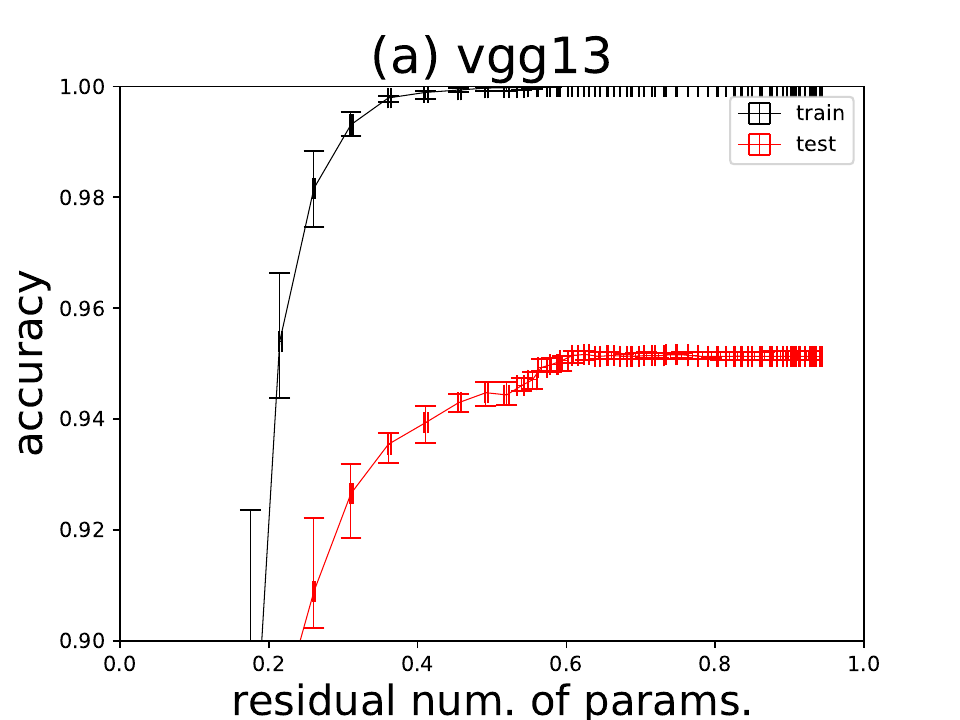}&
    \includegraphics[width=0.22\hsize, bb=0.000000 0.000000 460.800000 345.600000]{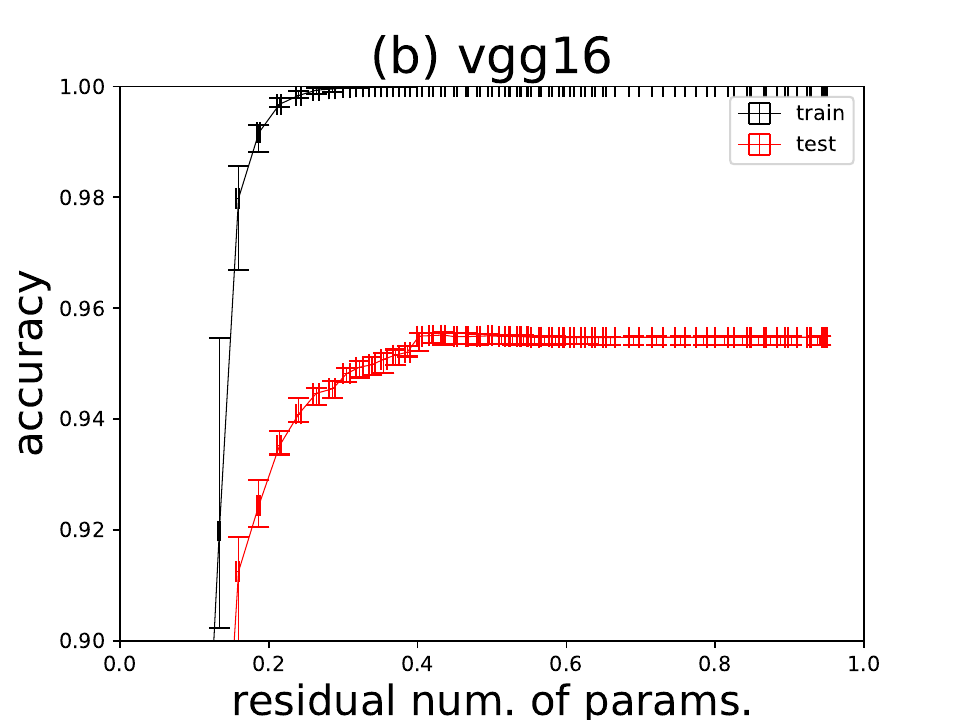} &
    \includegraphics[width=0.22\hsize, bb=0.000000 0.000000 460.800000 345.600000]{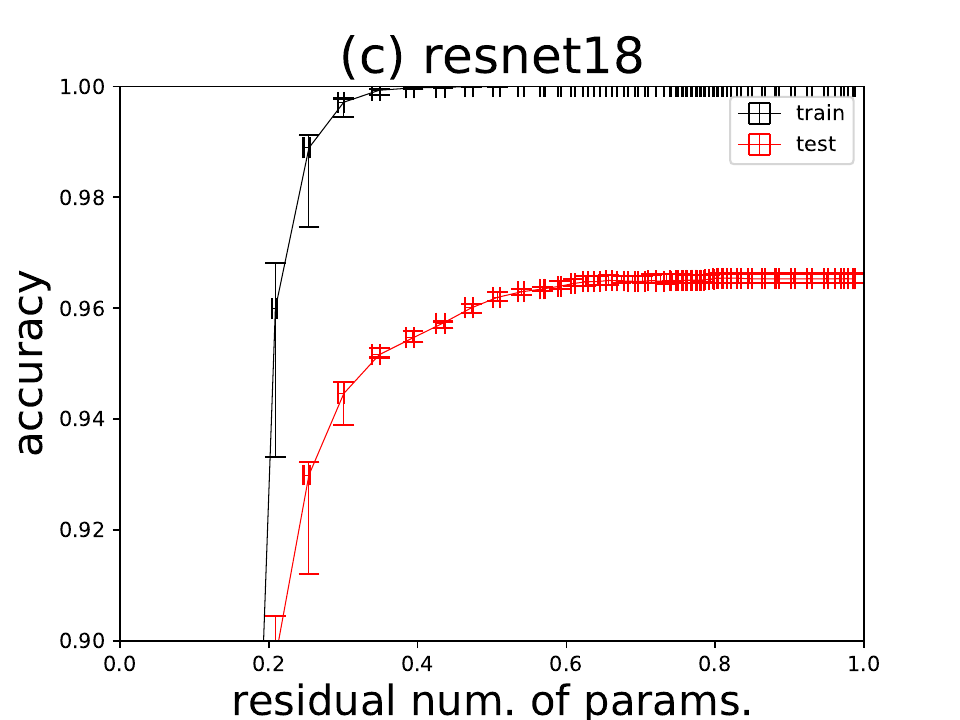}&
    \includegraphics[width=0.22\hsize, bb=0.000000 0.000000 460.800000 345.600000]{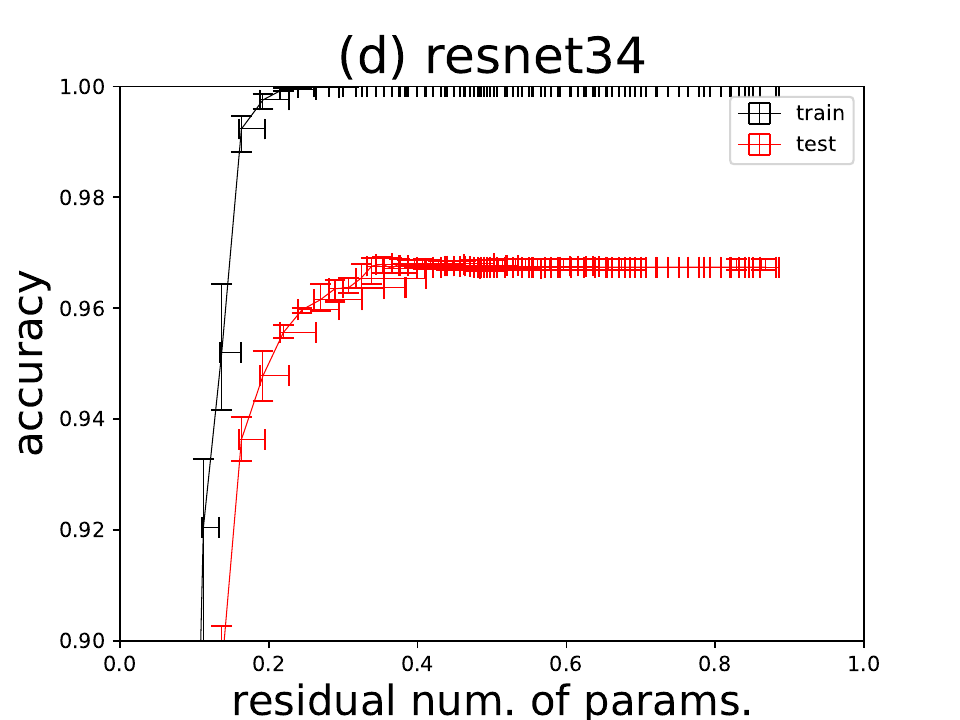}
  \end{tabular}
  \caption{
    Compression rate dependence of the ID (CIFAR-10) classification accuracy
    for the (a) VGG-13, (b) VGG-16, (c) ResNet-18, and (d) ResNet-34 models.
  }
  \label{fig:app_compress-accuracy}
\end{figure}

\section{Effect of compression}
\label{sec:app_compress}
In this Appendix section, we present details of the projection-based OOD detection.
We utilize the alignment of model parameters and features via the projections of features to weight matrices with their singular values considered.
More precisely, we conduct the following procedure.
\begin{enumerate}
  \item Perform singular-value decomposition of the $l$-th layer weight matrix $W^{(l)}$: $W^{(l)}=U^{(l)}S^{(l)}(V^{(l)})^{\mathsf T}$.
  \item Remove irrelevant singular values such that $s_k^{(l)}/s_{0}^{(l)} < \varepsilon$ where $s_k^{(l)}$ is the $k$-th largest singular value of $W^{(l)}$, i.e., $s_0 \geq s_1 \geq \cdots $.
  \item Construct the dimensionality-aware projection matrix $V_{\varepsilon}^{(l)}$ from singular vectors corresponding to remaining singular values.
  \item Project the $l$-th layer feature $x^{(l)}$ to the weight as $x_{p,\varepsilon}^{(l)}=(V_{\varepsilon}^{(l)})^{\mathsf T} x^{(l)}$.
  \item Regard samples with smaller values of $||x_{p,\varepsilon}^{(l)}|| / ||x^{(l)}|| $ or $||x_{p,\varepsilon}^{(l)}||$ as OOD samples.
\end{enumerate}
$\varepsilon$ is typically taken to be $10^{-2}$.
In the following, we check that the overall behavior is not sensitive to the change in $\varepsilon$ in a moderate range where the ID classification accuracy is retained, although it changes the observed values in a quantitative sense.

Fig. \ref{fig:app_compress-nparams} exhibits the threshold $\varepsilon$ dependence of the residual number of parameters in the DNN in the range $10^{-6}\leq \varepsilon < 1$:
the accuracy versus the residual number of parameters is shown in Fig. \ref{fig:app_compress-accuracy}.
The residual number of parameters is normalized by the original one; that is, it is maximally 1 for non-compressed network and almost 0 when most of parameters are removed.
These results indicate that the ID test accuracy can be retained even if we remove around half of the parameters.

\begin{figure}[t]
\centering
  \begin{tabular}{cccc}
    \includegraphics[width=0.22\hsize, bb=0.000000 0.000000 460.800000 345.600000]{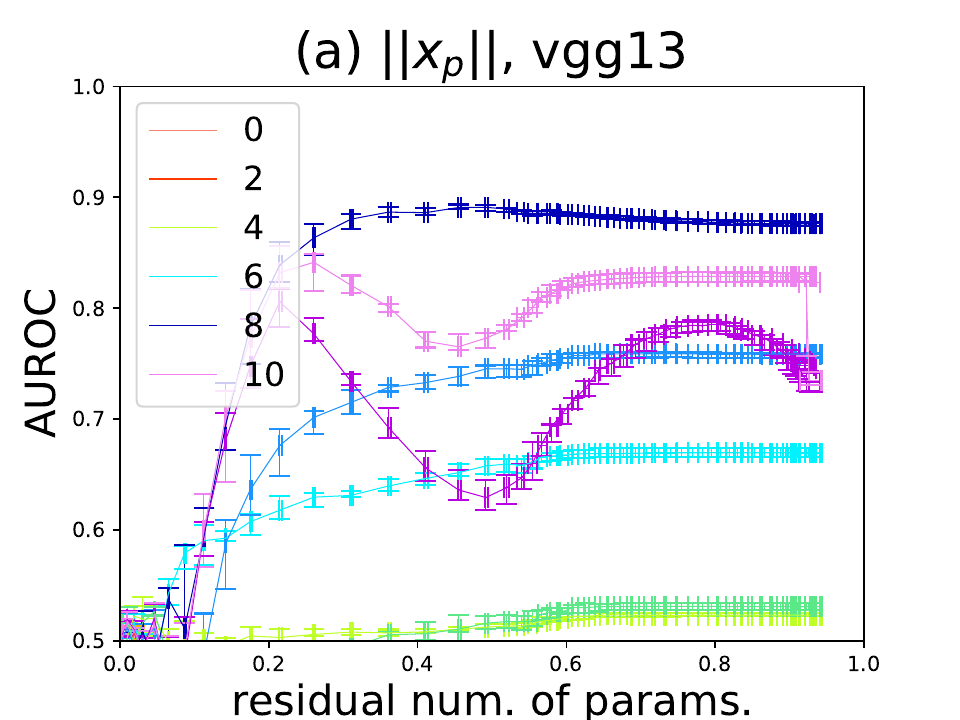}&
    \includegraphics[width=0.22\hsize, bb=0.000000 0.000000 460.800000 345.600000]{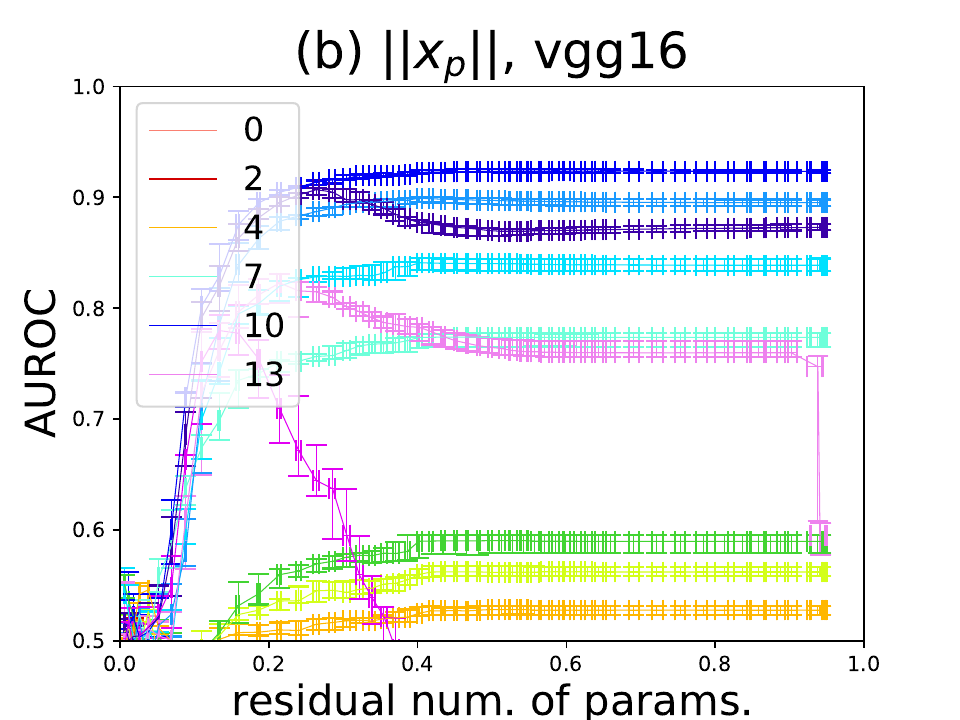}&
    \includegraphics[width=0.22\hsize, bb=0.000000 0.000000 460.800000 345.600000]{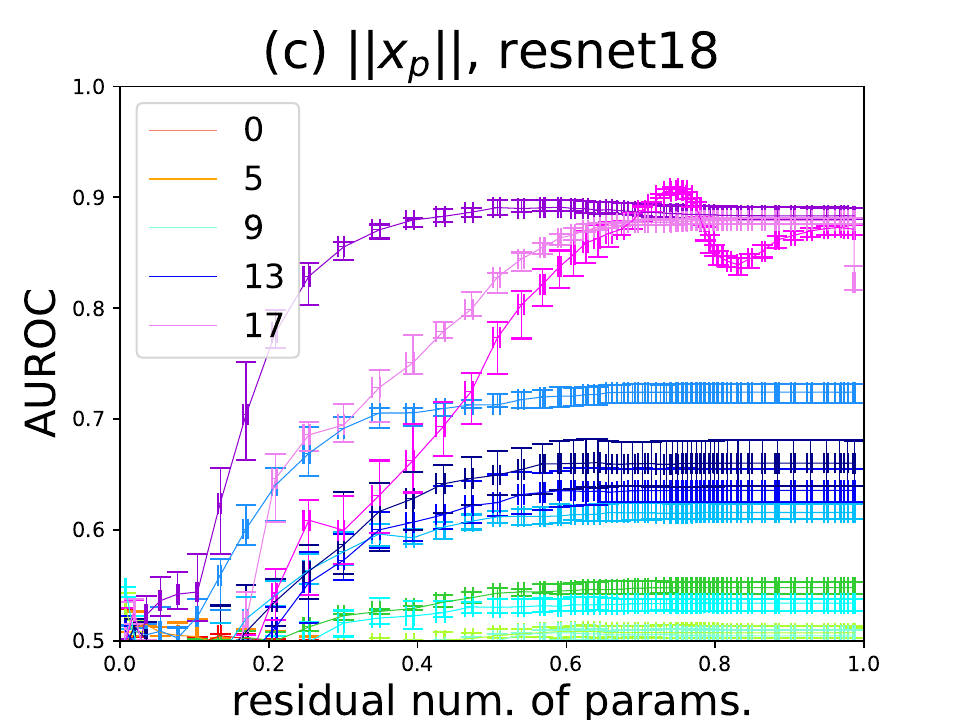}&
    \includegraphics[width=0.22\hsize, bb=0.000000 0.000000 460.800000 345.600000]{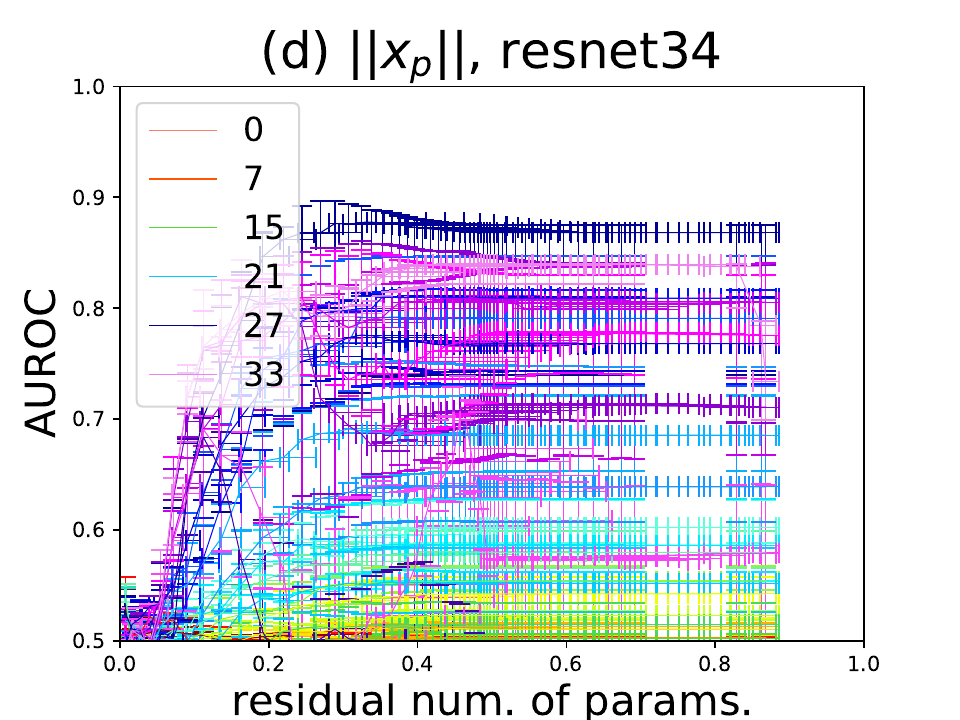}
    \\
    \includegraphics[width=0.22\hsize, bb=0.000000 0.000000 460.800000 345.600000]{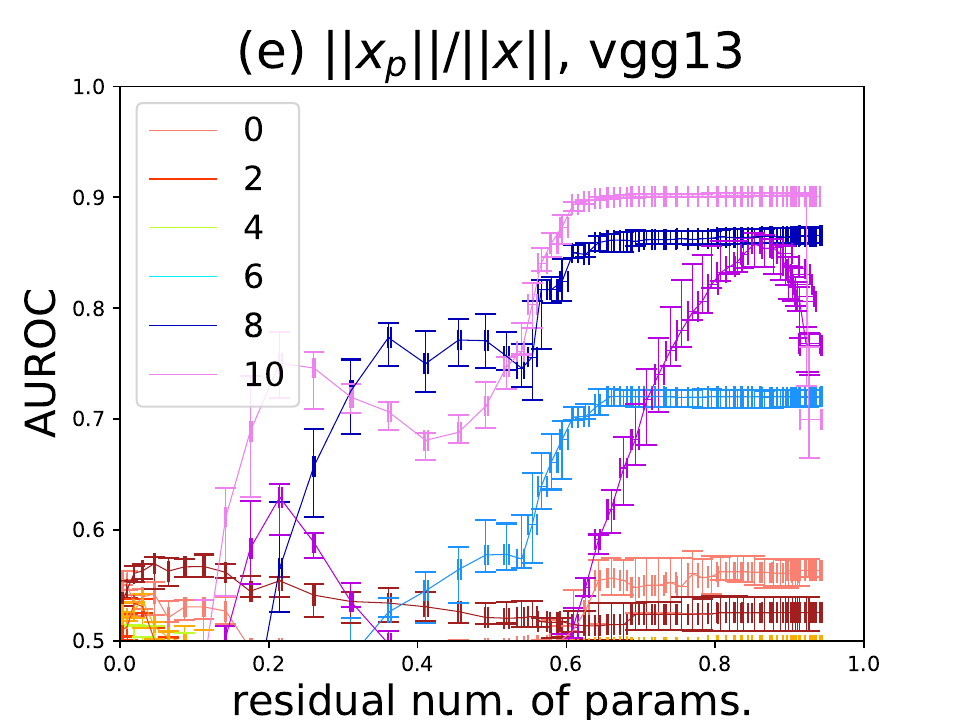}&
    \includegraphics[width=0.22\hsize, bb=0.000000 0.000000 460.800000 345.600000]{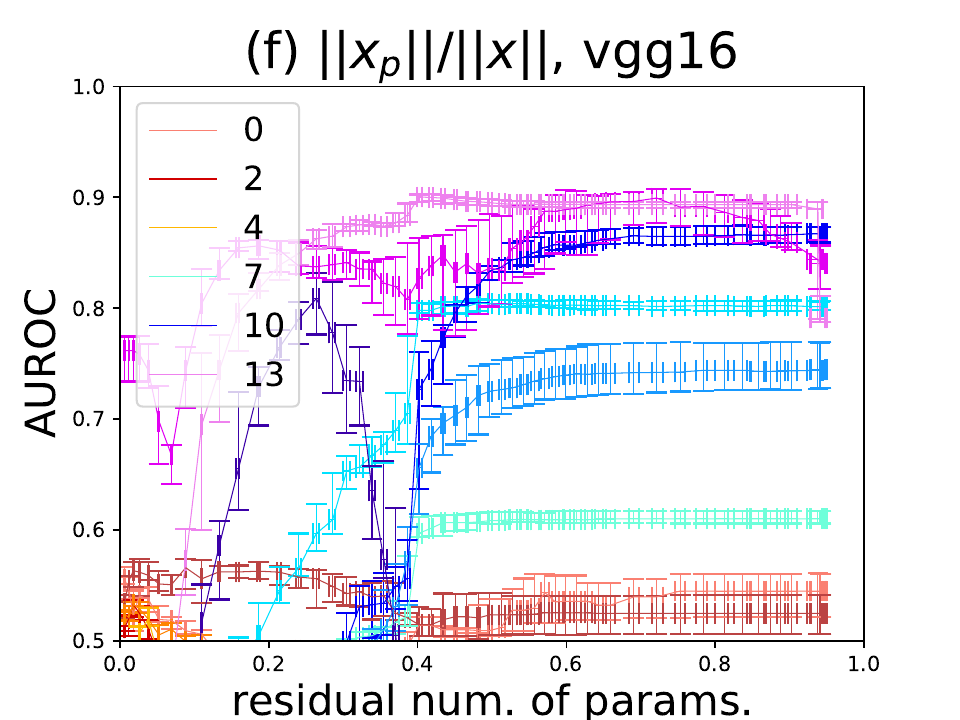}&
    \includegraphics[width=0.22\hsize, bb=0.000000 0.000000 460.800000 345.600000]{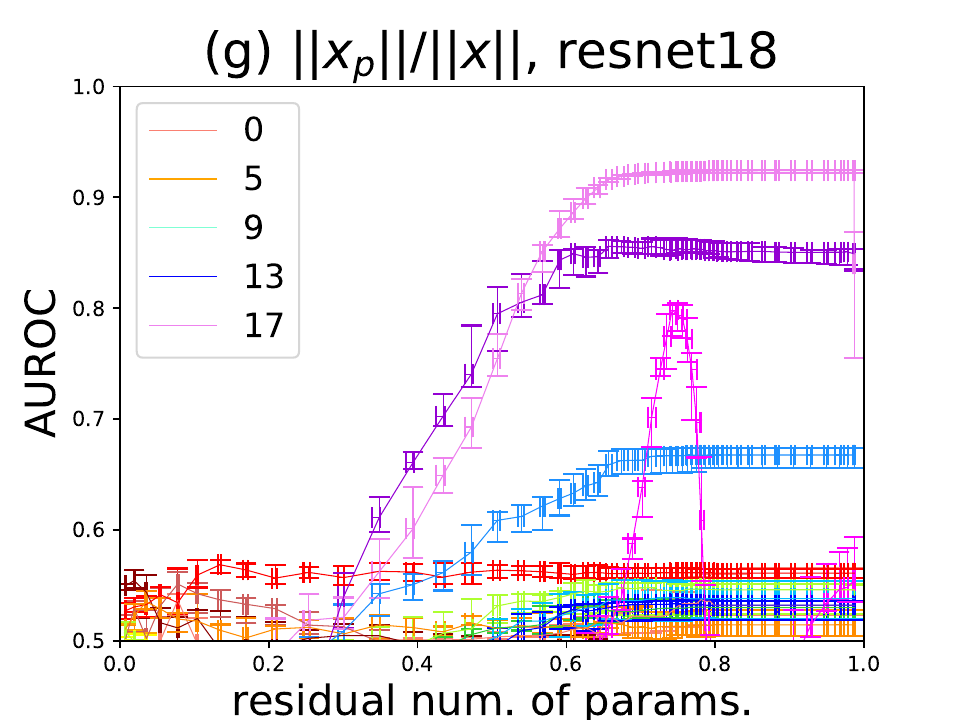}&
    \includegraphics[width=0.22\hsize, bb=0.000000 0.000000 460.800000 345.600000]{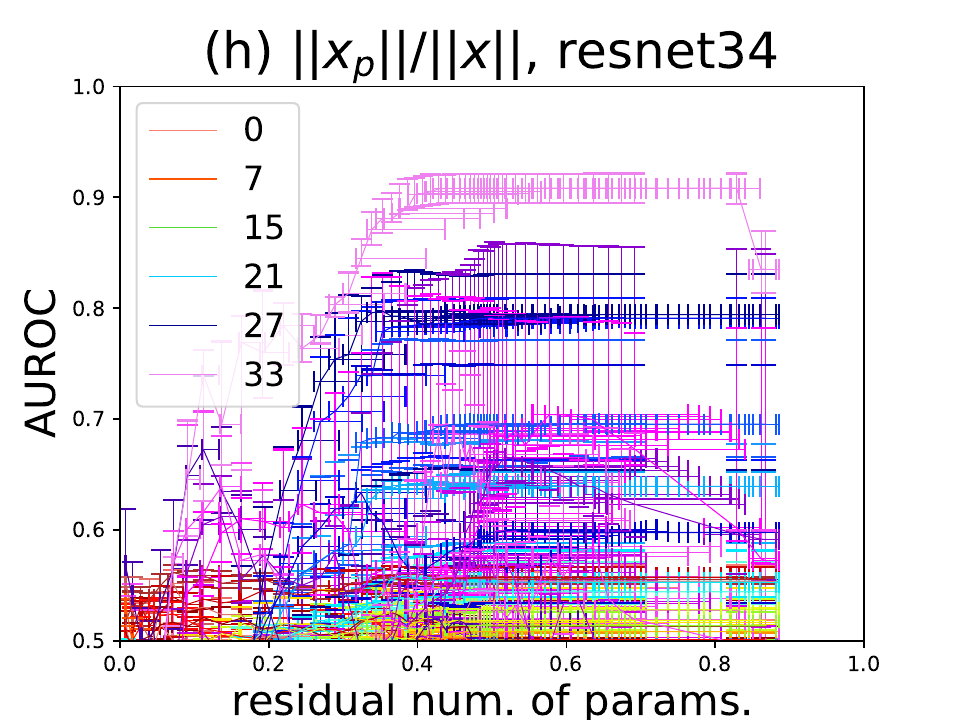}
  \end{tabular}
  \caption{
    Compression rate dependence of the OOD (CIFAR-100) detection AUROC
    for the (a,e) VGG-13, (b,e) VGG-16, (c,e) ResNet-18, and (d,e) ResNet-34 models.
    The detection scores adopted in the upper (a--d) and lower (e-h) figures are $||x_{p,\varepsilon}||$ and $||x_{p,\varepsilon}||/||x||$, respectively.
    The colors of the lines represent the layers used for detection.
  }
  \label{fig:app_compress-auroc_cifar10-cifar100}
\end{figure}

To observe full OOD detection performance, in Fig. \ref{fig:app_compress-auroc_cifar10-cifar100},
we show the AUROC vs the residual number of parameters for various layers.
The AUROC is measured using the dataset CIFAR-10 as ID and CIFAR-100 as OOD.
We can see that, except for the region with less parameters, the AUROCs are flat for most layers, indicating that the value of $\varepsilon$ does not affect the overall layer dependence shown in Sec. \ref{sec:performance} in the main text.
Also, a few AUROCs dependent on the number of parameters are not among those exhibiting the best AUROC compared with other layers.

\begin{figure}[t]
\centering
  \begin{tabular}{cccc}
    \includegraphics[width=0.22\hsize, bb=0.000000 0.000000 460.800000 345.600000]{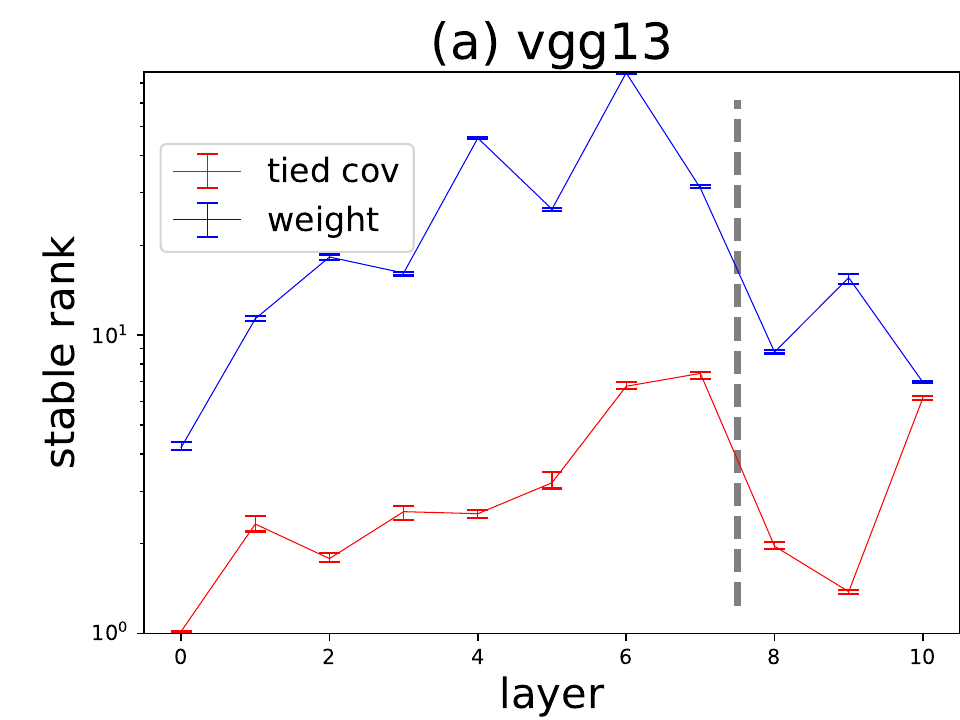}&
    \includegraphics[width=0.22\hsize, bb=0.000000 0.000000 460.800000 345.600000]{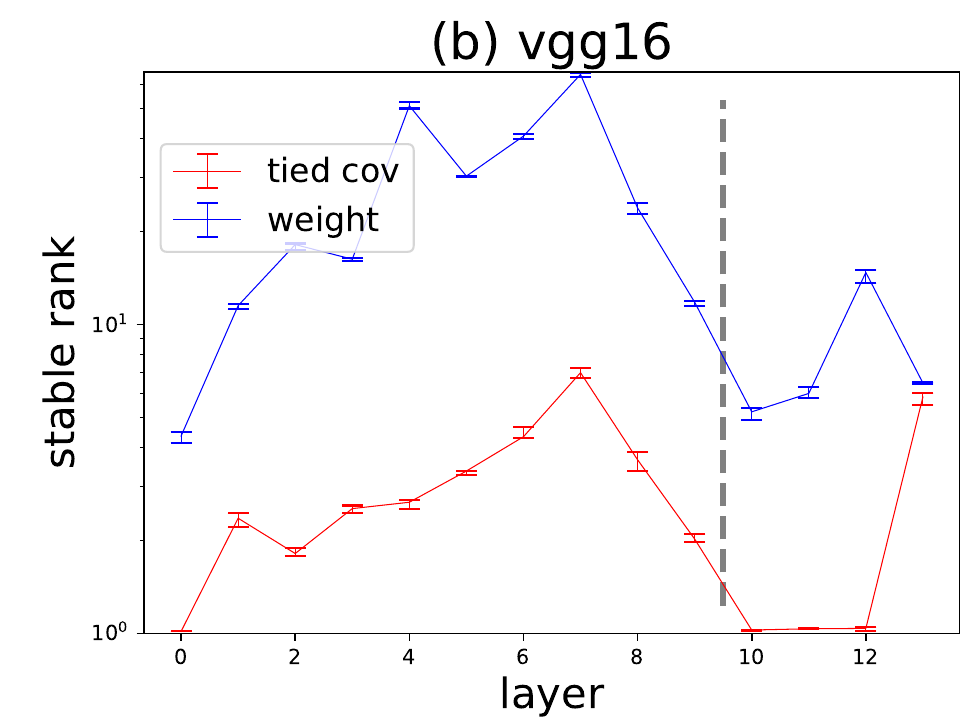}&
    \includegraphics[width=0.22\hsize, bb=0.000000 0.000000 460.800000 345.600000]{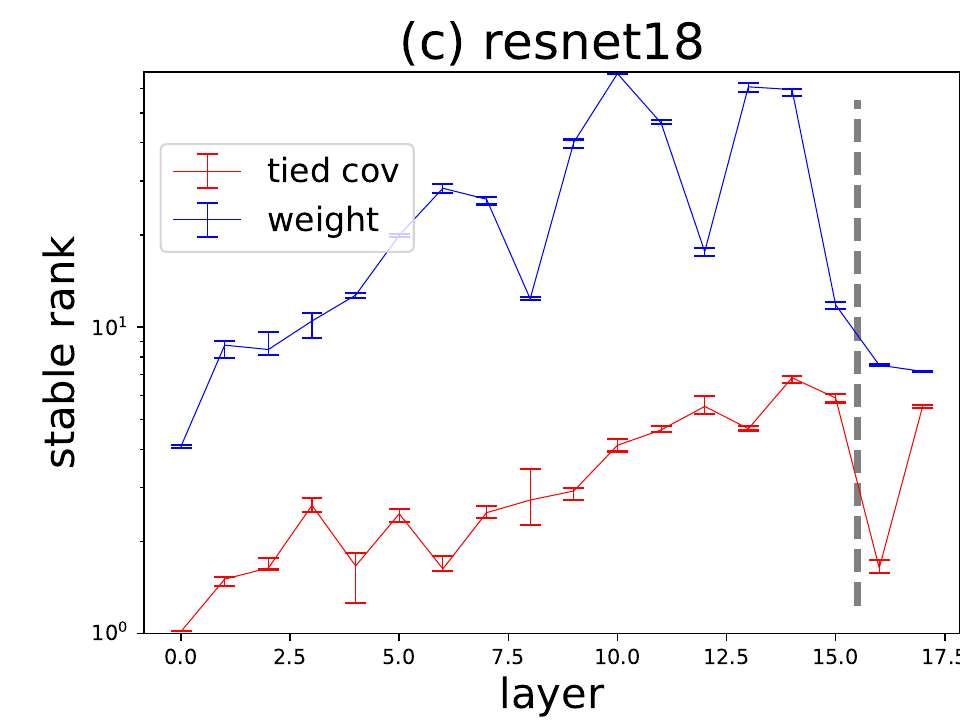}&
    \includegraphics[width=0.22\hsize, bb=0.000000 0.000000 460.800000 345.600000]{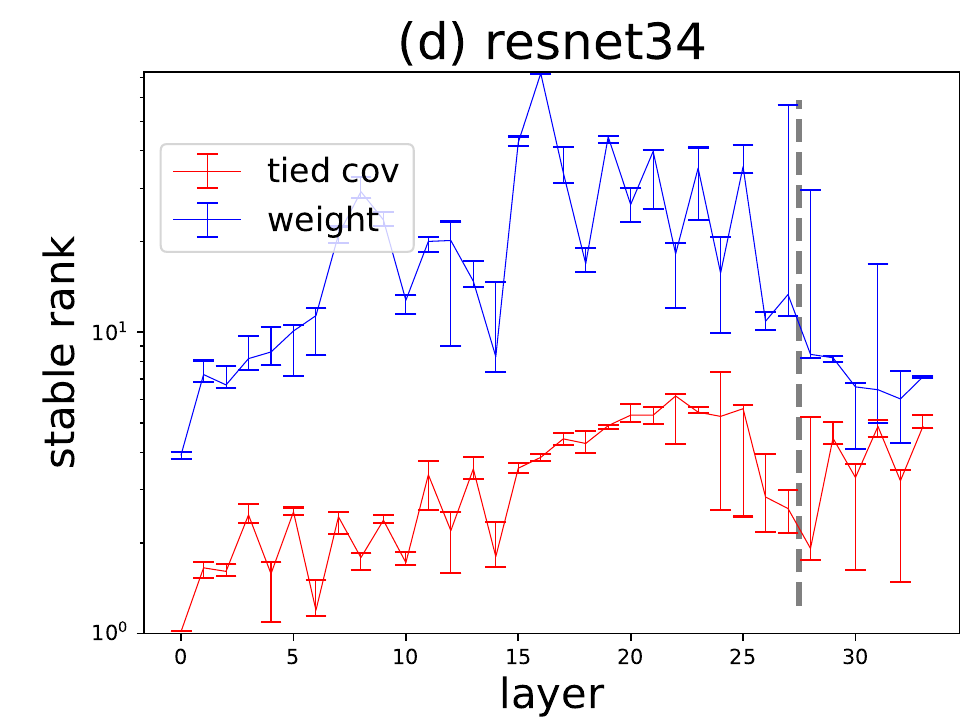}
  \end{tabular}
  \caption{
    Layer dependence of the stable rank of the covariance matrix $\overline{\Sigma}$ and the weight matrix $W$
    for the (a) VGG-13, (b) VGG-16, (c) ResNet-18, and (d) ResNet-34 models.
    The dashed line indicates the transition layer.
    Low dimensionalization of features and weights occurs at almost the same layer.
    See Sec. \ref{sec:stablerank} in the main text for more details.
  }
  \label{fig:app_stableranks}
\end{figure}

\begin{figure}[t]
\centering
  \begin{tabular}{cccc}
    \includegraphics[width=0.22\hsize, bb=0.000000 0.000000 460.800000 345.600000]{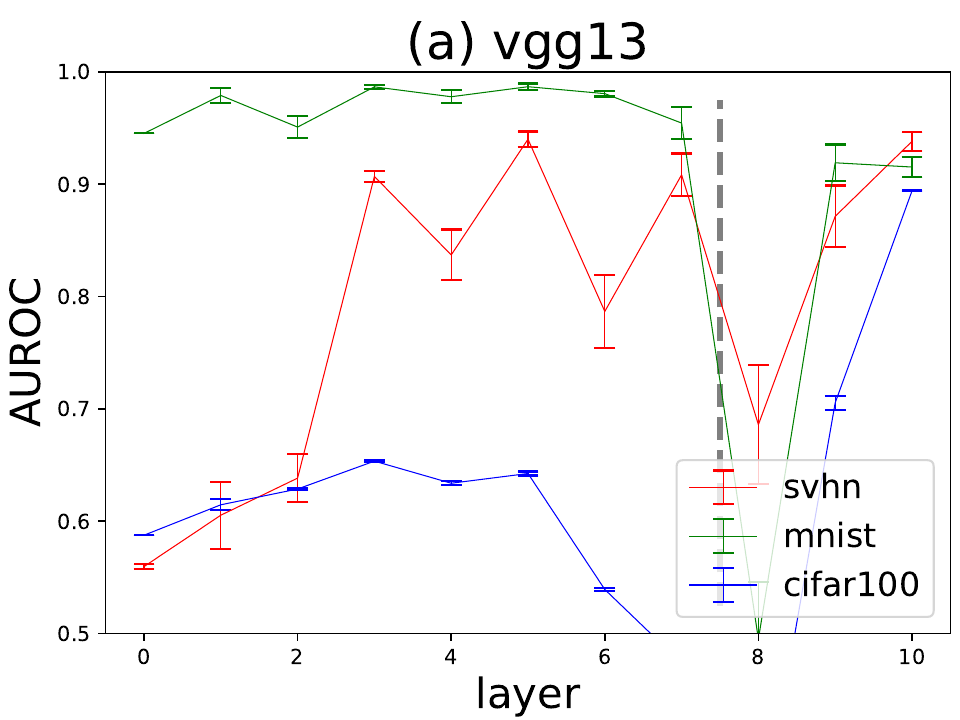}&
    \includegraphics[width=0.22\hsize, bb=0.000000 0.000000 460.800000 345.600000]{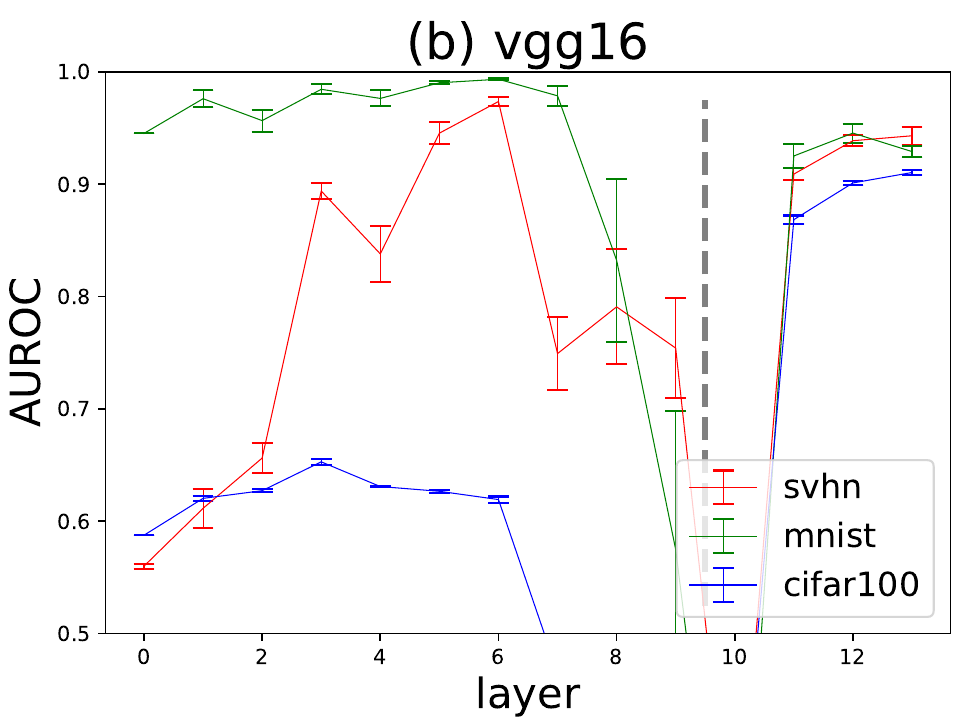}&
    \includegraphics[width=0.22\hsize, bb=0.000000 0.000000 460.800000 345.600000]{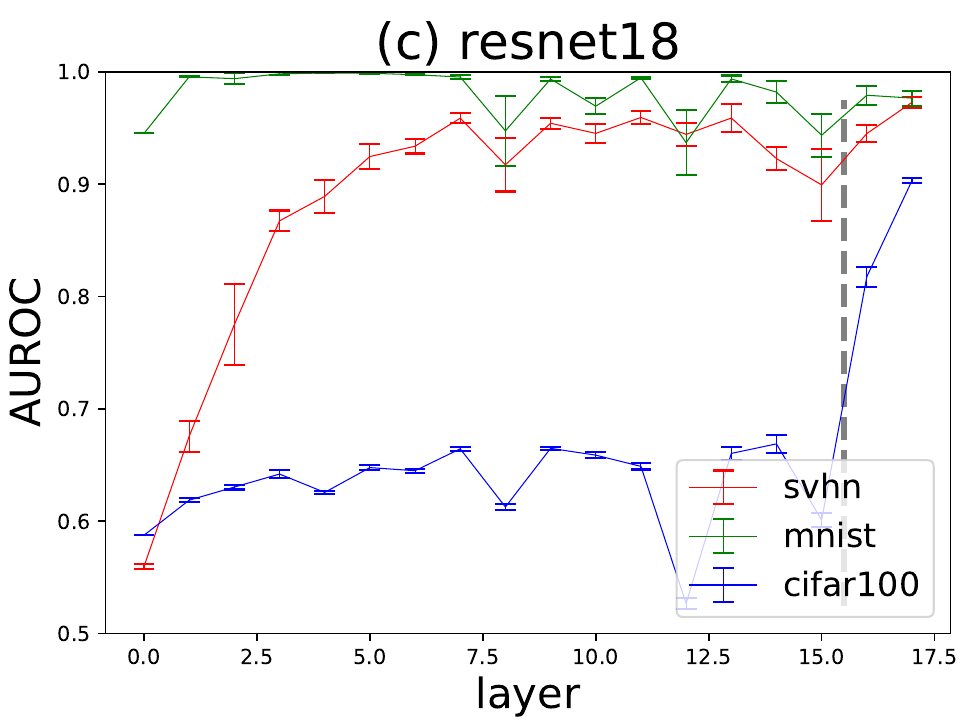}&
    \includegraphics[width=0.22\hsize, bb=0.000000 0.000000 460.800000 345.600000]{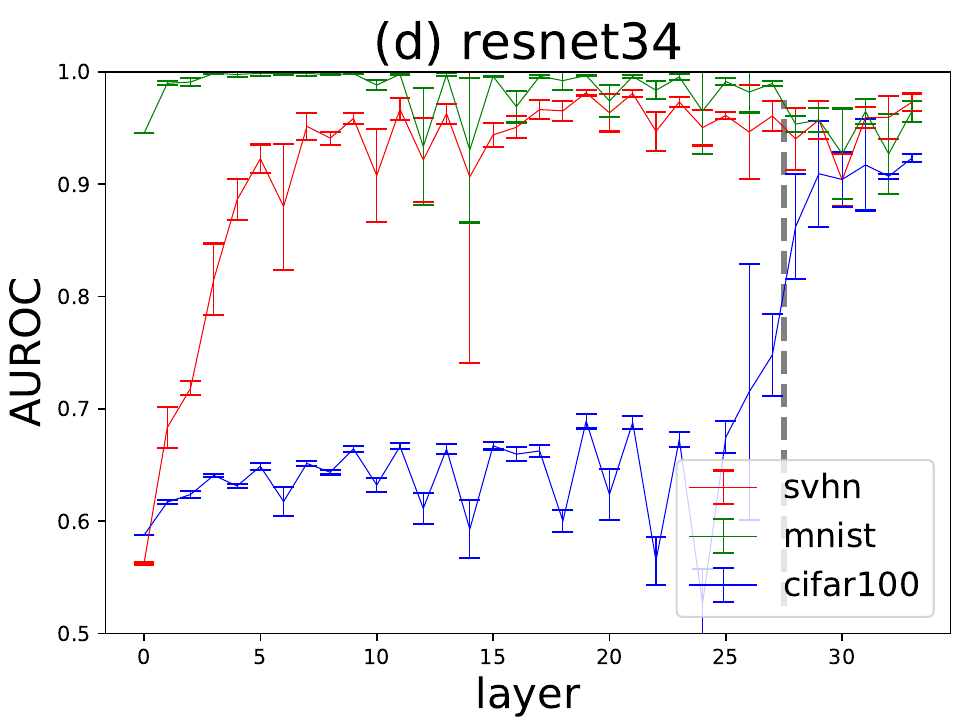}
  \end{tabular}
  \caption{
    Layer dependence of the AUROC detected by the Mahalanobis distance $M(x)$
    for the (a) VGG-13, (b) VGG-16, (c) ResNet-18, and (d) ResNet-34 models.
    Different line colors represent the different OOD datasets evaluated.
    The dashed line indicates the transition layer.
    The measured AUROCs are stabilized after transition independent of the models and datasets.
    See Sec. \ref{sec:performance} in the main text for more details.
  }
  \label{fig:app_auroc-cov}
\end{figure}

\begin{figure}[t]
\centering
  \begin{tabular}{cccc}
    \includegraphics[width=0.22\hsize, bb=0.000000 0.000000 460.800000 345.600000]{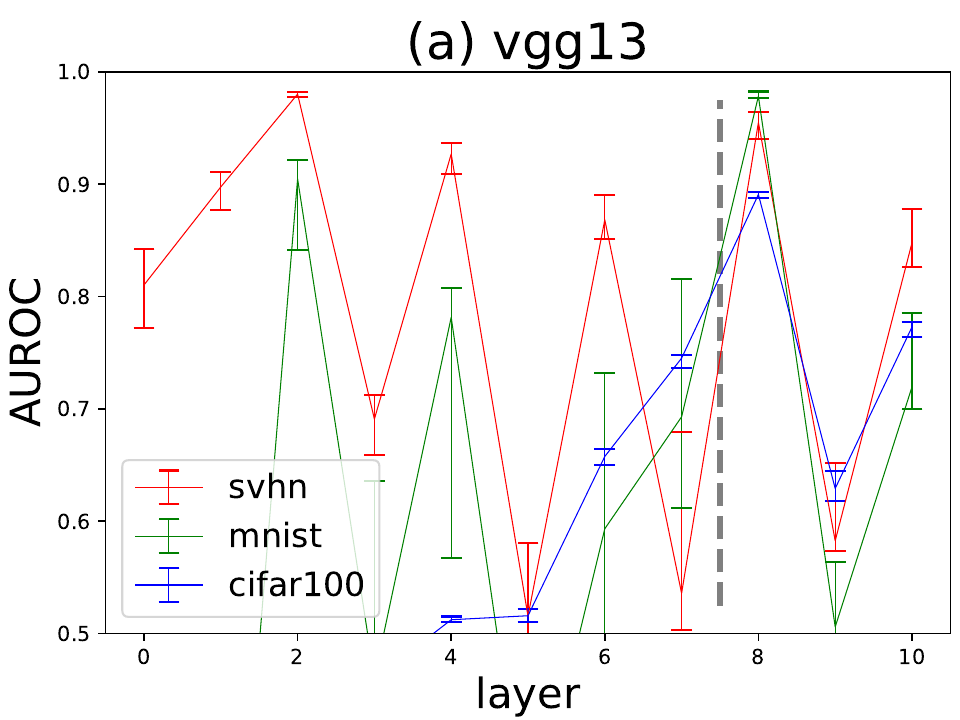}&
    \includegraphics[width=0.22\hsize, bb=0.000000 0.000000 460.800000 345.600000]{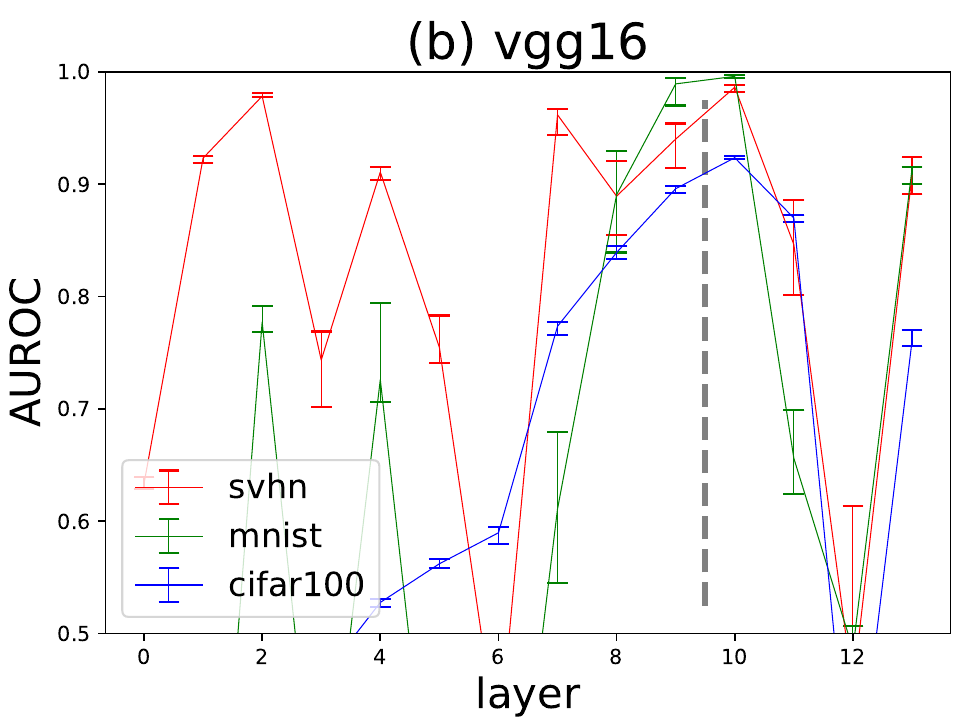}&
    \includegraphics[width=0.22\hsize, bb=0.000000 0.000000 460.800000 345.600000]{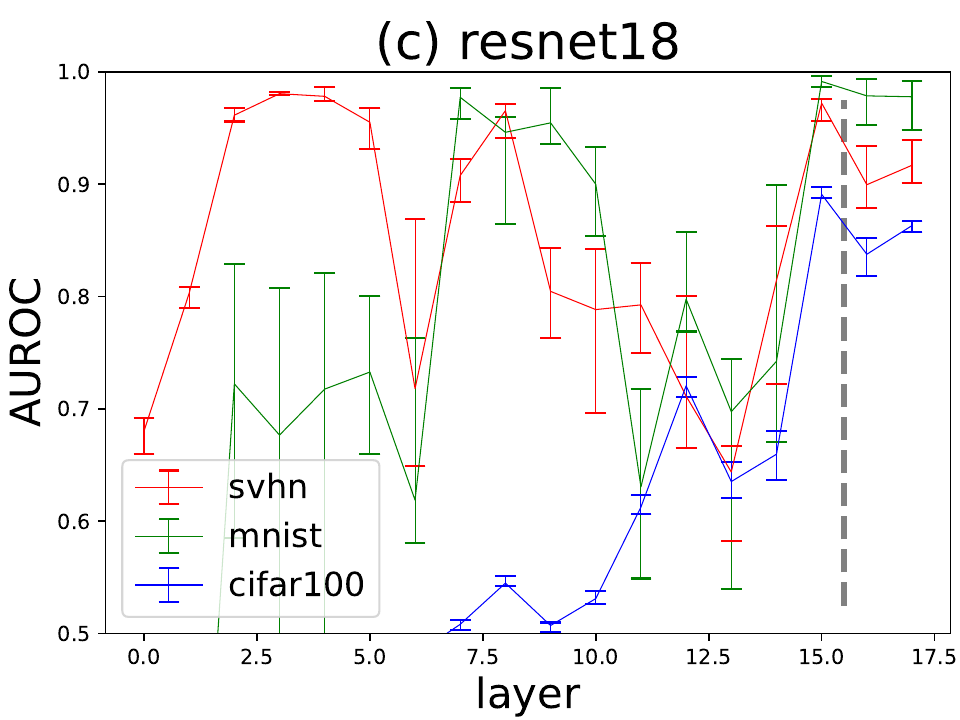}&
    \includegraphics[width=0.22\hsize, bb=0.000000 0.000000 460.800000 345.600000]{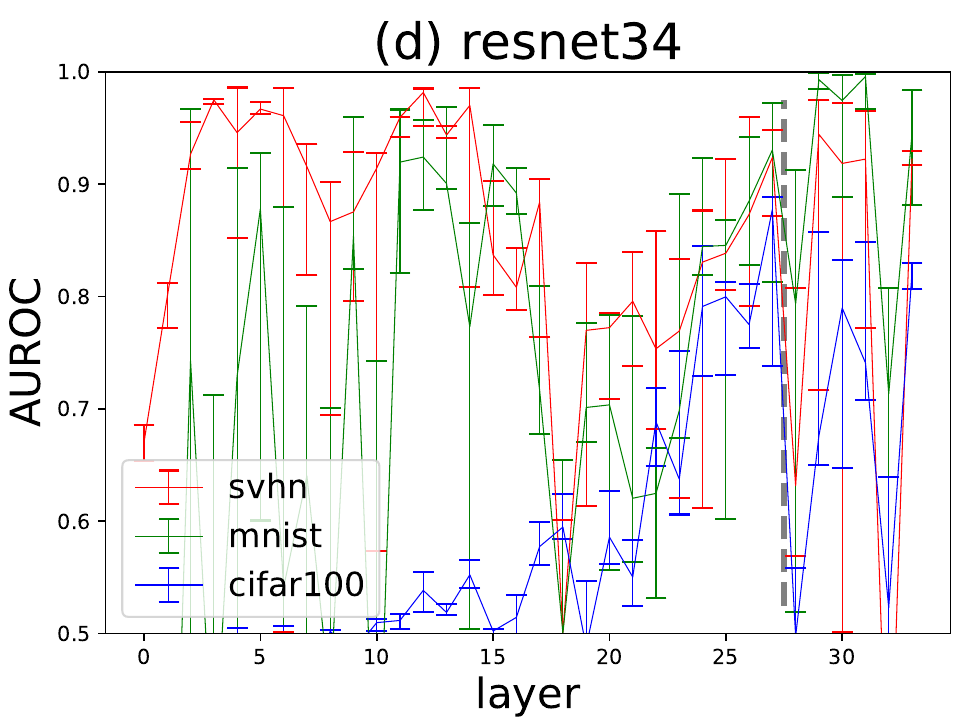}
  \end{tabular}
  \caption{
    Layer dependence of the AUROC detected by the projected norm $||x_{p,\varepsilon}||$
    for the (a) VGG-13, (b) VGG-16, (c) ResNet-18, and (d) ResNet-34 models.
    Different line colors represent the different OOD datasets evaluated.
    The dashed line indicates the transition layer.
    The projection-based discrimination between ID and OOD samples becomes clear just at the transition layer.
    See Sec. \ref{sec:performance} in the main text for more details.
  }
  \label{fig:app_auroc-projection}
\end{figure}

\section{Treatment of convolutional layer}
\label{sec:app_conv}
In this Appendix section, we explain how the matrix representing the convolutional layer is prepared.
Basically, the convolutional layer linearly transforms the input vector with $\mathbb{R}^{H\times P_y\times P_x}$ into the output vector $\mathbb{R}^{H'\times P_y'\times P_x'}$.
Here, $H, P_y$, and $P_x$ are respectively the number of channels, the number of vertical pixels, and the number of horizontal pixels of the input feature map,
while  $H', P_y'$, and $P_x'$ are respectively the number of channels, vertical pixels, and horizontal pixels of the output feature map.
When we neglect the bias term, this tensor-type transformation can be represented by a matrix with a huge number of input--output elements, $\mathbb{R}^{(H'P_y'P_x')\times(HP_yP_x)}$, corresponding to the same representation as the fully connected layer.
It is hard to directly treat this matrix, but we can utilize the locality of the convolutional operation \cite{CNN}.
The convolutional layer shares the same weight matrix among pixels, so it would be reasonable just to consider the local input--output linear transformation.
Let $F_y, F_x$ be the vertical and horizontal size of the convolutional kernel (typically both are 3). 
Then, the local linear transformation of the convolutional layer can be represented by a matrix with $\mathbb{R}^{(H')\times (HF_yF_x)}$ elements.
Because $F_x, F_y$ is much smaller than $P_x, P_x', P_y, P_y'$ in typical cases, we can significantly reduce the computational cost of treating the convolutional layer.
In the main text, we analyze this local linear operator for each convolutional layer to compute the stable rank and the projection.

\section{Complete results for different model architectures}
\label{sec:app_full-model}
\begin{figure}[t]
\centering
  \begin{tabular}{cccc}
    \includegraphics[width=0.22\hsize, bb=0.000000 0.000000 460.800000 345.600000]{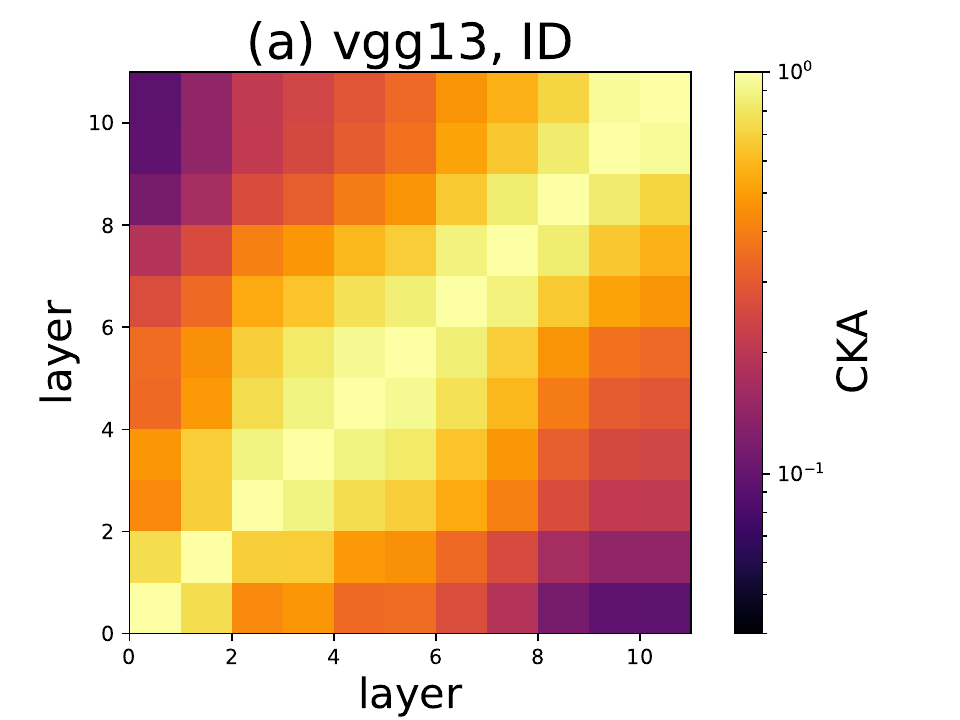}&
    \includegraphics[width=0.22\hsize, bb=0.000000 0.000000 460.800000 345.600000]{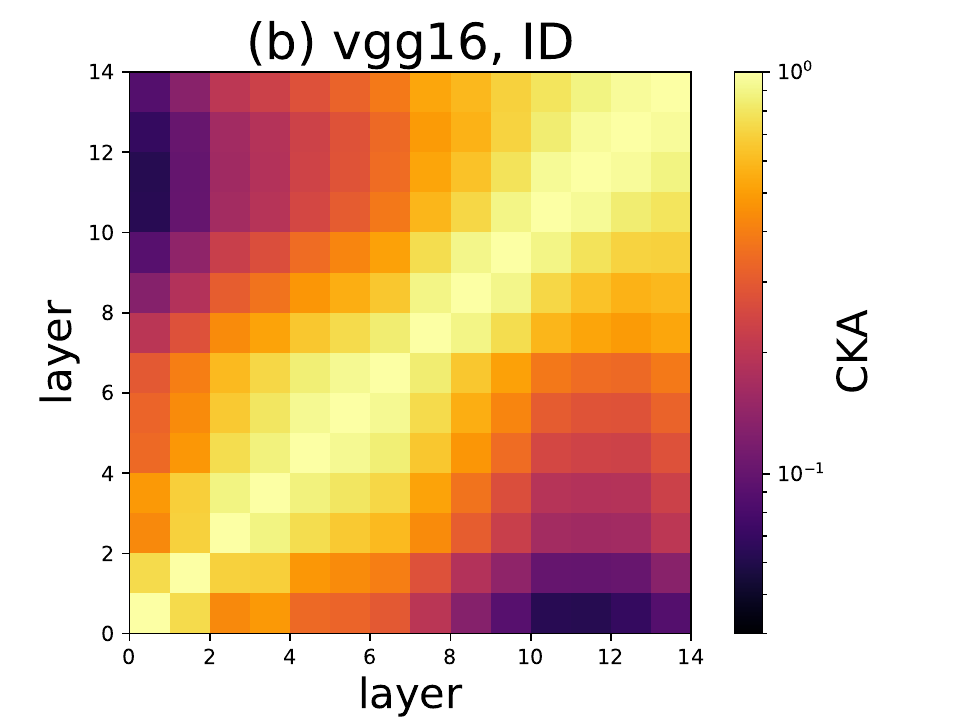}&
    \includegraphics[width=0.22\hsize, bb=0.000000 0.000000 460.800000 345.600000]{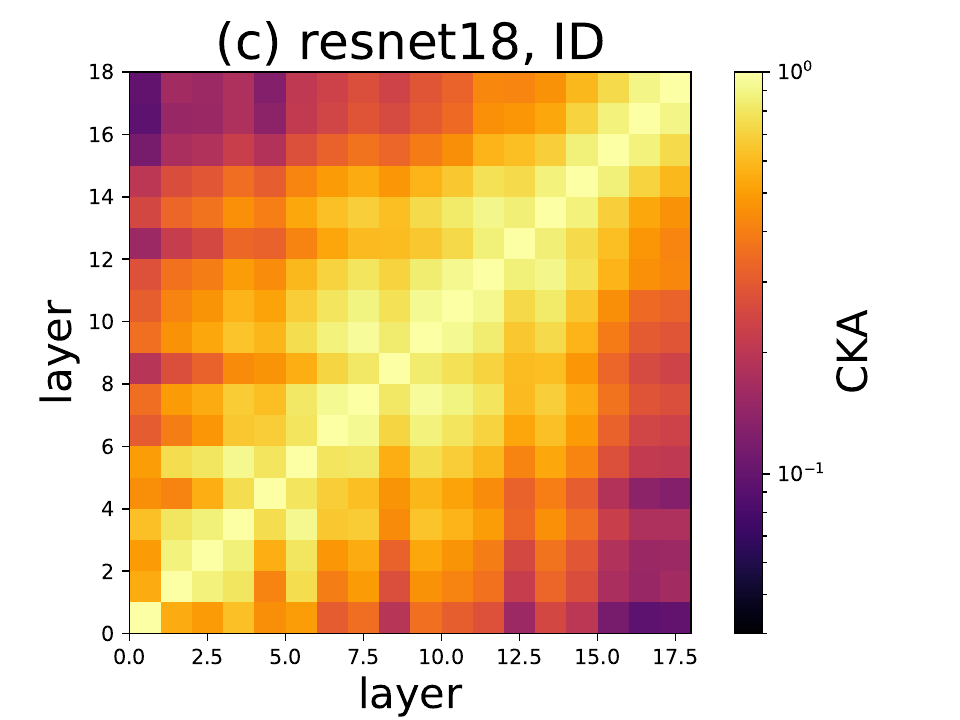}&
    \includegraphics[width=0.22\hsize, bb=0.000000 0.000000 460.800000 345.600000]{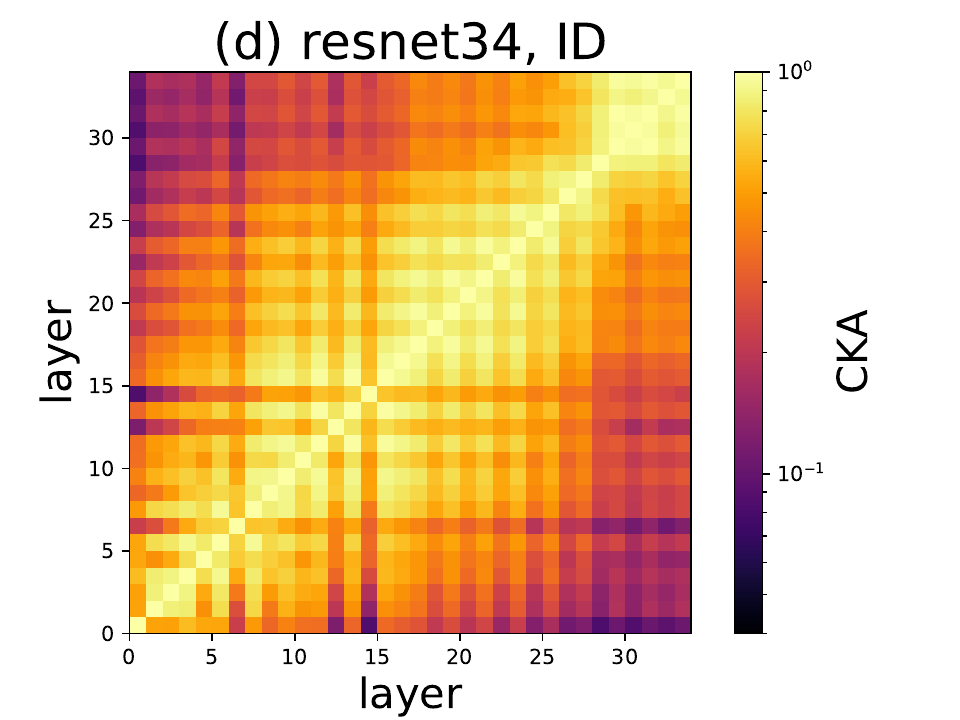}
    \\
    \includegraphics[width=0.22\hsize, bb=0.000000 0.000000 460.800000 345.600000]{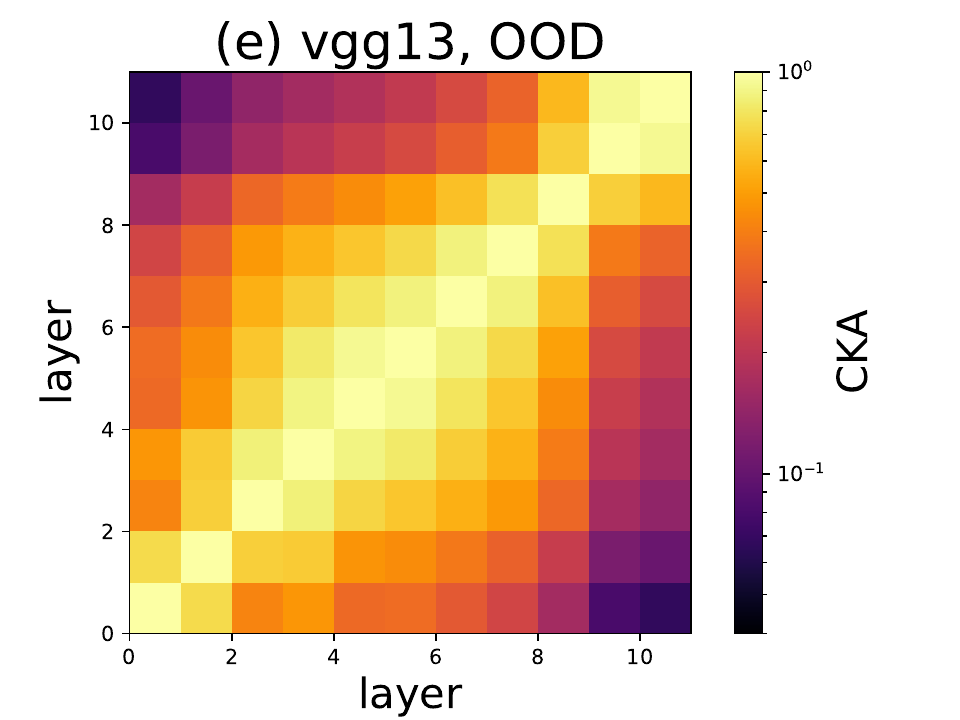}&
    \includegraphics[width=0.22\hsize, bb=0.000000 0.000000 460.800000 345.600000]{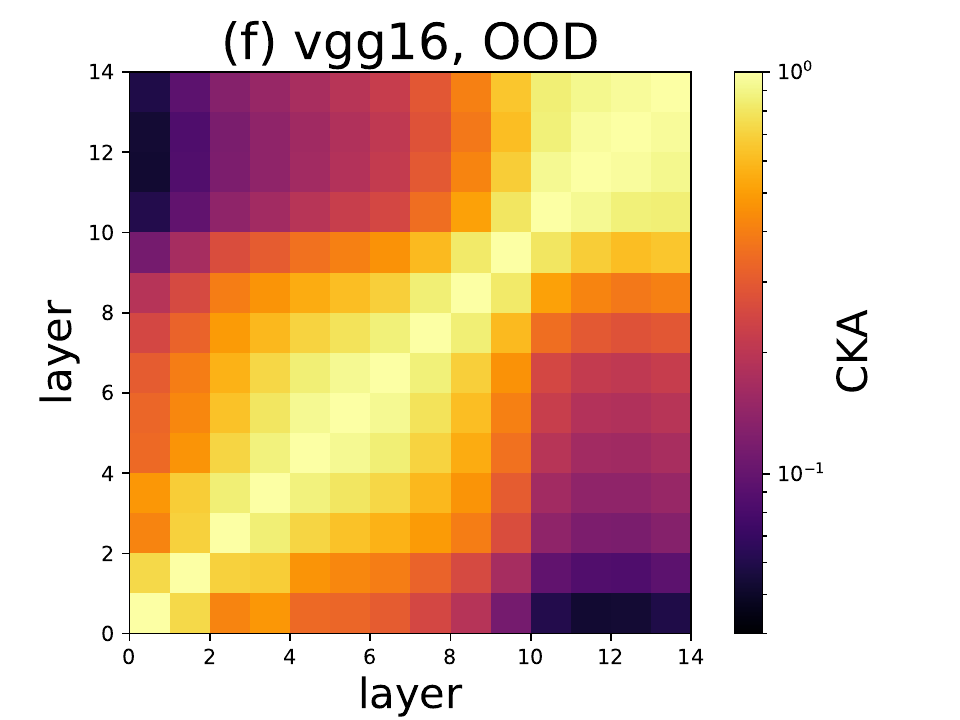}&
    \includegraphics[width=0.22\hsize, bb=0.000000 0.000000 460.800000 345.600000]{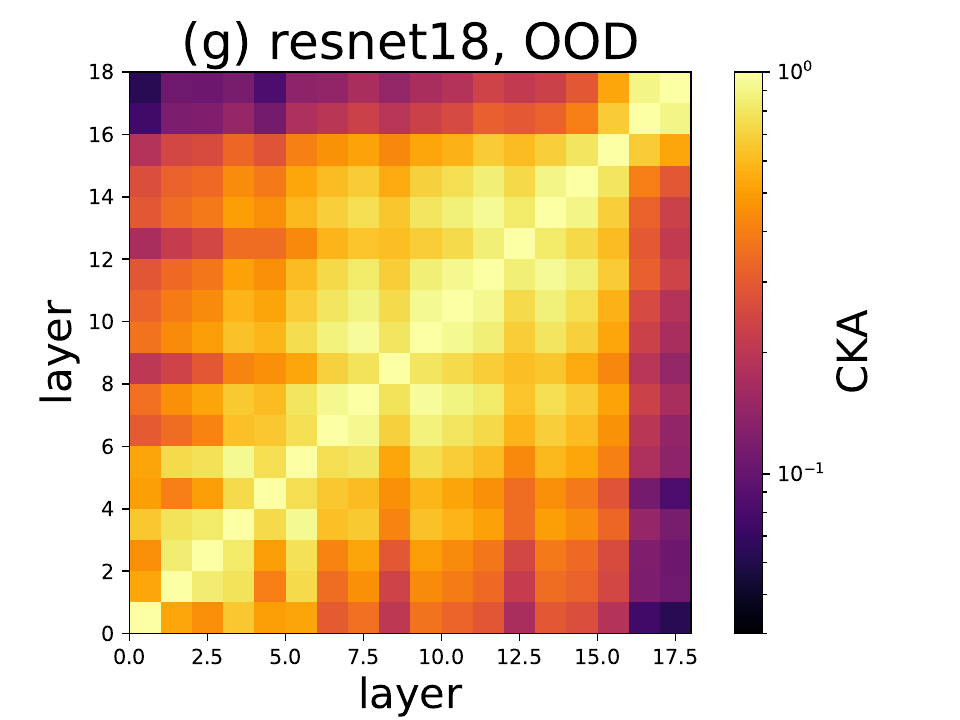}&
    \includegraphics[width=0.22\hsize, bb=0.000000 0.000000 460.800000 345.600000]{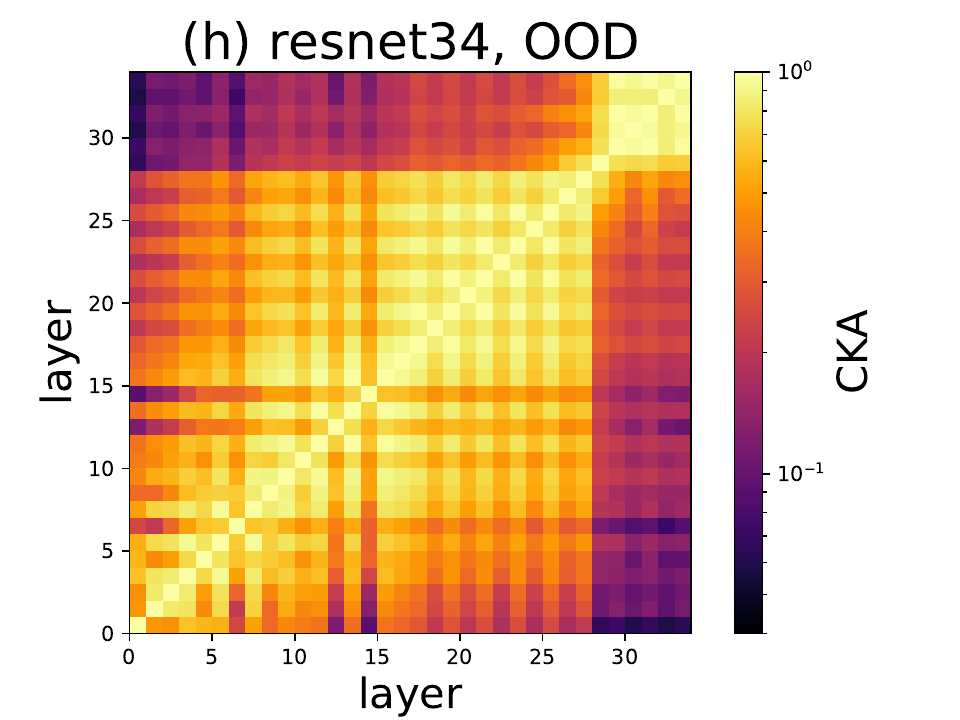}
  \end{tabular}
  \caption{
    CKA of features at various layers
    for the (a,e) VGG-13, (b,f) VGG-16, (c,g) ResNet-18, and (d,h) ResNet-34 models.
    The upper four figures (a--d) show CKA of ID (CIFAR-10) samples,
    while the lower figures (e--h) show CKA of OOD (CIFAR-100) samples.
    In each figure, the horizontal and vertical axes represent layers, and the color bar represents CKA.
    The block-like saturations appear both for ID and OOD samples around the transition layer.
    See Sec. \ref{sec:cka} in the main text for more details.
  }
  \label{fig:app_cka-layers}
\end{figure}

\begin{figure}[t]
\centering
  \begin{tabular}{cccc}
    \includegraphics[width=0.22\hsize, bb=0.000000 0.000000 460.800000 345.600000]{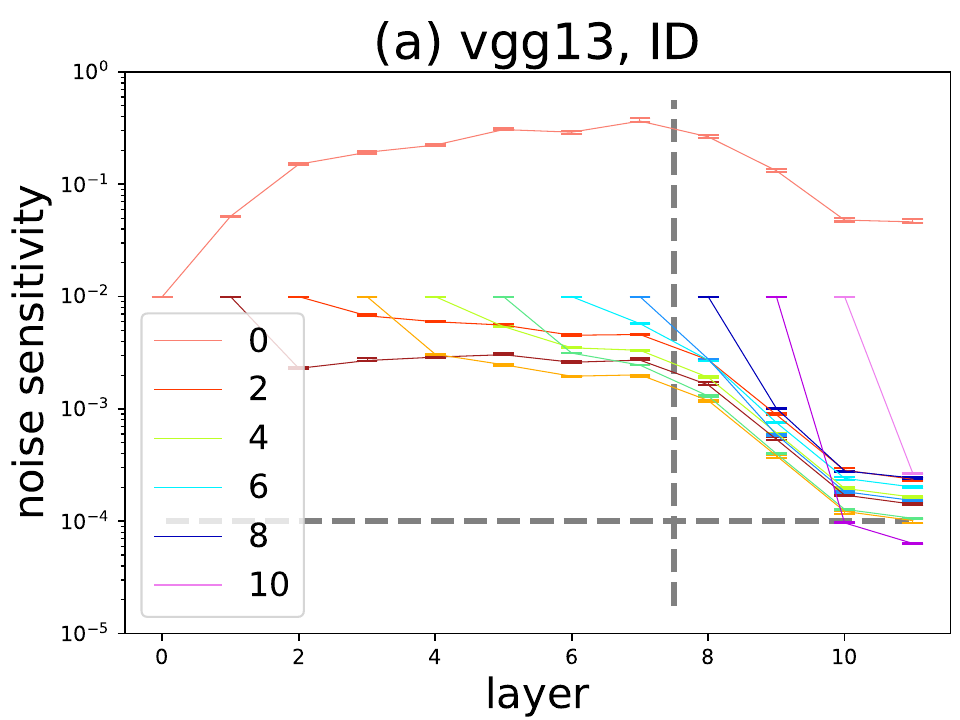}&
    \includegraphics[width=0.22\hsize, bb=0.000000 0.000000 460.800000 345.600000]{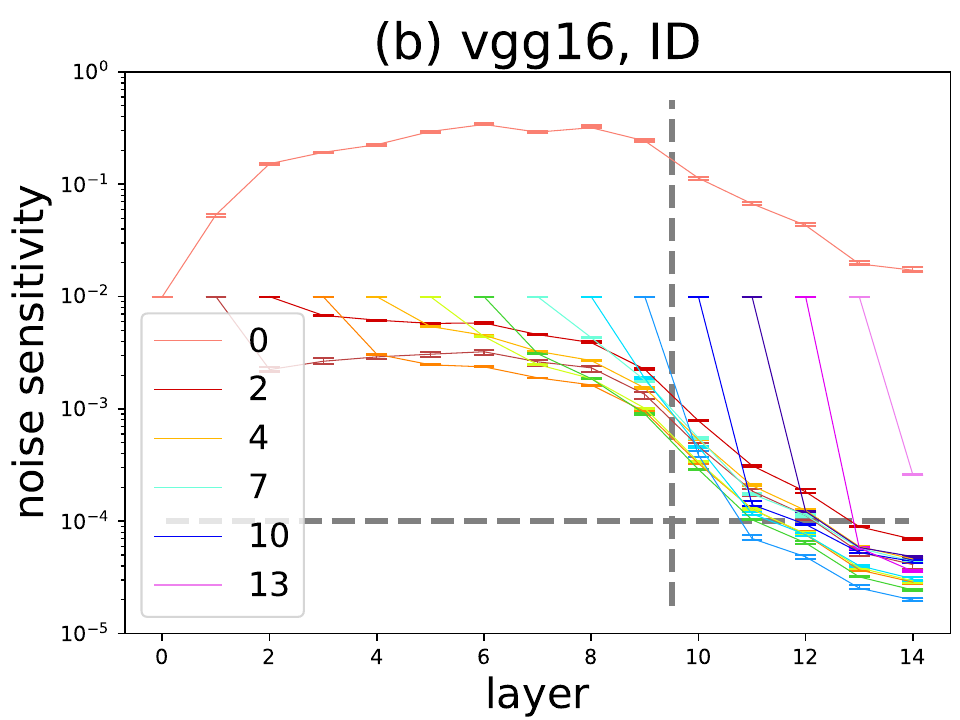}&
    \includegraphics[width=0.22\hsize, bb=0.000000 0.000000 460.800000 345.600000]{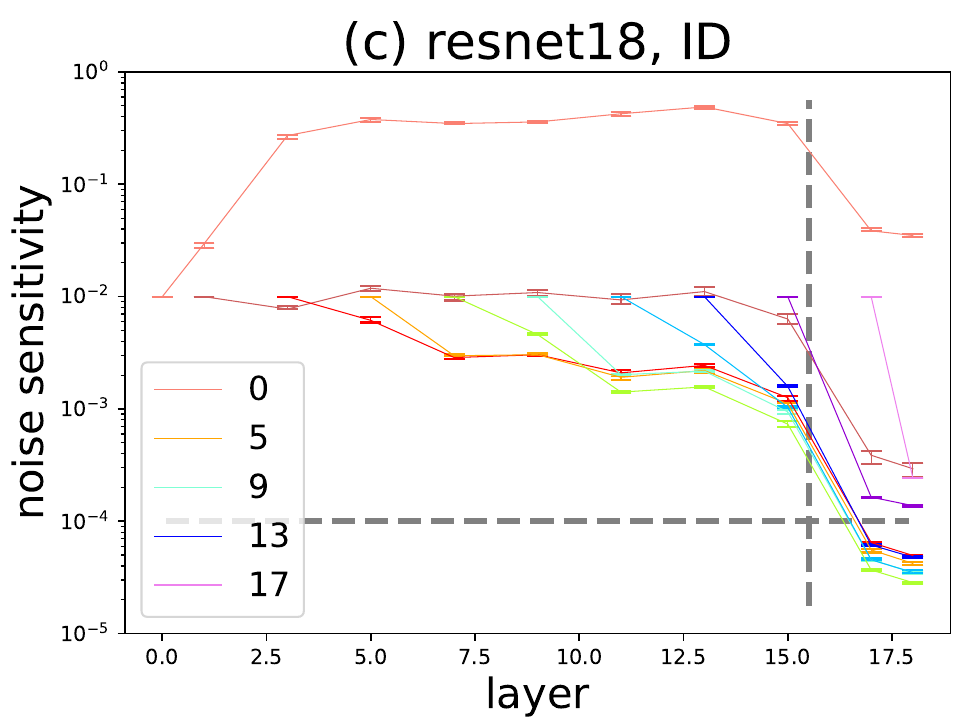}&
    \includegraphics[width=0.22\hsize, bb=0.000000 0.000000 460.800000 345.600000]{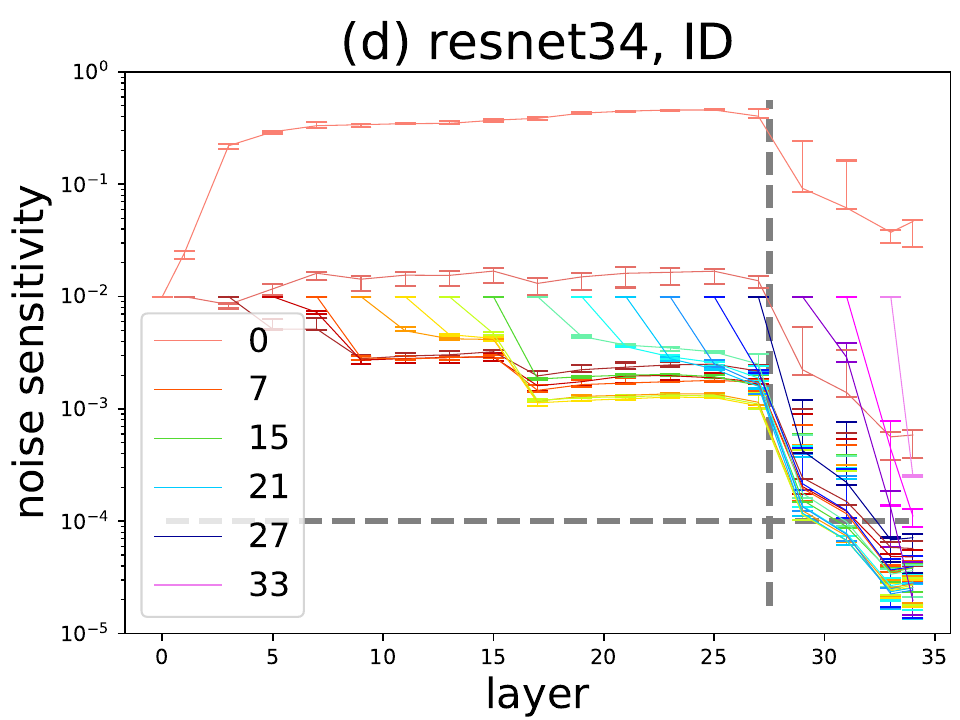}
    \\
    \includegraphics[width=0.22\hsize, bb=0.000000 0.000000 460.800000 345.600000]{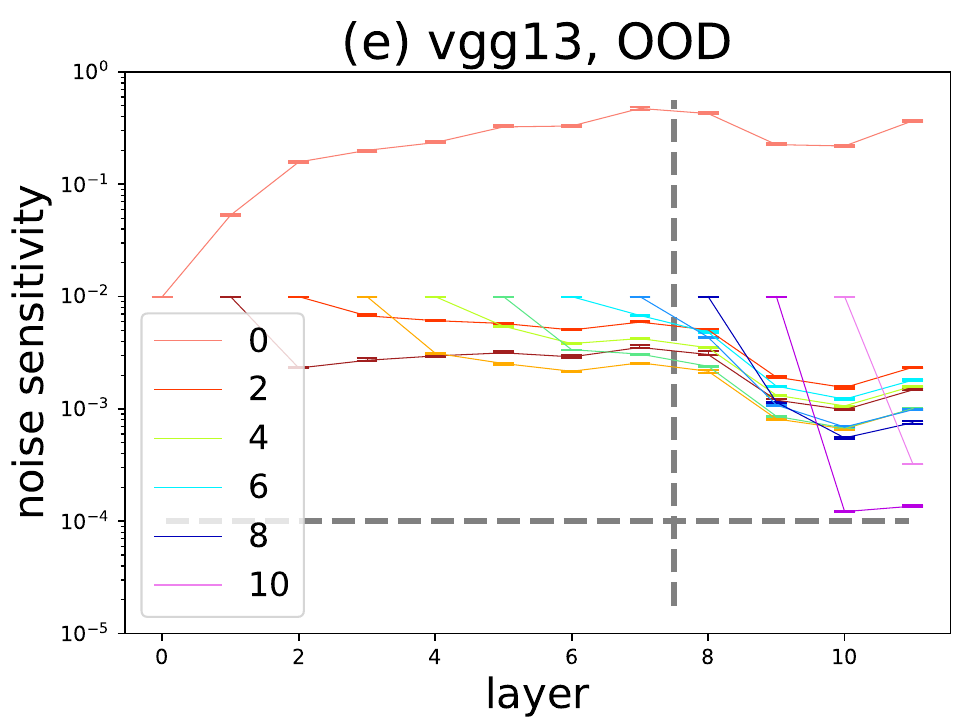}&
    \includegraphics[width=0.22\hsize, bb=0.000000 0.000000 460.800000 345.600000]{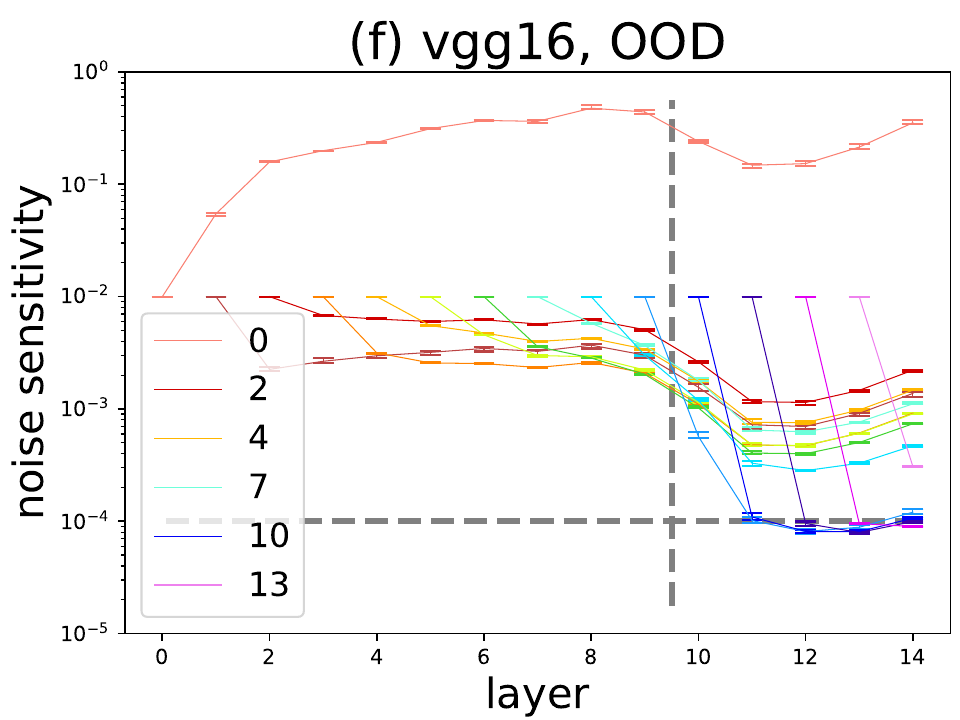}&
    \includegraphics[width=0.22\hsize, bb=0.000000 0.000000 460.800000 345.600000]{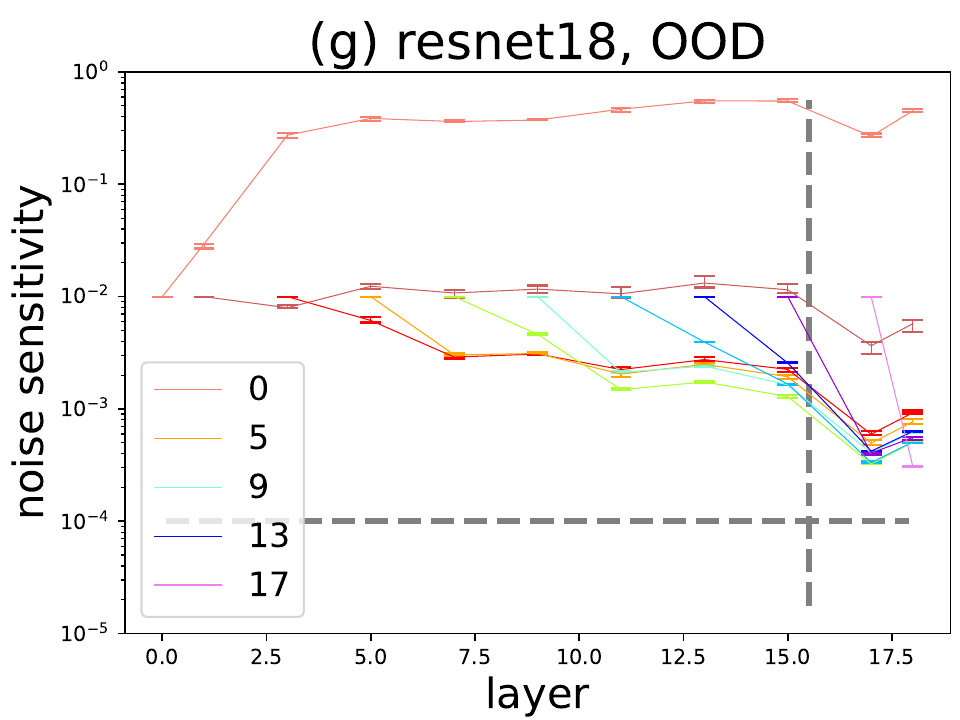}&
    \includegraphics[width=0.22\hsize, bb=0.000000 0.000000 460.800000 345.600000]{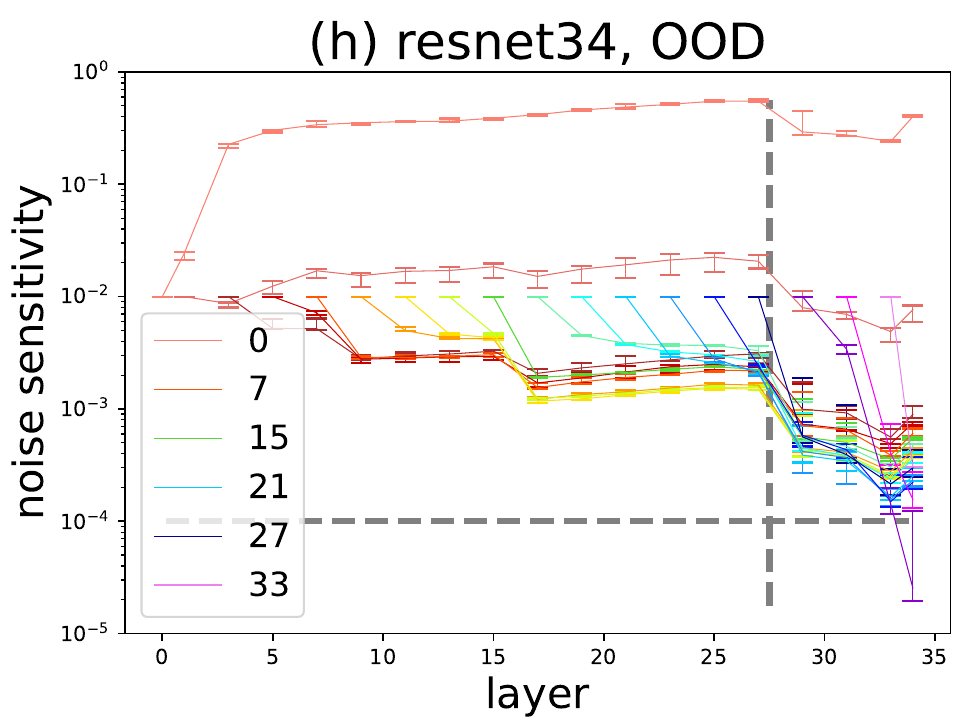}
  \end{tabular}
  \caption{
    Noise sensitivity
    for the (a,e) VGG-13, (b,f) VGG-16, (c,g) ResNet-18, and (d,h) ResNet-34 models.
    The upper (a--d) and lower (e--g) figures show the noise sensitivities
    of ID (CIFAR-10) samples and OOD (CIFAR-100) samples, respectively.
    In each figure, the horizontal axis is the layer
    and the vertical axis is the corresponding noise sensitivity.
    The different colors represent the different input layers where noise is injected.
    The dashed vertical line indicates the transition layer.
    The horizontal line indicates the difference between ID and OOD samples.
    The OOD samples are more sensitive to noise injection compared with ID samples.
    See Sec. \ref{sec:sensitivity} in the main text for more details.
  }
  \label{fig:app_sensitivity}
\end{figure}

\begin{figure}[t]
\centering
  \begin{tabular}{cccc}
    \includegraphics[width=0.22\hsize, bb=0.000000 0.000000 460.800000 345.600000]{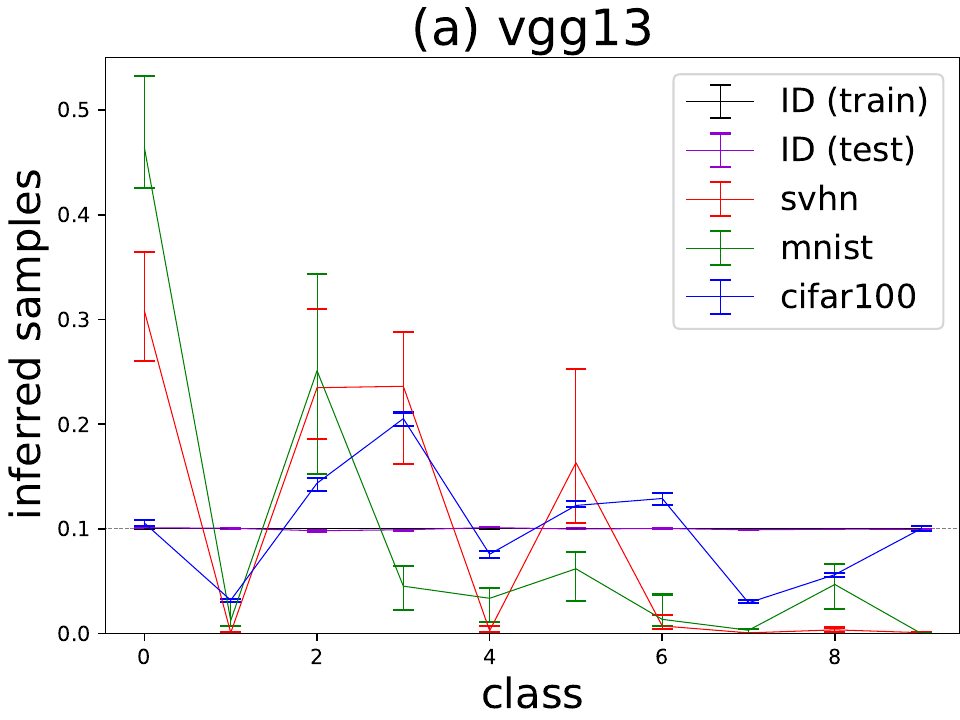}&
    \includegraphics[width=0.22\hsize, bb=0.000000 0.000000 460.800000 345.600000]{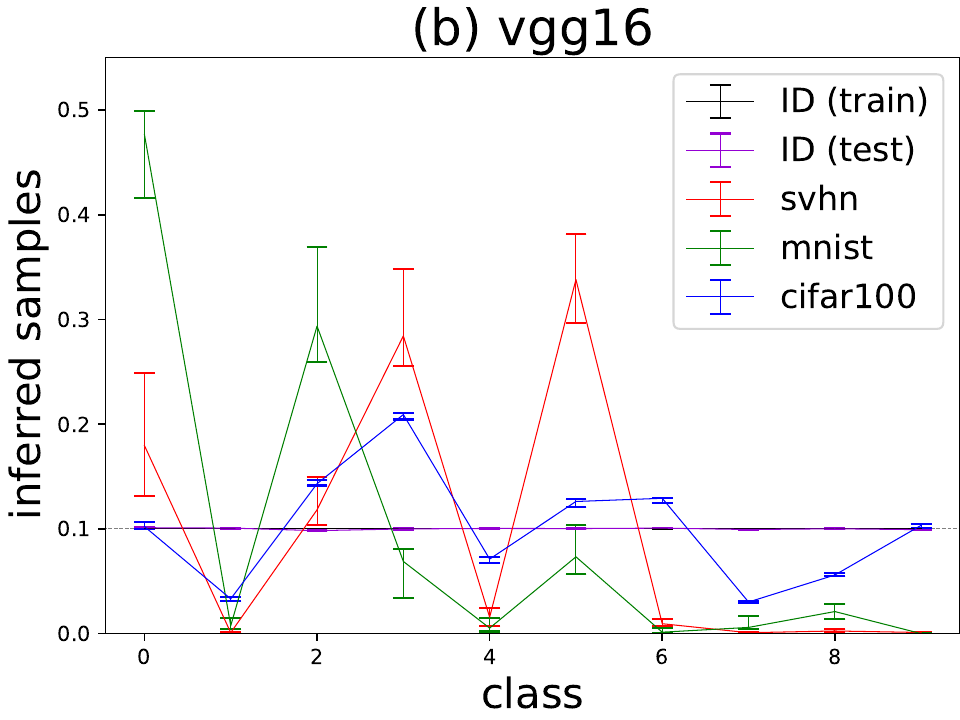}&
    \includegraphics[width=0.22\hsize, bb=0.000000 0.000000 460.800000 345.600000]{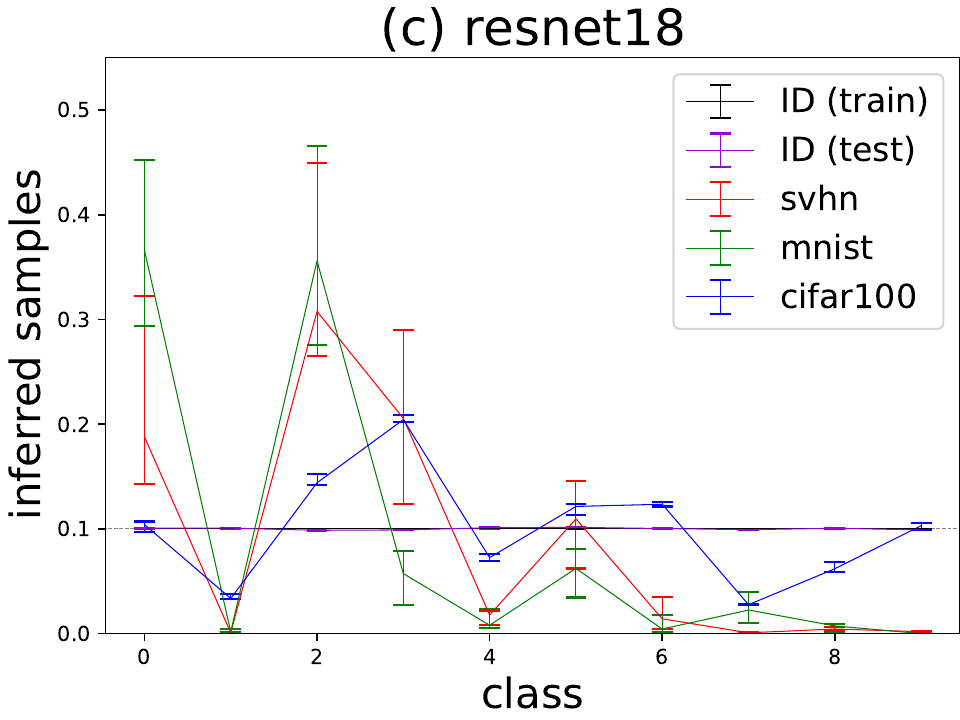}&
    \includegraphics[width=0.22\hsize, bb=0.000000 0.000000 460.800000 345.600000]{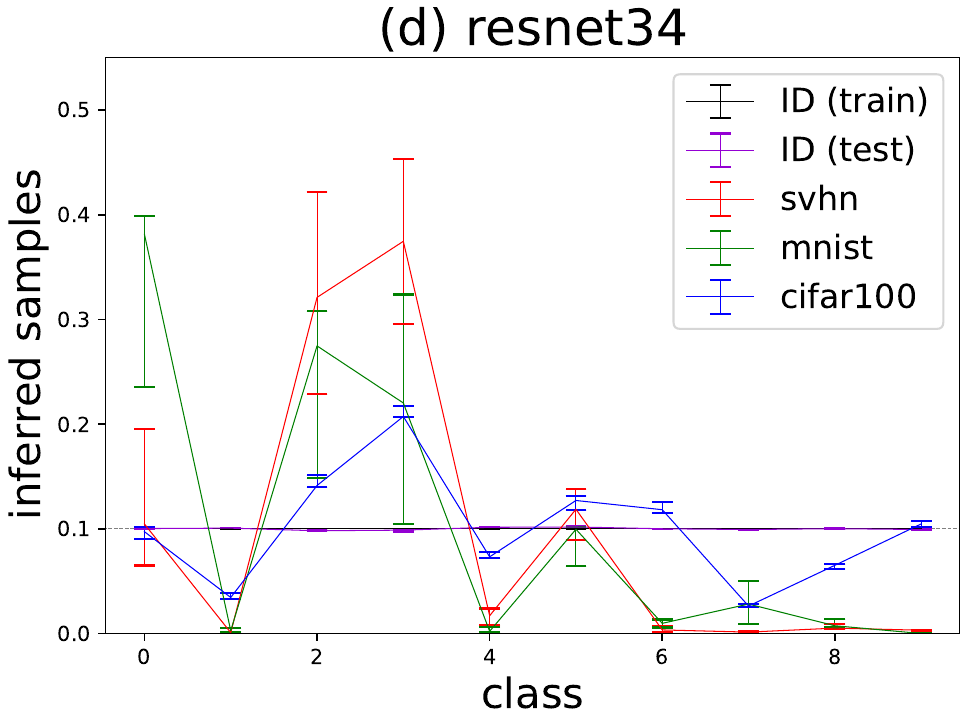}
  \end{tabular}
  \caption{
    Rates of inferred samples to each class
    for the (a) VGG-13, (b) VGG-16, (c) ResNet-18, and (d) ResNet-34 models.
    Different line colors represent the different datasets.
    The inference of OOD samples is highly imbalanced compared with that of ID (CIFAR-10) samples.
    See Sec. \ref{sec:inference} in the main text for more details.
  }
  \label{fig:app_inference}
\end{figure}

In this Appendix section, we provide complete results given in the main text for different model architectures.
Although the individual results may be rather noisy,
we can observe the same behaviors as discussed in the main text.
This consistent correlation verifies the picture in the main text.
The detailed discussion is only provided in the main text to avoid duplication, so it is recommended to see results while corresponding this Appendix and the main text in Sec. \ref{sec:results} as follows: 
Fig. \ref{fig:app_stableranks} and Fig. \ref{fig:stableranks}, 
Fig. \ref{fig:app_auroc-cov} and Fig. \ref{fig:auroc} (a),
Fig. \ref{fig:app_auroc-projection} and Fig. \ref{fig:auroc} (b),
Fig. \ref{fig:app_cka-layers} and Fig. \ref{fig:cka-layers},
Fig. \ref{fig:app_sensitivity} and Fig. \ref{fig:sensitivity},
Fig. \ref{fig:app_inference} and Fig. \ref{fig:inference}.

\section{Experimental setup}
\label{sec:app_related_detail}

We train some DNNs using momentum ($m=0.9$) stochastic gradient descent (SGD) with weight decay coefficient $\lambda_{\mathrm{wdecay}}=5\times10^{-4}$.
The loss function is the standard cross-entropy loss.
The learning rate is decreased from 0.1 to 0 following a cosine scheduler with 128 images per minibatch.
We perform momentum SGD up to 1200 epochs by applying the following data augmentation series to the ID samples:
random crop, random horizontal flip, normalization, and random erasing.
We optimize the various network architectures of VGG-13, VGG-16 \cite{VGG}, ResNet-18, and ResNet-34 \cite{ResNet} with batch normalization \cite{BatchNorm} for 10 random initializations in the same manner given above.
We show the results using quantiles; the median represents the typical value, while the 0.25 and 0.75 quantiles represent the statistical error.
In the following analysis, we treat the CIFAR-10 \cite{CIFAR} dataset as ID, and CIFAR-100 \cite{CIFAR}, SVHN \cite{SVHN}, MNIST \cite{MNIST} as OOD.
In this setting, CIFAR-100 is OOD but close to ID (CIFAR-10), while the others are relatively far from ID.

We investigate the properties of ID and OOD samples from various aspects as stated in Sec. \ref{sec:relatedworks} in the main text.
Another major quantity is the area under the receiver-operating characteristic curve (AUROC),
representing the deviation of OOD from ID as described by Ref. \cite{OOD-baseline}.
This quantity is normalized between zero and one, and the higher value means ID and OOD are well distinguished.
Therefore, this can be used to quantify performance of OOD detection.
We compare three methods quantitatively:
the probability-based method \cite{OOD-baseline} that regards samples with large/small maximal softmax probabilities as ID/OOD,
the feature-based method \cite{OOD-Mahalanobis} using Mahalanobis distances for OOD detection,
and the projection-based method utilizing the alignment of features and weights specified in Sec. \ref{sec:related_projection} in the main text and Appendix \ref{sec:app_conv}, \ref{sec:app_compress}.
For the projection-based method, we evaluate NuSA \cite{OOD-NuSA} and the modified one, where singular vectors with smaller singular values are removed.
See \ref{sec:app_compress} for more details.
In computing the AUROC, we randomly select the same number of samples ($=10000$ samples) from the ID and OOD test datasets.

\section{Comprehensive analysis of OOD detection performance}
\label{sec:app_comprehensive}
For the additional verification of the effectiveness of our modification to the projection-based detection, 
we summarize in Tab. \ref{tab:comparison} OOD detection performance detected by the ratio of norm, $||x_{p,\varepsilon}^{(l)}|| / ||x^{(l)}||$ at the penultimate fully connected layer.
Our method is quite similar to NuSA in this case \cite{OOD-NuSA}.
However, NuSA uses full projections ($\varepsilon=0$), while we eliminate irrelevant singular vectors from the projection matrix by focusing on the corresponding singular values (typically $\varepsilon\sim 10^{-2}$).
See Appendix \ref{sec:app_compress} for more details.
Comparing these two projection-based methods, this minor modification significantly improves not only detection performance but also its stability.
Also, our modified projection-based method tends to outperform the probability-based method \cite{OOD-baseline}.
Compared with the feature-based method, our projection-based method exhibits similar detection performance in cases with close-to-ID OOD samples, where CIFAR-10 and CIFAR-100 are treated as ID and OOD datasets.
For much easier OOD samples, the projection-based method tends to outperform the feature-based method.
In summary, our modification of the projection-based method improves the OOD detection performance well enough to be comparable or better than other methods.

\begin{table*}[t]
  \caption{
    Summary of OOD detection performance (AUROC) in various situations.
    All results were reproduced in our experiments.
    Values inside the parentheses () represent the error, that is, the average of deviations from the median to the 0.25 and 0.75 quantiles.
    For example, $0.913(8)$ means $0.913\pm 0.008$, $0.892(16)$ means $0.892\pm 0.016$.
  }
  \begin{center}
    \begin{tabular}{cccccc}
      \hline \hline
      \multirow{2}{*}{
        \begin{tabular}{c}
          \textbf{ID dataset}\\
          \textbf{(Model)}
        \end{tabular}
      } 
      & \multirow{2}{*}{\textbf{Detection score}}
      & \multicolumn{4}{c}{\textbf{OOD dataset}} \\
      & & CIFAR-10 & CIFAR-100 & SVHN & MNIST \\
      \hline \hline
      \multirow{4}{*}{
        \begin{tabular}{c}
          CIFAR-10\\
          (VGG-13)
        \end{tabular}
      }
      & Probability \cite{OOD-baseline}&
      (ID) &
      0.886(1) &
      0.944(5) &
      0.927(7) \\
      & Feature \cite{OOD-Mahalanobis}&
      (ID) &
      0.898(1) &
      0.948(4) &
      0.936(5) \\
      & Projection \cite{OOD-NuSA}&
      (ID) &
      0.70(6) &
      0.62(7) &
      0.47(16) \\
      & Projection (ours) &
      (ID) &
      0.901(1) &
      0.960(4) &
      0.978(4) \\
      \hline
      \multirow{4}{*}{
        \begin{tabular}{c}
          CIFAR-10\\
          (ResNet-18)
        \end{tabular}
      }
      & Probability \cite{OOD-baseline}&
      (ID) &
      0.905(1) &
      0.956(6) & 
      0.953(9) \\
      & Feature \cite{OOD-Mahalanobis}&
      (ID) &
      0.919(1)  &
      0.958(5) &
      0.956(7) \\
      & Projection \cite{OOD-NuSA}&
      (ID) &
      0.84(5) &
      0.88(6) &
      0.85(14) \\
      & Projection (ours) &
      (ID) &
      0.924(1) &
      0.969(3) &
      0.982(4) \\
      \hline \hline
      \multirow{4}{*}{
        \begin{tabular}{c}
          SVHN\\
          (VGG-13)
        \end{tabular}
      }
      & Probability \cite{OOD-baseline}&
      0.959(1) &
      0.954(1) &
      (ID) &
      0.949(9) \\
      & Feature \cite{OOD-Mahalanobis}&
      0.965(1) &
      0.961(2) &
      (ID) &
      0.921(9) \\
      & Projection \cite{OOD-NuSA}&
      0.86(7) &
      0.85(6) &
      (ID) &
      0.71(20) \\
      & Projection (ours) &
      0.977(2) &
      0.973(2) &
      (ID) &
      0.965(5) \\
      \hline
      \multirow{4}{*}{
        \begin{tabular}{c}
          SVHN\\
          (ResNet-18)
        \end{tabular}
      }
      & Probability \cite{OOD-baseline}&
      0.950(2) &
      0.944(3) &
      (ID) &
      0.983(3) \\
      & Feature \cite{OOD-Mahalanobis}&
      0.970(2) &
      0.967(3) &
      (ID) &
      0.918(11) \\
      & Projection \cite{OOD-NuSA}&
      0.90(3) &
      0.90(3) &
      (ID) &
      0.43(28) \\
      & Projection (ours) &
      0.970(3) &
      0.965(4) &
      (ID) &
      0.996(2) \\
      \hline \hline
    \end{tabular}
    \label{tab:app_comparison}
  \end{center}
\end{table*}

We note here that the performance of the feature-based detection in Tab. \ref{tab:comparison} is given by the single-layer detection at the penultimate layer,
not by the ensemble of multiple layers.
The layer ensemble method by Ref. \cite{OOD-Mahalanobis} uplifts the detection accuracy of easier OOD samples.
In contrast, it has a negative effect on the harder OOD samples.
Actually, we checked that the AUROC value to detect CIFAR-100 OOD dataset using the ensemble method adopted by Ref. \cite{OOD-Mahalanobis} is just around 0.86 for models trained by CIFAR-10,
while single-layer detection can achieve AUROC $\sim 0.90$ as shown in Tab. \ref{tab:comparison}.
Also, the layer ensemble requires a lot of memory to save covariances, which is not suitable especially for resource-limited hardware.
More seriously, the ensemble method by Ref. \cite{OOD-Mahalanobis} requires some OOD samples, 
although it would be practically inaccessible in cases where we do not know what kinds of OOD samples are contaminated.
Therefore, if we are interested in close-to-ID OOD detection,
we should not use the layer ensemble, considering its performance, computational cost, and accessibility to OOD samples.
Again, our projection-based method tends to outperform single-layer feature-based methods.

The projection-based method is also superior to the feature-based method in terms of computational cost, especially if the number of classes $K$ is much smaller than the feature dimensions at the OOD detection layer.
The preparation of the tied covariance requires $\mathcal{O}(N_{\mathrm{train}})$ preprocessing,
and the computation of the Mahalanobis distance requires $\mathcal{O}(K)$ evaluations to find the minimum values for classes.
The projection-based method, on the other hand, merely computes the projection to singular vectors,
leading to the reduction of $\mathcal{O}(N_{\mathrm{train}})$ preprocessing and $\mathcal{O}(K)$ evaluations.
Also, the projection-based method is more memory-efficient compared with the feature-based method, even in single-layer detection.
The null space of the low-dimensional covariance makes a large contribution to the feature-based detection as demonstrated by Ref. \cite{OOD-Mahalanobis-PCA}.
In fact, we checked that the best number of dimensions for the feature-based OOD detection is around 200--400 dimensions out of 512 dimensions at the penultimate layer.
Meanwhile, the projection-based method eliminates the null space of the low-dimensional weight,
leaving only the dimension of the number of classes.
These facts make the projection-based method more efficient than the feature-based one, especially in hardware with limited computational resources.



\section{Computation of similarities}
\label{sec:app_similarity}
We provide details of how centered kernel alignment (CKA) is computed.
We drop the index and $\mathcal{D}$ for simplicity.

\begin{figure}[t]
\centering
  \begin{tabular}{cccc}
    \includegraphics[width=0.22\hsize, bb=0.000000 0.000000 460.800000 345.600000]{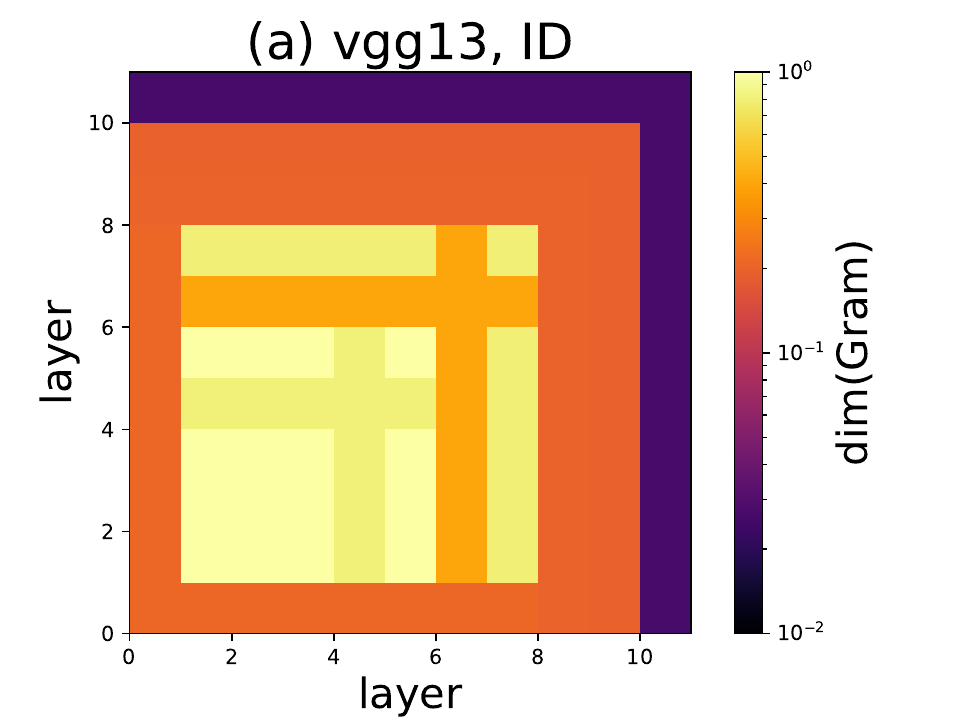}&
    \includegraphics[width=0.22\hsize, bb=0.000000 0.000000 460.800000 345.600000]{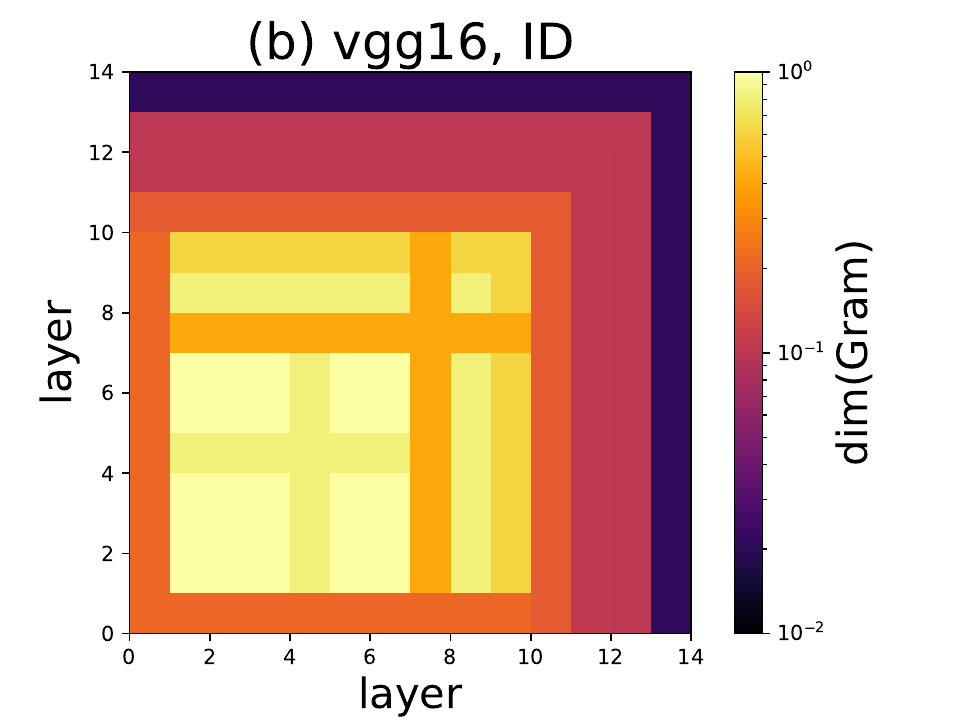}&
    \includegraphics[width=0.22\hsize, bb=0.000000 0.000000 460.800000 345.600000]{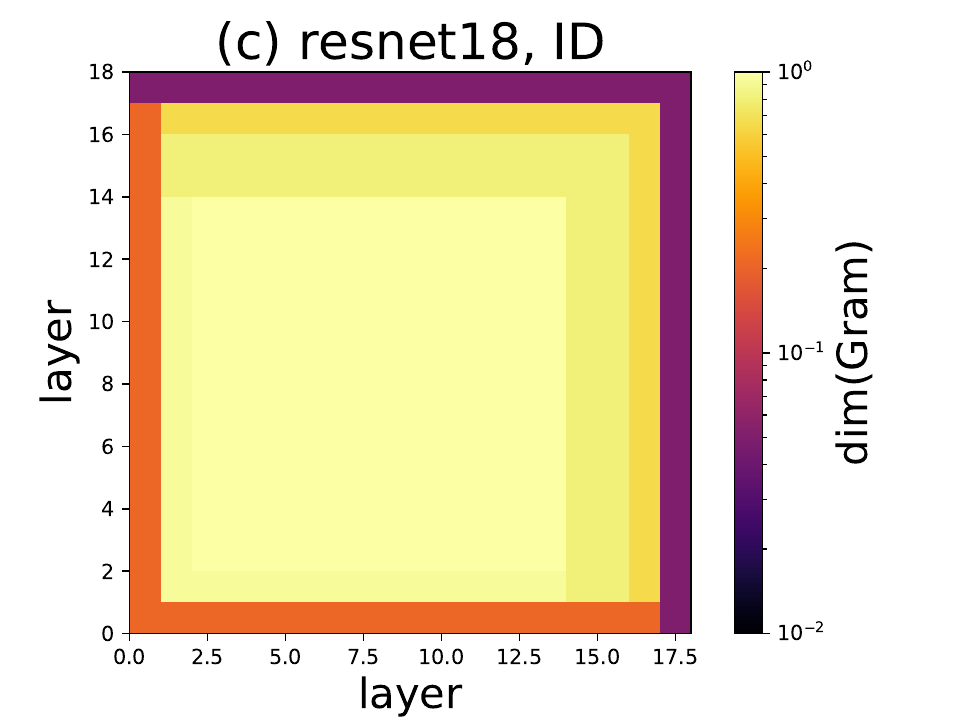}&
    \includegraphics[width=0.22\hsize, bb=0.000000 0.000000 460.800000 345.600000]{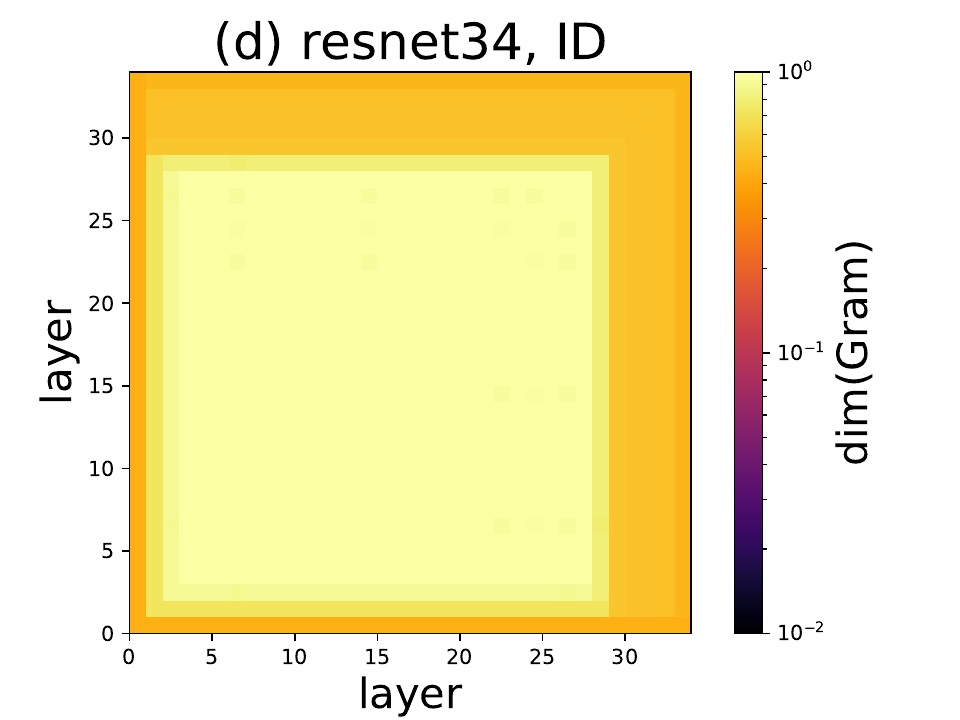}
  \end{tabular}
  \caption{
    Residual numbers of the dimensions of Gram matrices 
    for the (a) VGG-13, (b) VGG-16, (c) ResNet-18, and (d) ResNet-34 models
    with smaller eigenvalues removed by the threshold $\varepsilon_{\mathrm{gram}}$.
  }
  \label{fig:app_dims-gram}
\end{figure}

\begin{figure}[t]
\centering
  \begin{tabular}{cccc}
    \includegraphics[width=0.22\hsize, bb=0.000000 0.000000 460.800000 345.600000]{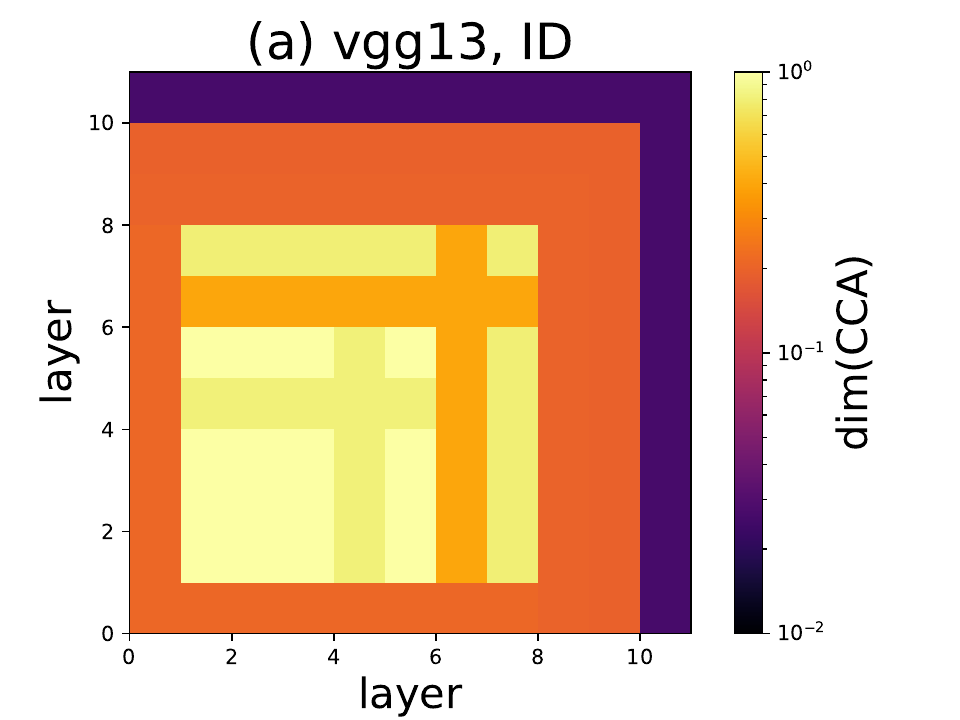}&
    \includegraphics[width=0.22\hsize, bb=0.000000 0.000000 460.800000 345.600000]{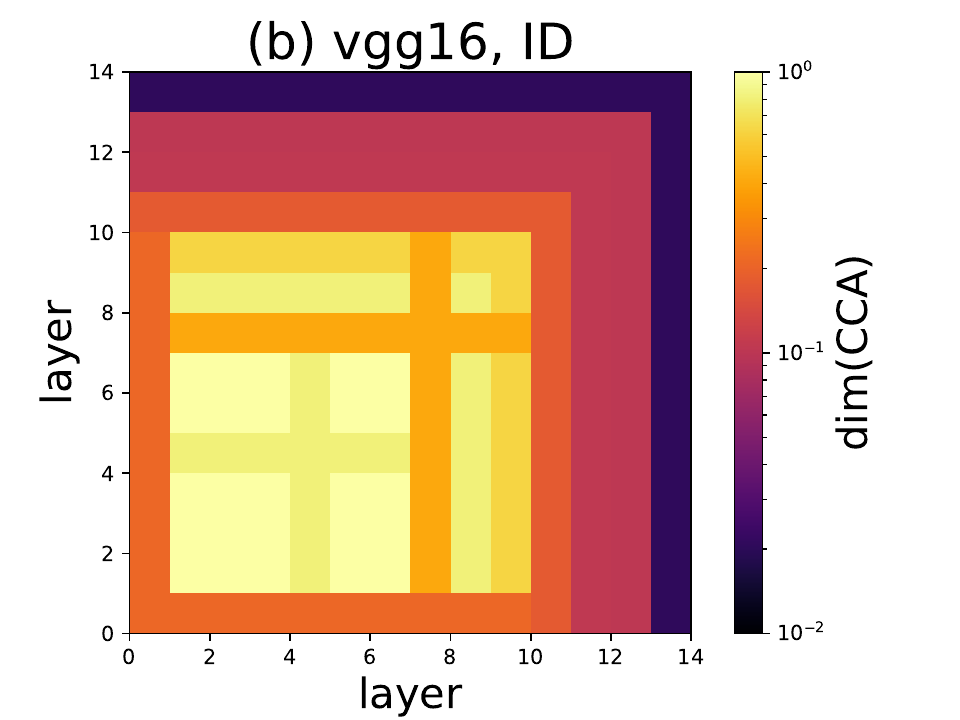}&
    \includegraphics[width=0.22\hsize, bb=0.000000 0.000000 460.800000 345.600000]{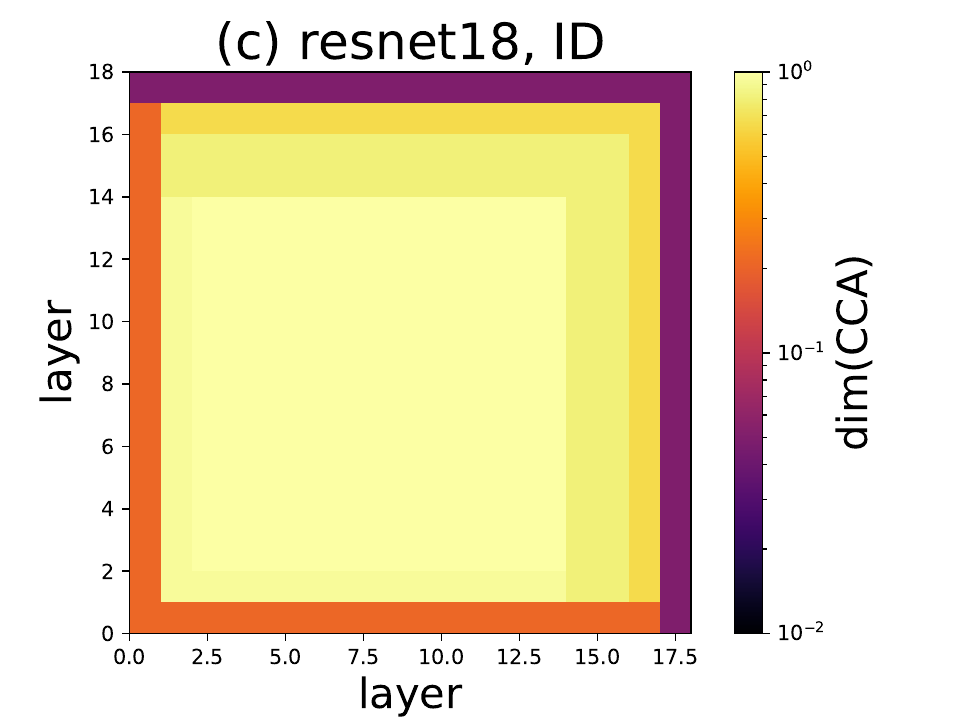}&
    \includegraphics[width=0.22\hsize, bb=0.000000 0.000000 460.800000 345.600000]{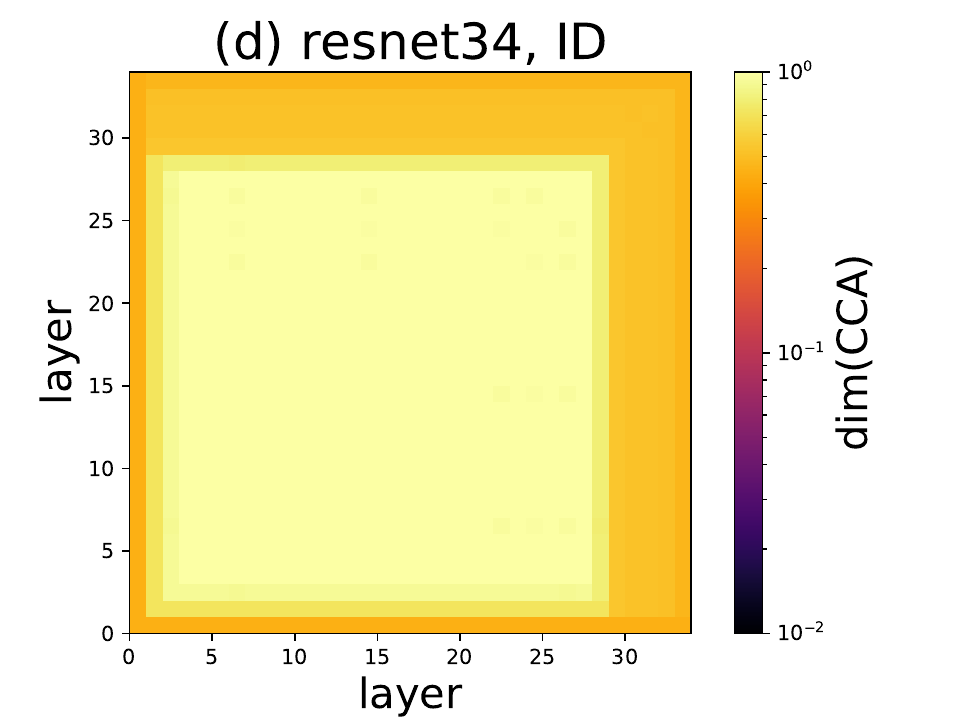}
  \end{tabular}
  \caption{
    Residual numbers of dimensions of CCA matrices
    for the (a) VGG-13, (b) VGG-16, (c) ResNet-18, and (d) ResNet-34 models
    with smaller singular values removed by the threshold $\varepsilon_{\mathrm{sim}}$.
  }
  \label{fig:app_dims-cca}
\end{figure}

\begin{figure}[t]
\centering
  \begin{tabular}{cccc}
    \includegraphics[width=0.22\hsize, bb=0.000000 0.000000 460.800000 345.600000]{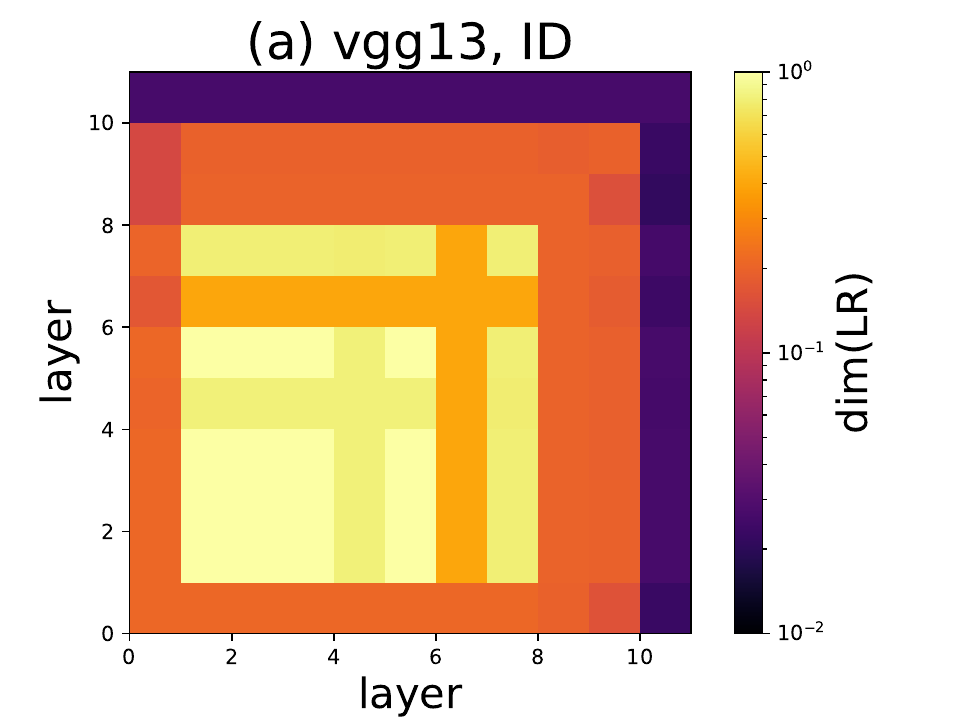}&
    \includegraphics[width=0.22\hsize, bb=0.000000 0.000000 460.800000 345.600000]{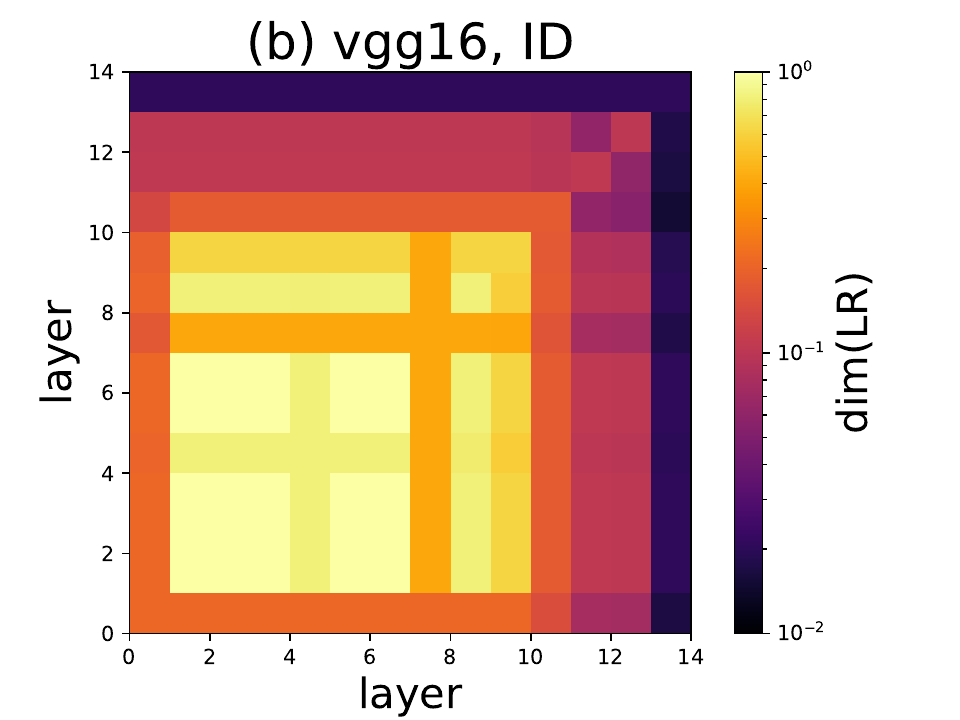}&
    \includegraphics[width=0.22\hsize, bb=0.000000 0.000000 460.800000 345.600000]{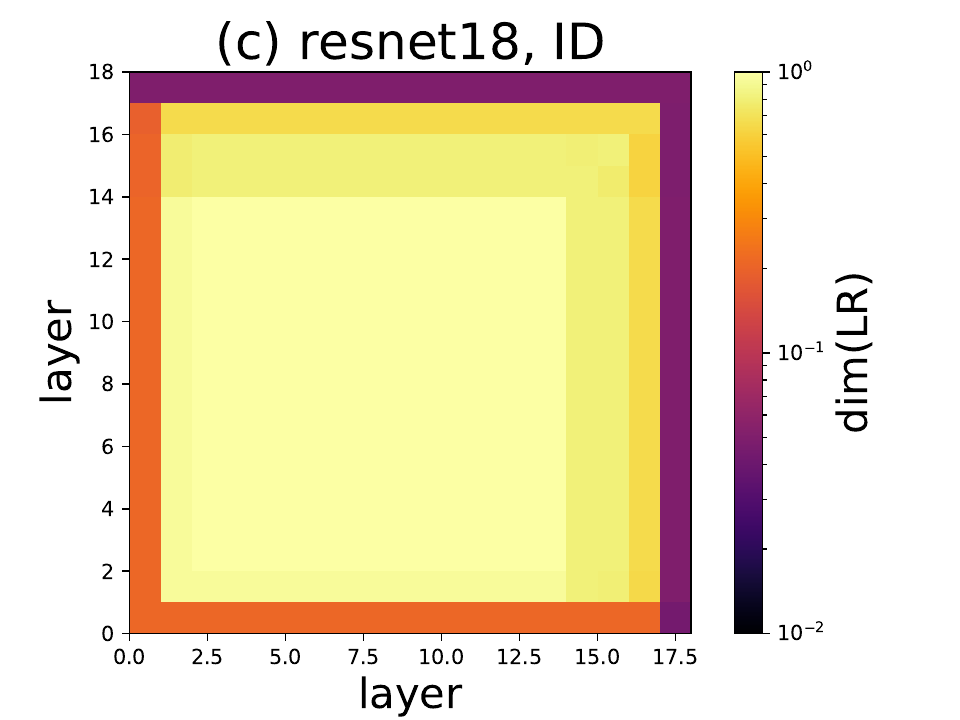}&
    \includegraphics[width=0.22\hsize, bb=0.000000 0.000000 460.800000 345.600000]{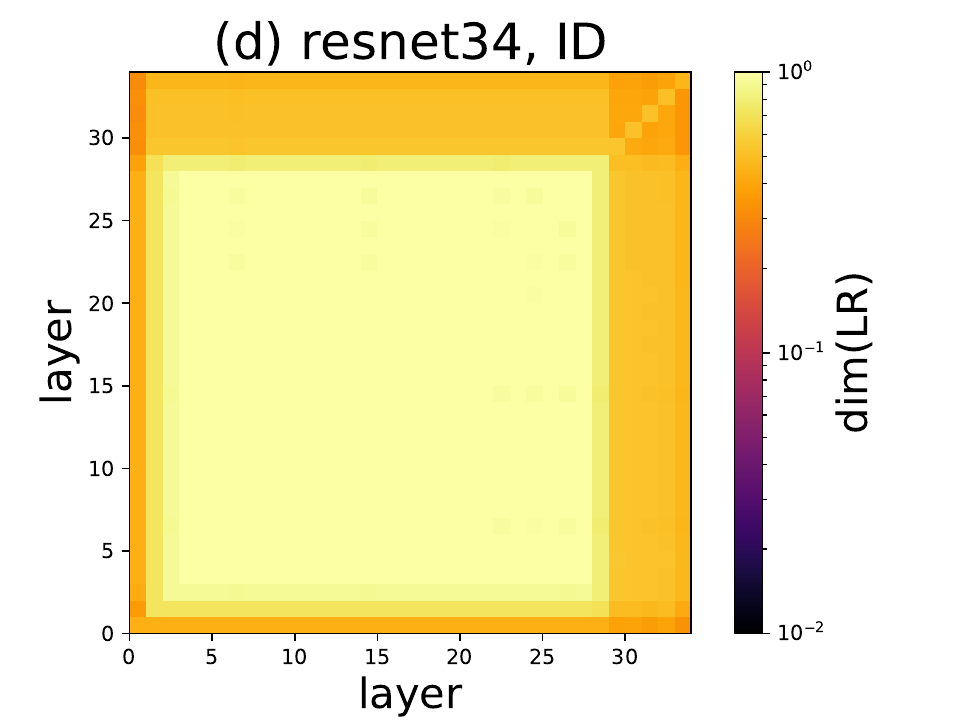}
  \end{tabular}
  \caption{
    Residual numbers of dimensions of LR matrices
    for the (a) VGG-13, (b) VGG-16, (c) ResNet-18, and (d) ResNet-34 models
    with smaller singular values removed by the threshold $\varepsilon_{\mathrm{sim}}$.
  }
  \label{fig:app_dims-lr}
\end{figure}

\begin{figure}[t]
\centering
  \begin{tabular}{cccc}
    \includegraphics[width=0.22\hsize, bb=0.000000 0.000000 460.800000 345.600000]{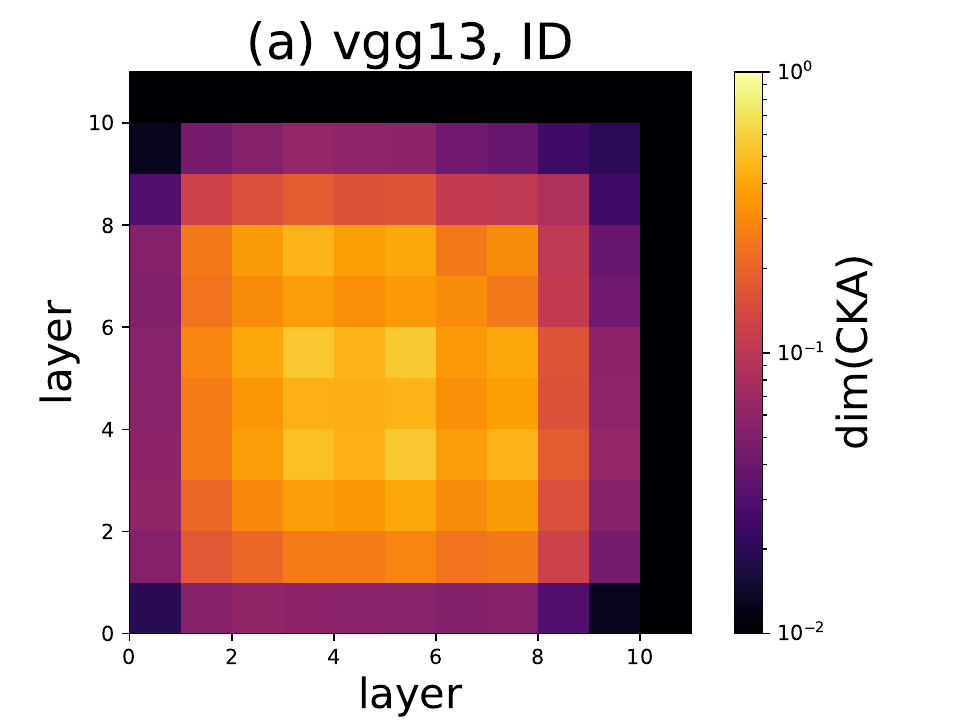}&
    \includegraphics[width=0.22\hsize, bb=0.000000 0.000000 460.800000 345.600000]{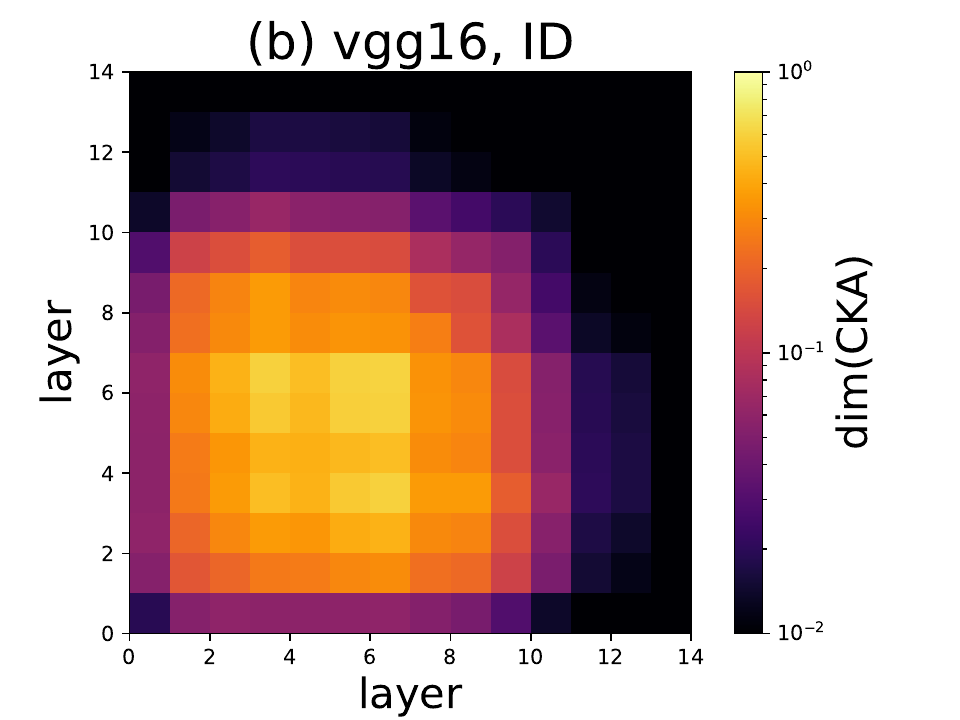}&
    \includegraphics[width=0.22\hsize, bb=0.000000 0.000000 460.800000 345.600000]{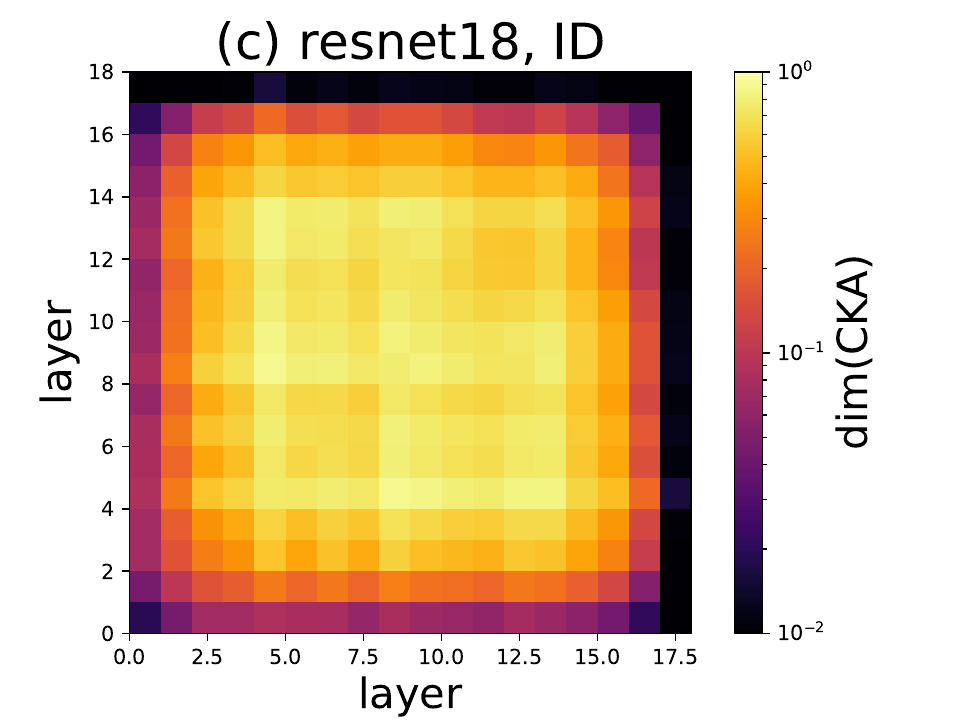}&
    \includegraphics[width=0.22\hsize, bb=0.000000 0.000000 460.800000 345.600000]{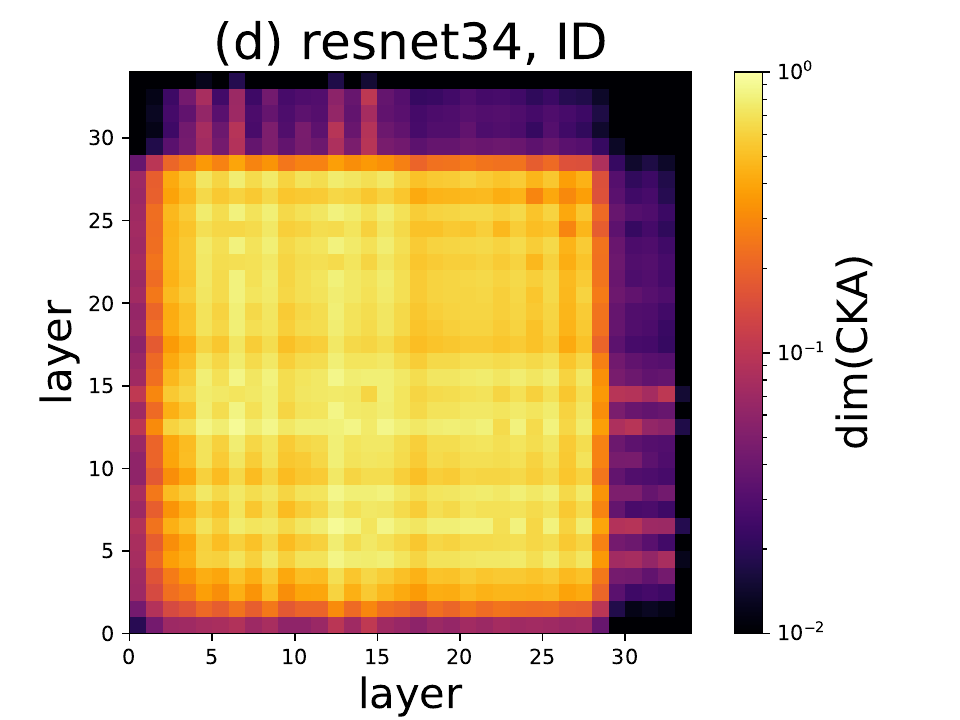}
  \end{tabular}
  \caption{
    Residual numbers of dimensions of CKA matrices
    for the (a) VGG-13, (b) VGG-16, (c) ResNet-18, and (d) ResNet-34 models
    with smaller singular values removed by the threshold $\varepsilon_{\mathrm{sim}}$.
  }
  \label{fig:app_dims-cka}
\end{figure}

\begin{figure}[t]
\centering
  \begin{tabular}{cccc}
    \includegraphics[width=0.22\hsize, bb=0.000000 0.000000 460.800000 345.600000]{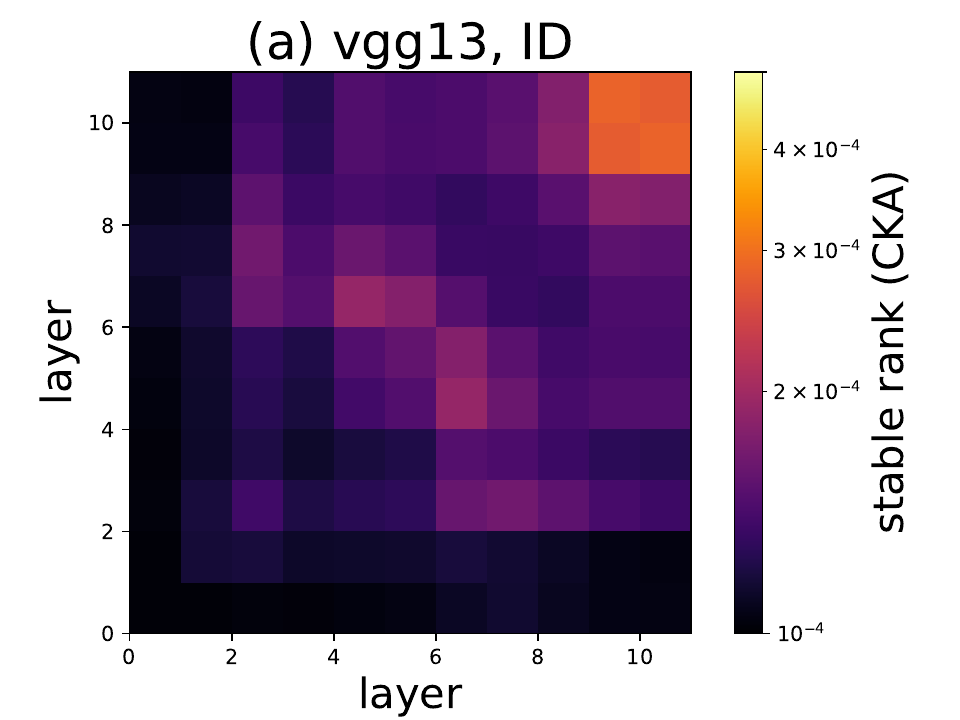}&
    \includegraphics[width=0.22\hsize, bb=0.000000 0.000000 460.800000 345.600000]{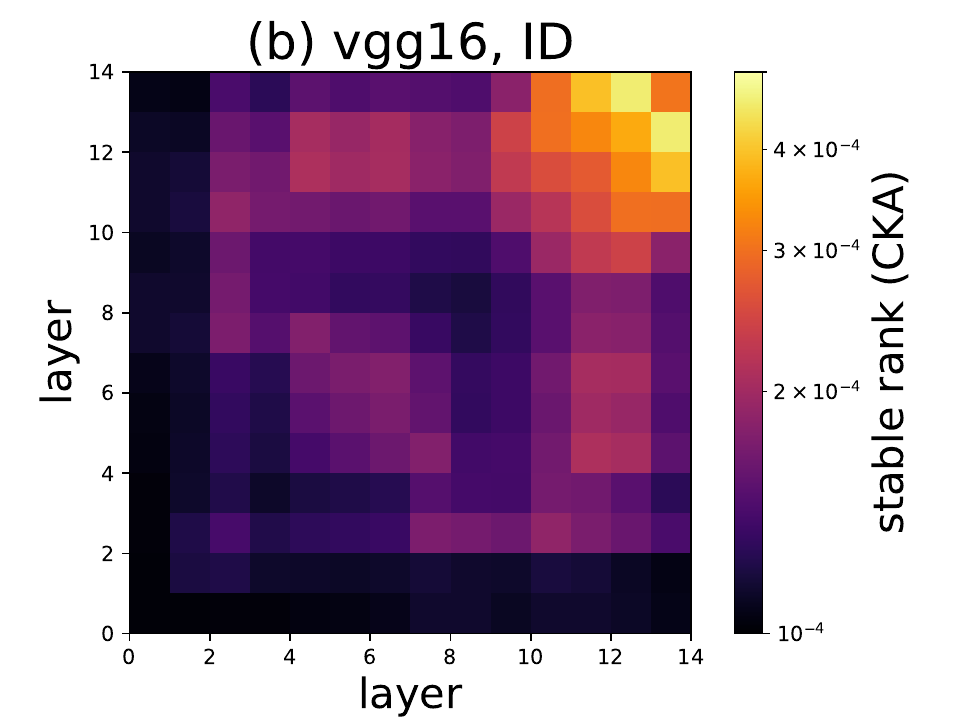}&
    \includegraphics[width=0.22\hsize, bb=0.000000 0.000000 460.800000 345.600000]{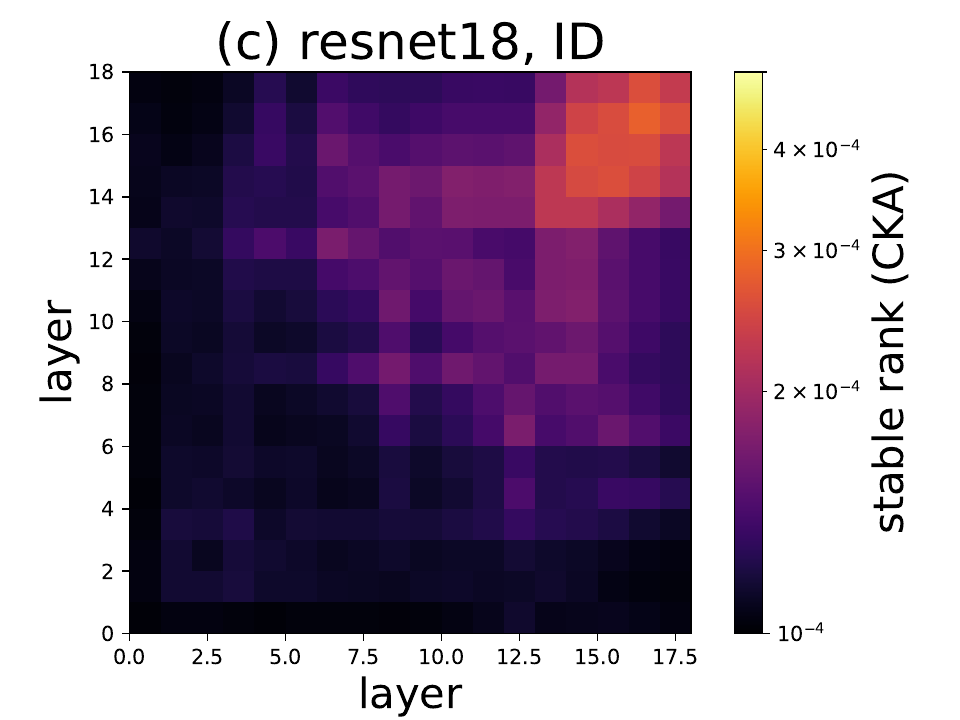}&
    \includegraphics[width=0.22\hsize, bb=0.000000 0.000000 460.800000 345.600000]{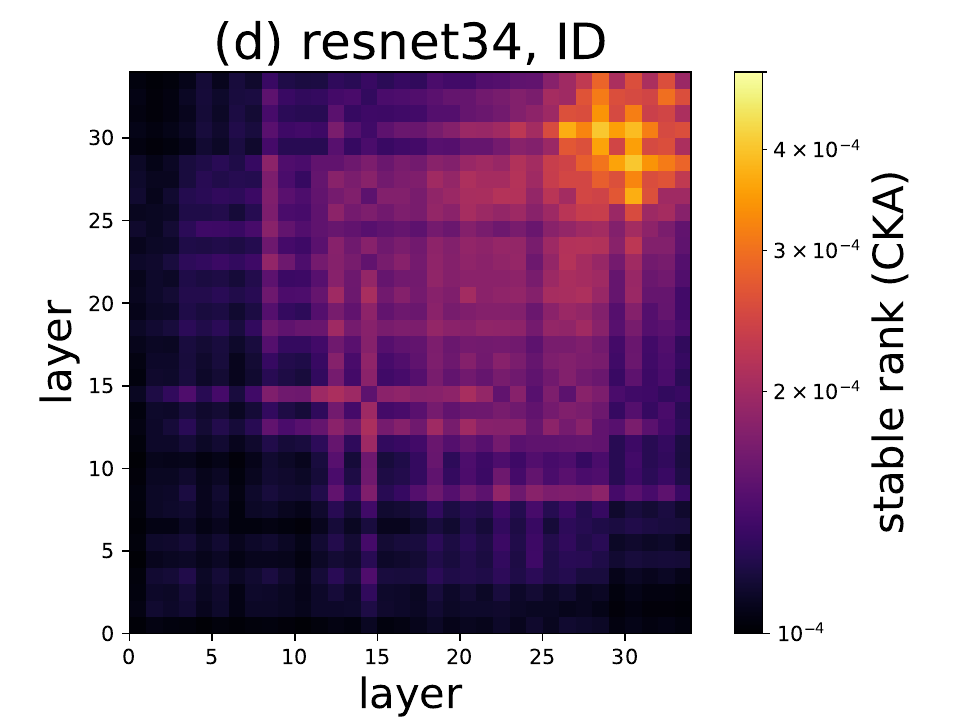}
    \\
    \includegraphics[width=0.22\hsize, bb=0.000000 0.000000 460.800000 345.600000]{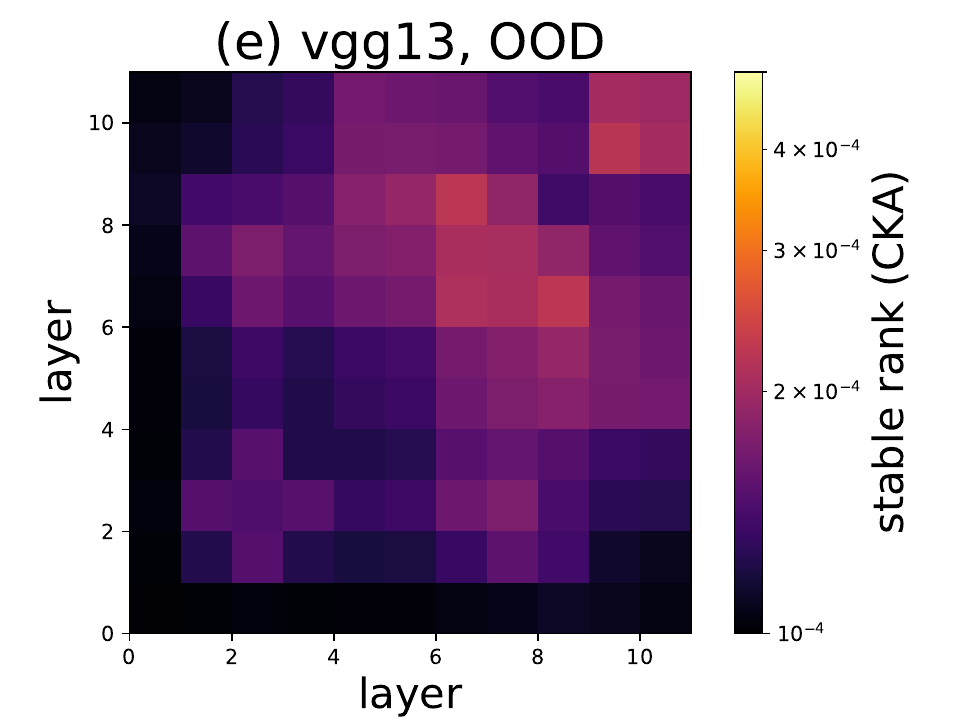}&
    \includegraphics[width=0.22\hsize, bb=0.000000 0.000000 460.800000 345.600000]{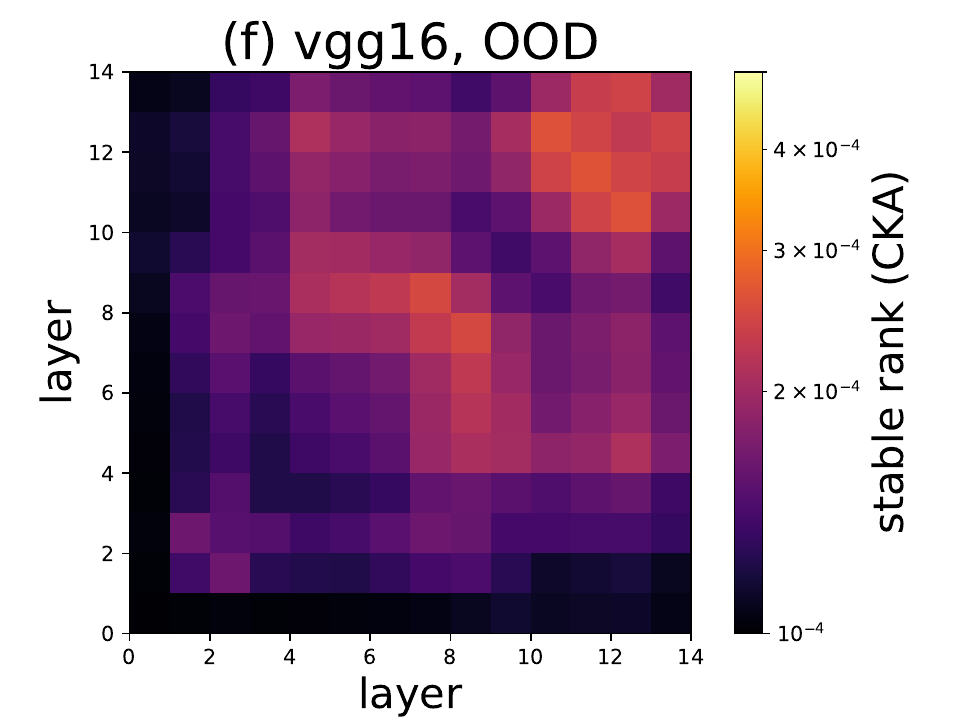}&
    \includegraphics[width=0.22\hsize, bb=0.000000 0.000000 460.800000 345.600000]{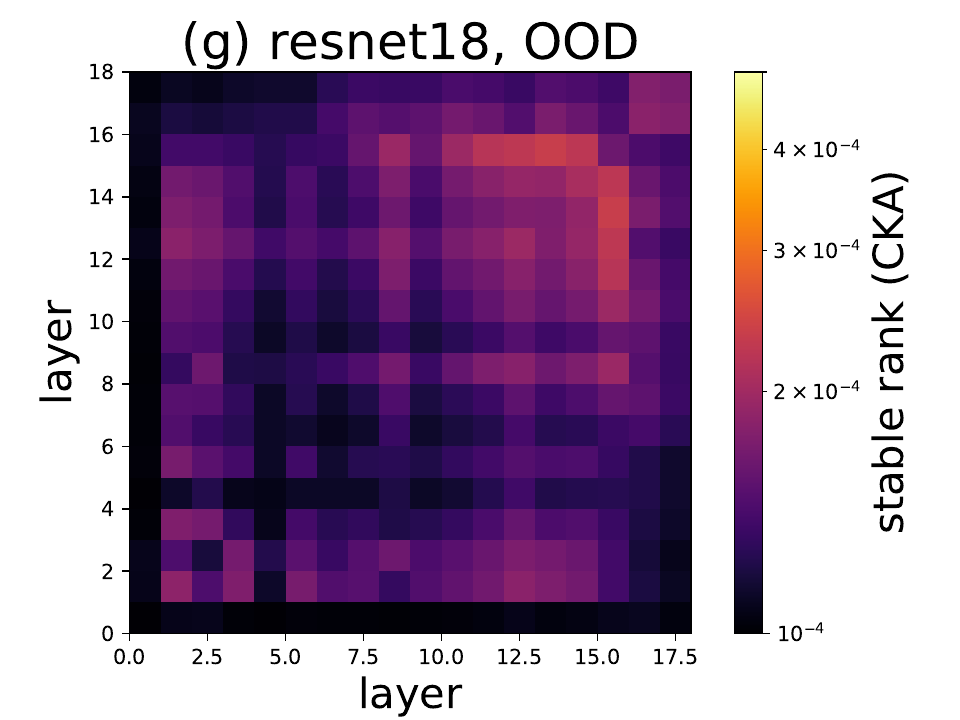}&
    \includegraphics[width=0.22\hsize, bb=0.000000 0.000000 460.800000 345.600000]{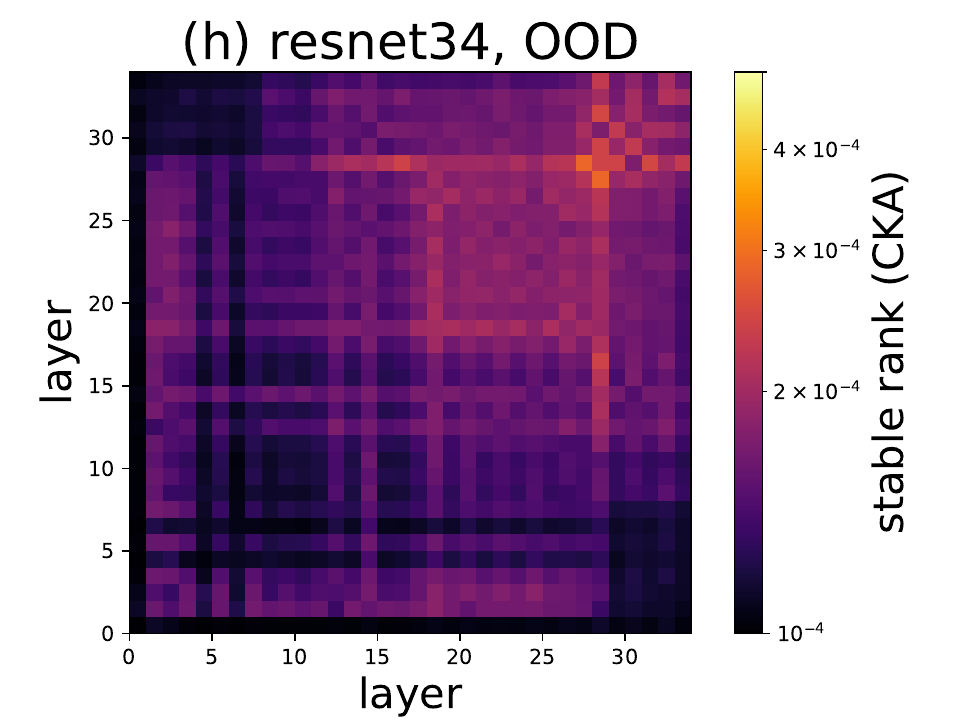}
  \end{tabular}
  \caption{
    Stable ranks of CKA matrices
    for the (a,e) VGG-13, (b,f) VGG-16, (c,g) ResNet-18, and (d,h) ResNet-34 models
    with smaller singular values removed by the threshold $\varepsilon_{\mathrm{sim}}$.
    The upper (a--d) and lower (e--g) figures show the stable ranks of the CKA matrices of ID (CIFAR-10) and OOD (CIFAR-100) samples, respectively.
  }
  \label{fig:app_srank-cka}
\end{figure}

\begin{enumerate}
  \item Compute inner products $(x_{i}^{(l)})^{\mathsf T} x_j^{(l)}$ for $j\geq i$ and save them.
  \item Obtain $K^{(l)}$ $=HK_0^{(l)}H$ as in Eq. \eqref{eq:cka}.
  \item Diagonalize $K^{(l)}$ $= P^{(l)} \Lambda^{(l)} (P^{(l)})^{\mathsf T}$.
  \item Remove eigenvalues such that $\lambda_i^{(l)}/\lambda_0^{(l)} < \varepsilon_{\mathrm{gram}}=10^{-6}$ and corresponding eigenvectors.
  \item Compute $D_{\varepsilon_{\mathrm{gram}}}^{(l_1,l_2)}$
    $=(P_{\varepsilon_{\mathrm{gram}}}^{(l_1)})^{\mathsf T} P_{\varepsilon_{\mathrm{gram}}}^{(l_2)}$
    before computing CKA given by the trace of $K^{(l_1)} K^{(l_2)}$.
  \item Diagonalize the positive semidefinite symmetric matrix
    $(\Lambda_{\varepsilon_{\mathrm{gram}}}^{(l_1)})^{1/2} D_{\varepsilon_{\mathrm{gram}}}^{(l_1,l_2)} \Lambda_{\varepsilon_{\mathrm{gram}}}^{(l_2)} (D_{\varepsilon_{\mathrm{gram}}}^{l_1,l_2})^{\mathsf T} (\Lambda_{\varepsilon_{\mathrm{gram}}}^{(l_1)})^{1/2} $
    $ = P_{\varepsilon_{\mathrm{gram}}}^{(l_1,l_2)} \Lambda_{\varepsilon_{\mathrm{gram}}}^{(l_1,l_2)} (P_{\varepsilon_{\mathrm{gram}}}^{(l_1,l_2)})^{\mathsf T}$,
    the trace of which is identical to $\mathrm{tr}(K^{(l_1)}K^{(l_2)})$ owing to the cyclic property of trace.
  \item Again remove irrelevant eigenvalues such that $\lambda_{\varepsilon_{\mathrm{gram}}, i}^{(l_1,l_2)}/\lambda_{\varepsilon_{\mathrm{gram}}, 0}^{(l_1,l_2)} < \varepsilon_{\mathrm{sim}}=10^{-6}$ and corresponding eigenvectors.
  \item Compute
    $\mathrm{CKA}_{\varepsilon_{\mathrm{gram}}, \varepsilon_{\mathrm{sim}}}(K^{(l_1)}, K^{(l_2)})$ \\
    $=\mathrm{tr}(\Lambda_{\varepsilon_{\mathrm{gram}}, \varepsilon_{\mathrm{sim}}}^{(l_1, l_2)}) / \sqrt{\mathrm{tr}((\Lambda_{\varepsilon_{\mathrm{gram}}}^{(l_1)})^2)\cdot \mathrm{tr}((\Lambda_{\varepsilon_{\mathrm{gram}}}^{(l_2)})^2)}$.
\end{enumerate}
Here, matrices with a subscript $\varepsilon$ represent the reconstructed ones using the remaining eigenvalues and eigenvectors similarly to the dimensionality-aware projection matrix in \ref{sec:app_compress}.
This quite resembles the procedure given in the singular vector canonical correlation analysis \cite{kernelCCA,CCA-retrieval,SVCCA}.

An advantage of this procedure is that we can simultaneously obtain other similarity scores from $D^{(l_1, l_2)}, \Lambda_{l_1}, \Lambda_{l_2}$.
For example, the similarity score given by the linear regression (LR) \cite{similarityrevisit} is
\begin{align*}
  \mathrm{LR}(K^{(l_1)}, K^{(l_2)}) &= \mathrm{tr}(D^{(l_1,l_2)} \Lambda_{l_2} D^{(l_2, l_1)}) / tr(\Lambda_{l_2}).
\end{align*}
As another example, the similarity score given by the canonical correlation analysis (CCA) can be represented by singular values of $D^{(l_1, l_2)}$ (e.g., $\left| D^{(l_1, l_2)} \right|_F^2 / \mathrm{dim}\left(D^{(l_1, l_2)}\right)$).
Note that $P^{l}$ in our notation is almost identical to $Q$ in Ref. \cite{similarityrevisit} except that the eigenvectors in $P^{(l)}$ refer to corresponding eigenvalues.

Another advantage of this procedure is that we can utilize the properties of the singular value decomposition.
As an example, we consider the case where we want to compare similarity scores with each other.
The decay of singular values in the similarity matrix reflects the stability in the sample selection in the dataset given that the dimension is determined by the number of samples.
If the smaller singular values of the similarity matrix are negligible compared with the largest one, additional samples would not change the similarity score as long as the sampling is performed equally.
On the other hand, an insufficient decay of singular values would lead to a sample-dependent similarity score.
A detailed analysis of similarity matrices might provide further information about such similarities.

We show the layer dependence of the number of residual singular values in similarity matrices of ID (CIFAR-10, test) samples in Figs. \ref{fig:app_dims-gram}--\ref{fig:app_dims-cka}.
Fig. \ref{fig:app_dims-gram} represents $\min(\mathrm{dim}(\Lambda_{\varepsilon_{\mathrm{gram}}}^{(l_1)}, \Lambda_{\varepsilon_{\mathrm{gram}}}^{(l_2)}))$,
and Figs. \ref{fig:app_dims-cca}--\ref{fig:app_dims-cka} represent the dimensions of CCA, LR, CKA matrices with smaller singular values removed in reference to $\varepsilon_{\mathrm{sim}}$.
From these figures, we can see that the dimension of the CCA matrices in Fig. \ref{fig:app_dims-cca} is identical to that in Fig. \ref{fig:app_dims-gram}, reflecting the fact that the singular values of $D^{(l_1,l_2)}$ are almost 1.
Similarly, the dimension of the LR matrices in Fig. \ref{fig:app_dims-lr} is quite similar to that in Fig. \ref{fig:app_dims-gram}, indicating that a single $\Lambda^{(l_1)}$ is not sufficient to set a threshold $\varepsilon_{\mathrm{sim}}$.
Compared with those, the dimension of the CKA matrices in Fig. \ref{fig:app_dims-cka} is significantly smaller than that in Fig. \ref{fig:app_dims-gram}, indicating that the simultaneous introduction of $\Lambda^{(l_1)}$ \textit{and} $\Lambda^{(l_2)}$ reduces the effective dimension of CKA.
This dimensional reduction in CKA likely originates from the multiplication of two gram matrices, $K^{(l_1)}$ and $K^{(l_2)}$, because the change of the squared value $x^2$ is larger than that of $x$.
Also, such a low-dimensional property of the CKA matrices would explain why the minibatch CKA is effective \cite{similarity-cnn}.

\begin{figure}[t]
\centering
  \begin{tabular}{cccc}
    \includegraphics[width=0.22\hsize, bb=0.000000 0.000000 460.800000 345.600000]{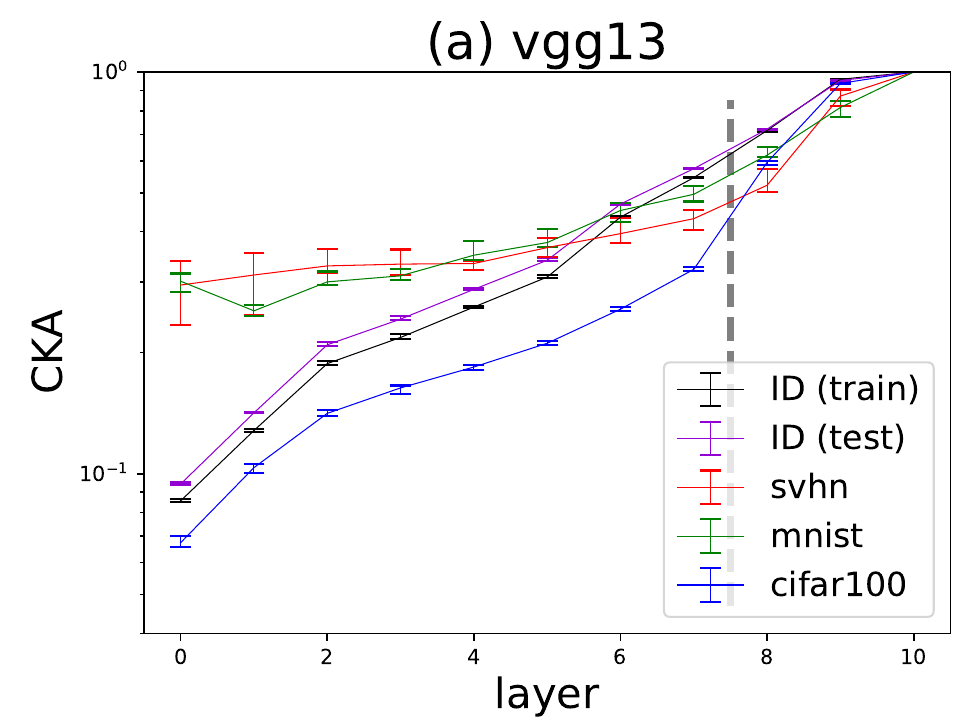}&
    \includegraphics[width=0.22\hsize, bb=0.000000 0.000000 460.800000 345.600000]{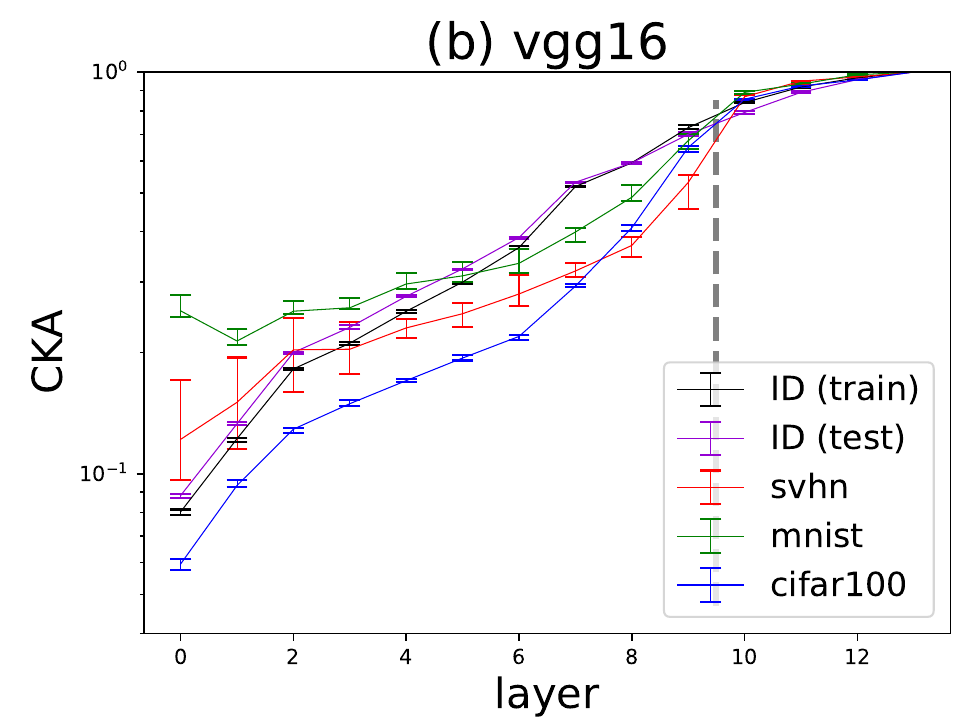}&
    \includegraphics[width=0.22\hsize, bb=0.000000 0.000000 460.800000 345.600000]{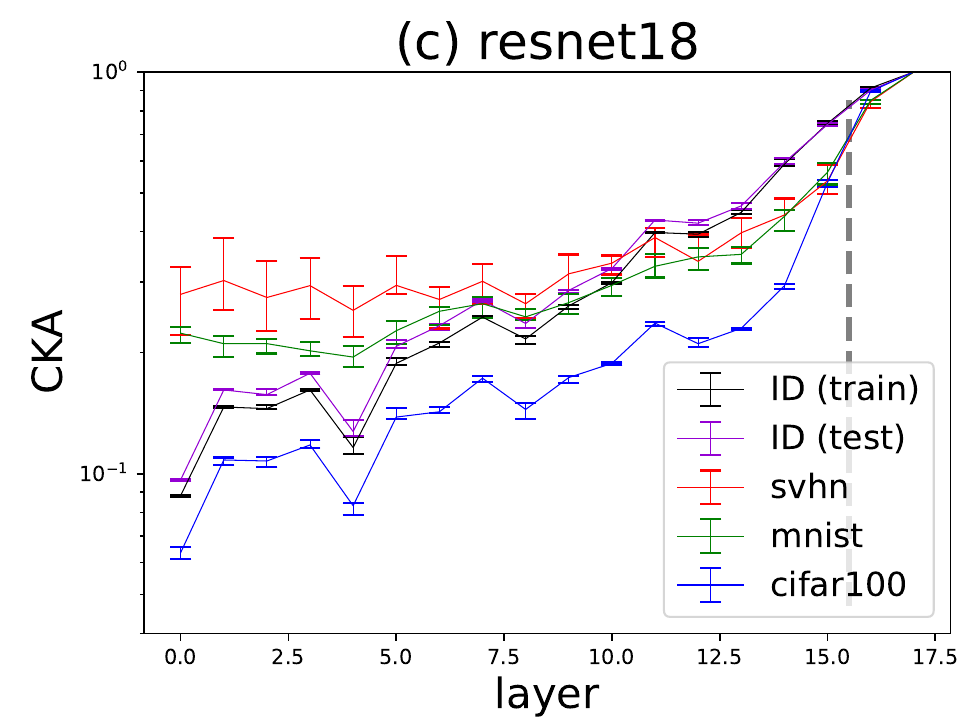}&
    \includegraphics[width=0.22\hsize, bb=0.000000 0.000000 460.800000 345.600000]{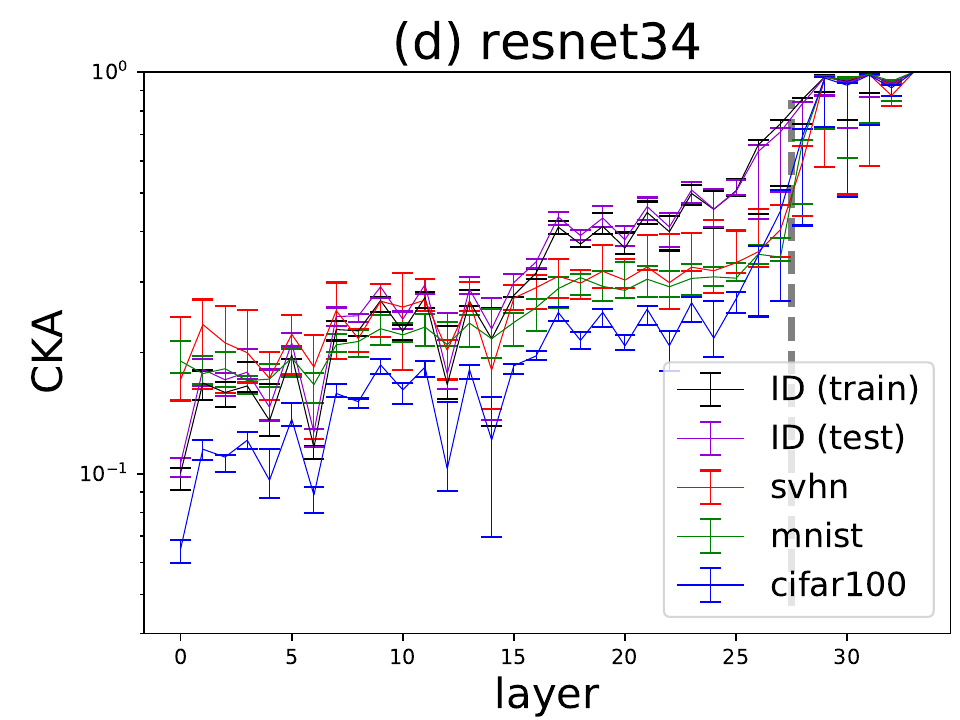}
  \end{tabular}
  \caption{
    CKA with the feature at the penultimate layer
    for the (a) VGG-13, (b) VGG-16, (c) ResNet-18, and (d) ResNet-34 models.
    Different line colors represent the different datasets.
    The dashed line indicates the transition layer.
    Just around the transition layer, CKA of OOD samples quickly decays compared with that of ID samples.
  }
  \label{fig:app_cka-penul}
\end{figure}

To further utilize the dimension of similarity matrices for analyzing OOD samples, we show the layer dependence of the stable ranks of CKA matrices of ID (CIFAR-10) and OOD (CIFAR-100) samples in Fig. \ref{fig:app_srank-cka}.
As can be seen from these figures, the stable ranks of interlayer CKA matrices of ID samples are comparably larger at deeper layers, while those of OOD samples are not.
This might be due to the fact that features of ID samples vary depending on the class of the input, while features of OOD samples only reflect the bias of the dataset.
Also, our low-rank property of OOD similarity is consistent with the observed low-rank feature vector of each OOD sample in Ref. \cite{OOD-RankFeat}.
Such a detailed evaluation of similarity matrices might also be informative for better understanding the DNN.

\section{CKA with the penultimate layer}
\label{sec:app_cka-penultimate}
As observed in Fig. \ref{fig:cka-layers} in the main text,
around the transition layer, we can find the gradual change of CKA in ID samples, while rather sharp in OOD samples.
To make it more visible, we extract the layer dependence of CKA with the features at the penultimate layer in Fig. \ref{fig:app_cka-penul},
corresponding to the top row or the right-most column in Figs. \ref{fig:cka-layers} (a-g) in the main text.
Focusing on the region around the transition layer, the decay of the similarity of OOD samples is much faster than that of ID samples around the transition layer represented by the dashed line.
This indicates that low dimensionalization cuts off most of the intrinsic information of OOD samples while it keeps a certain extent of ID samples.
A consistent behavior is also observed from the stable rank of CKA matrix in Fig. \ref{fig:app_srank-cka} in \ref{sec:app_similarity}, which would be highly related to the low-rank property of each OOD sample in Ref. \cite{OOD-RankFeat}.
In any event, such a loss of information would be a characteristic of OOD samples as described in Sec. \ref{sec:outline} in the main text.

When we focus on shallower layers far from the transition layer, 
far-from-ID OOD samples (SVHN, MNIST) exhibit higher CKA compared with ID samples and close-to-ID OOD samples (CIFAR-100).
This might be because the feature transformation in each layer gradually changes the features of ID samples while is independent of the features of far-from-ID OOD samples.
That is, the feature transformation gradually lowers CKA of ID samples, while retains that of far-from-ID OOD samples except the sharp decay around the transition layer.
This tendency can be seen in Fig. \ref{fig:app_cka-penul}.
For close-to-ID OOD samples, both of the gradual feature transformation and the transition layer lower CKA.
In fact, CKA of close-to-ID OOD samples is quite similar to that of ID samples except for the difference in constant multiplication originating from the sharp decay around the transition layer.

\begin{figure}[t]
\centering
  \begin{tabular}{cccc}
    \includegraphics[width=0.22\hsize, bb=0.000000 0.000000 460.800000 345.600000]{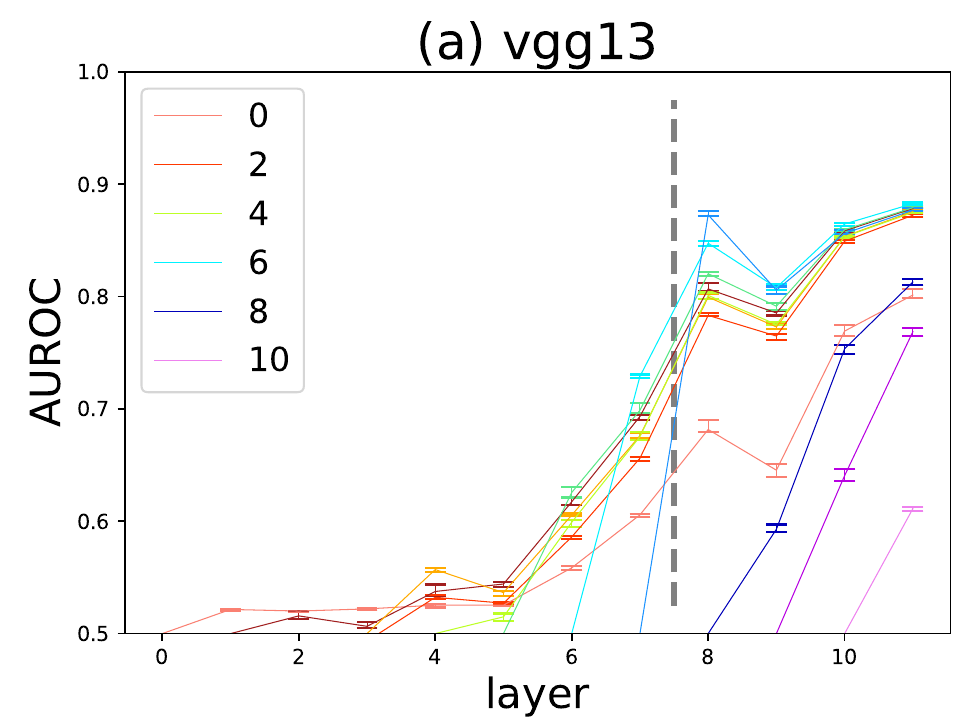}&
    \includegraphics[width=0.22\hsize, bb=0.000000 0.000000 460.800000 345.600000]{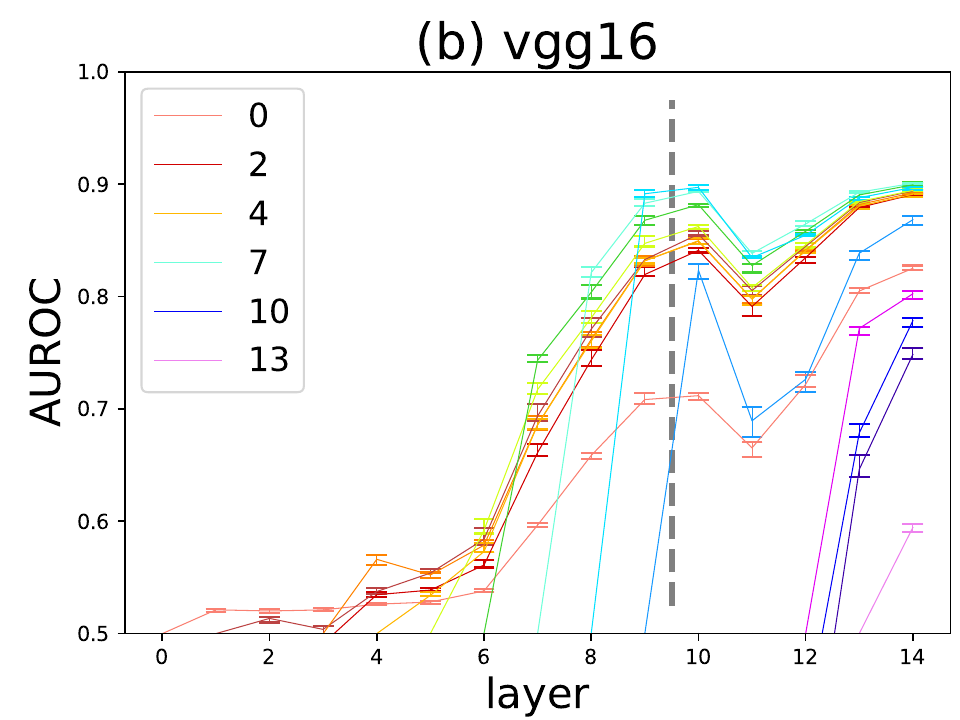}&
    \includegraphics[width=0.22\hsize, bb=0.000000 0.000000 460.800000 345.600000]{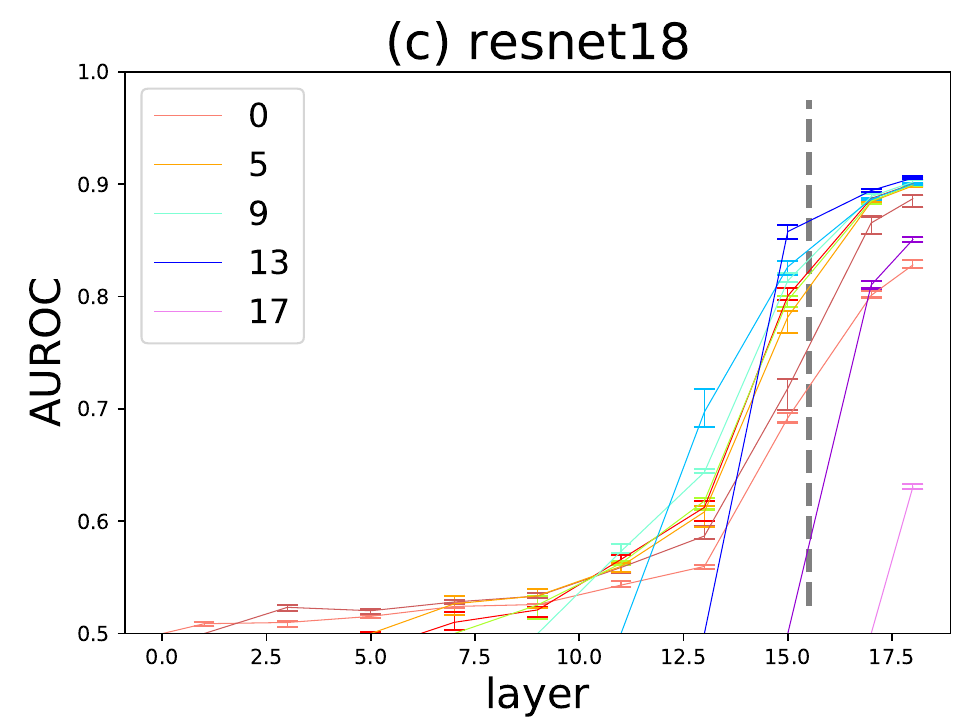}&
    \includegraphics[width=0.22\hsize, bb=0.000000 0.000000 460.800000 345.600000]{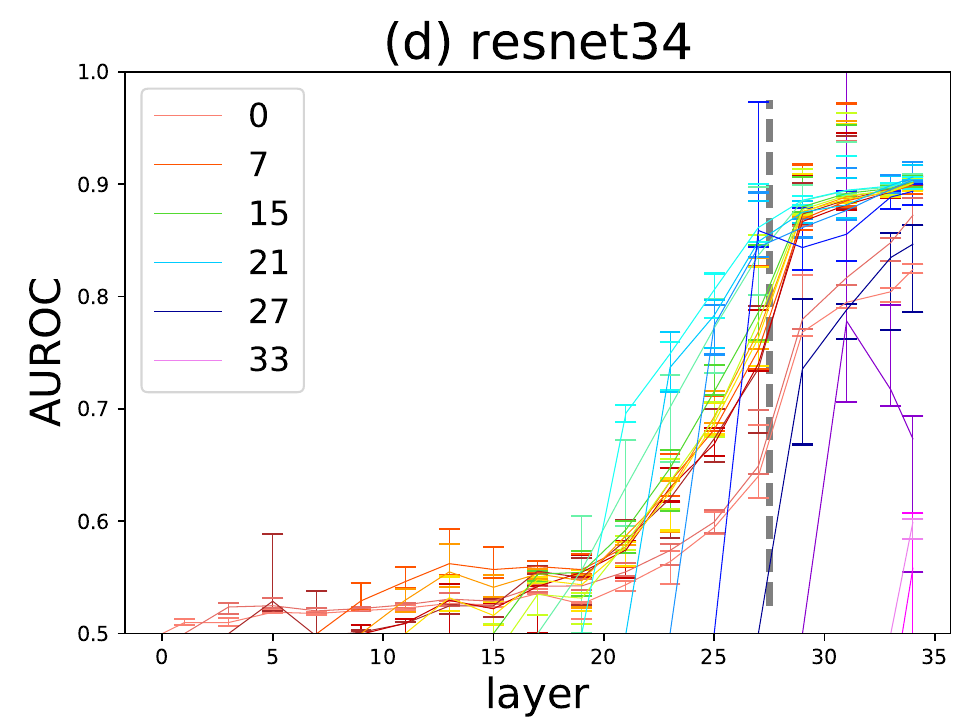}
  \end{tabular}
  \caption{
    Layer dependence of the AUROC detected by the noise sensitivity
    for the (a) VGG-13, (b) VGG-16, (c) ResNet-18, and (d) ResNet-34 models.
    The horizontal axis is the layer and the vertical axis is the corresponding AUROC.
    The different colors represent the different input layers where noise is injected.
  }
  \label{fig:app_sensitivity_AUROC}
\end{figure}

\section{OOD detection using noise sensitivity}
\label{sec:app_auroc-sensitivity}

Fig. \ref{fig:app_sensitivity_AUROC} represents the layer dependence of AUROC detected by the noise sensitivity when ID and OOD are CIFAR-10 and CIFAR-100, respectively.
We can see that, for accurate OOD detection, the noise-injected layer should be shallower than the transition layer, while the sensitivity-measured layer should be deeper than the transition layer.
This corresponds well with the observed behavior in the main text in Sec. \ref{sec:results}.
Although the differences are not very significant, the best noise-injected layer tends to be the second shallowest layer, while the best sensitivity-measured layer tends to be the deepest layer.
The AUROC is located at $\sim 0.90$, which is a little better than the probability-based detection ($\sim 0.88$) but a little worse than the feature-based one ($\sim 0.92$).


\begin{figure}[t]
\centering
  \begin{tabular}{cccc}
    \includegraphics[width=0.22\hsize, bb=0.000000 0.000000 460.800000 345.600000]{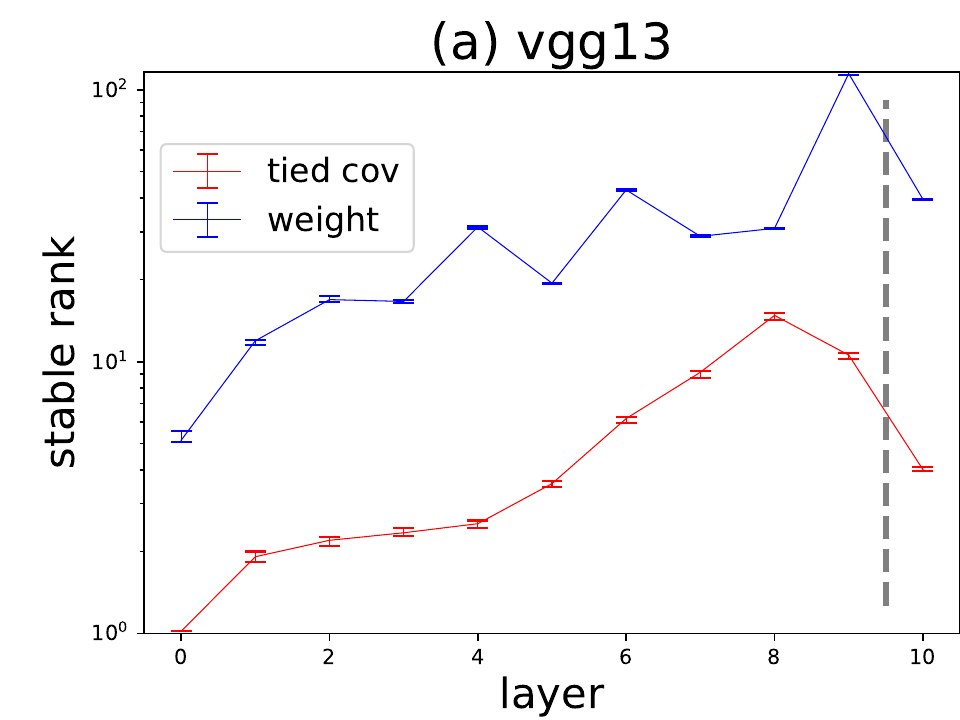}&
    \includegraphics[width=0.22\hsize, bb=0.000000 0.000000 460.800000 345.600000]{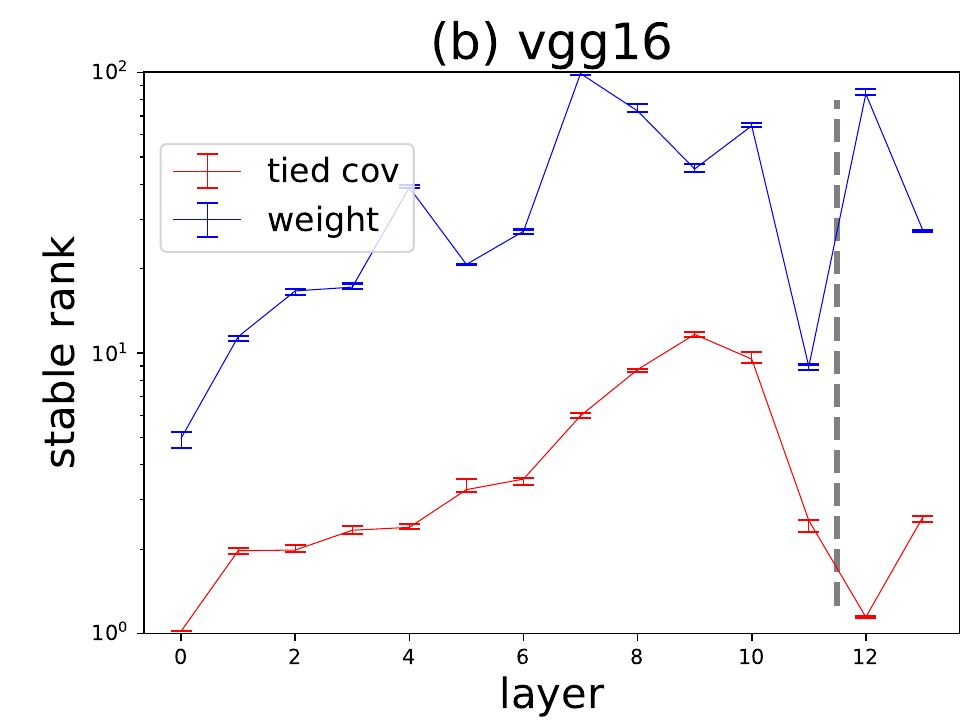}&
    \includegraphics[width=0.22\hsize, bb=0.000000 0.000000 460.800000 345.600000]{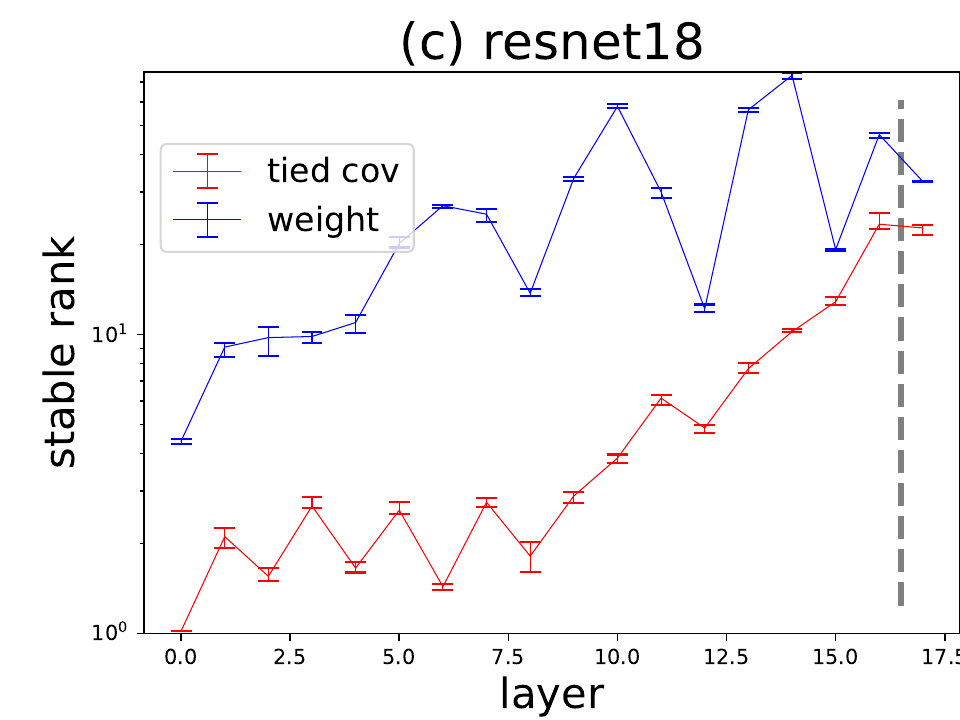}&
    \includegraphics[width=0.22\hsize, bb=0.000000 0.000000 460.800000 345.600000]{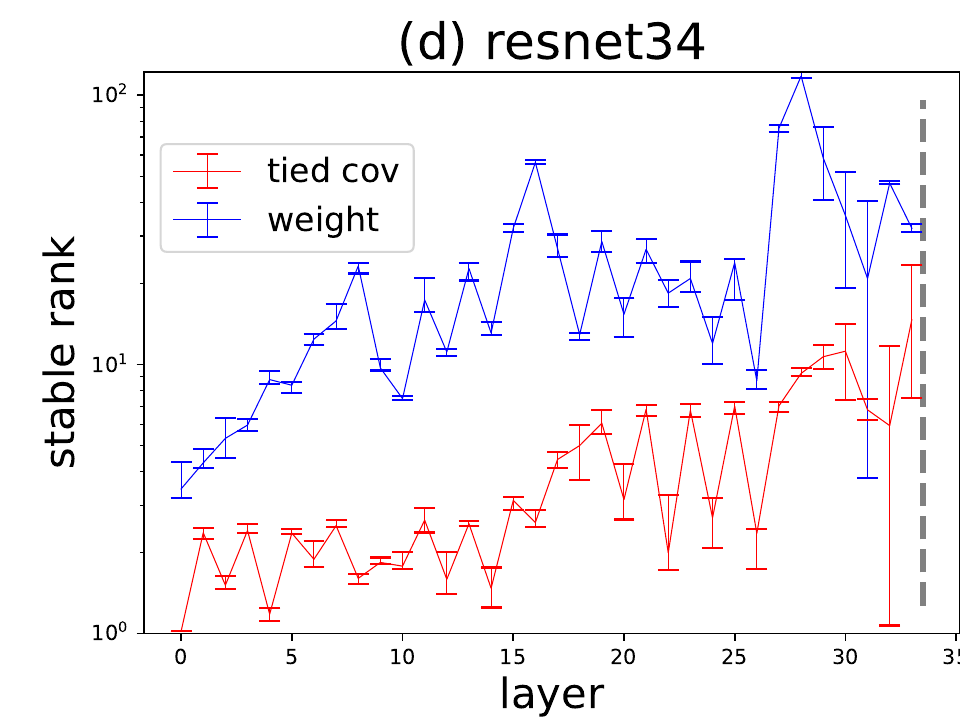}
  \end{tabular}
  \caption{
    Layer dependence of the stable rank of the covariance matrix $\overline{\Sigma}$ and the weight matrix $W$
    for the (a) VGG-13, (b) VGG-16, (c) ResNet-18, and (d) ResNet-34 models.
  }
  \label{fig:app_stableranks_cifar100}
\end{figure}

\begin{figure}[t]
\centering
  \begin{tabular}{cccc}
    \includegraphics[width=0.22\hsize, bb=0.000000 0.000000 460.800000 345.600000]{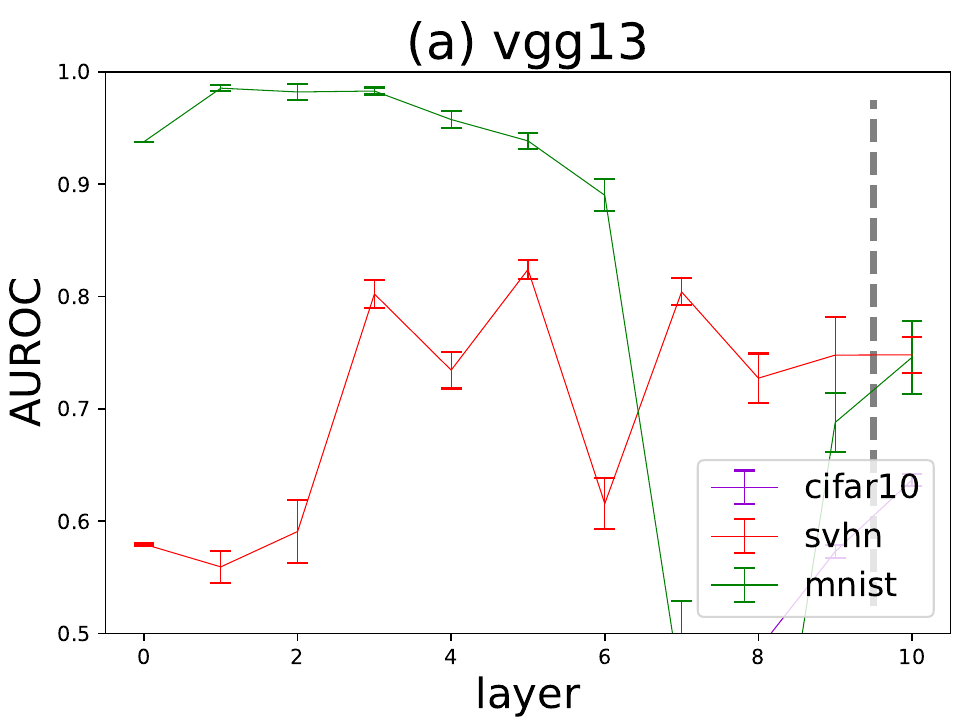}&
    \includegraphics[width=0.22\hsize, bb=0.000000 0.000000 460.800000 345.600000]{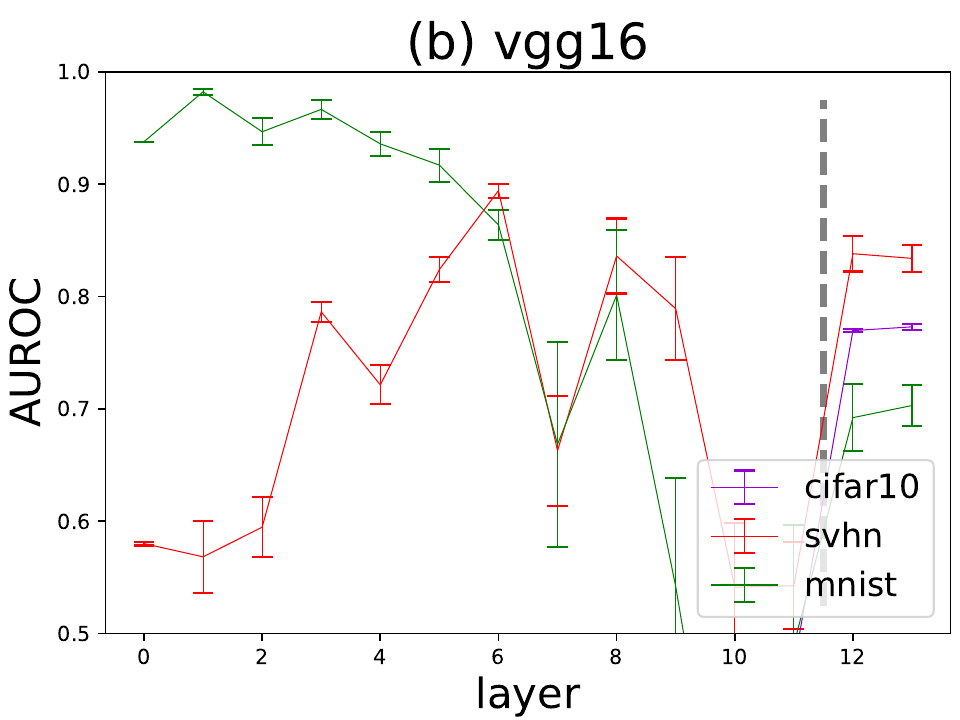}&
    \includegraphics[width=0.22\hsize, bb=0.000000 0.000000 460.800000 345.600000]{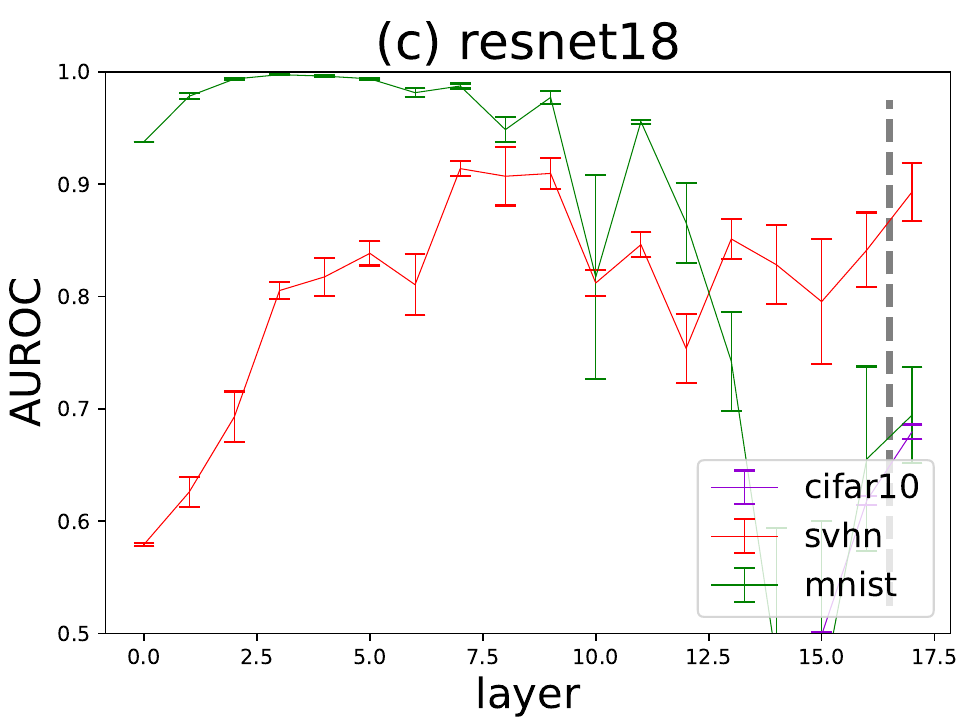}&
    \includegraphics[width=0.22\hsize, bb=0.000000 0.000000 460.800000 345.600000]{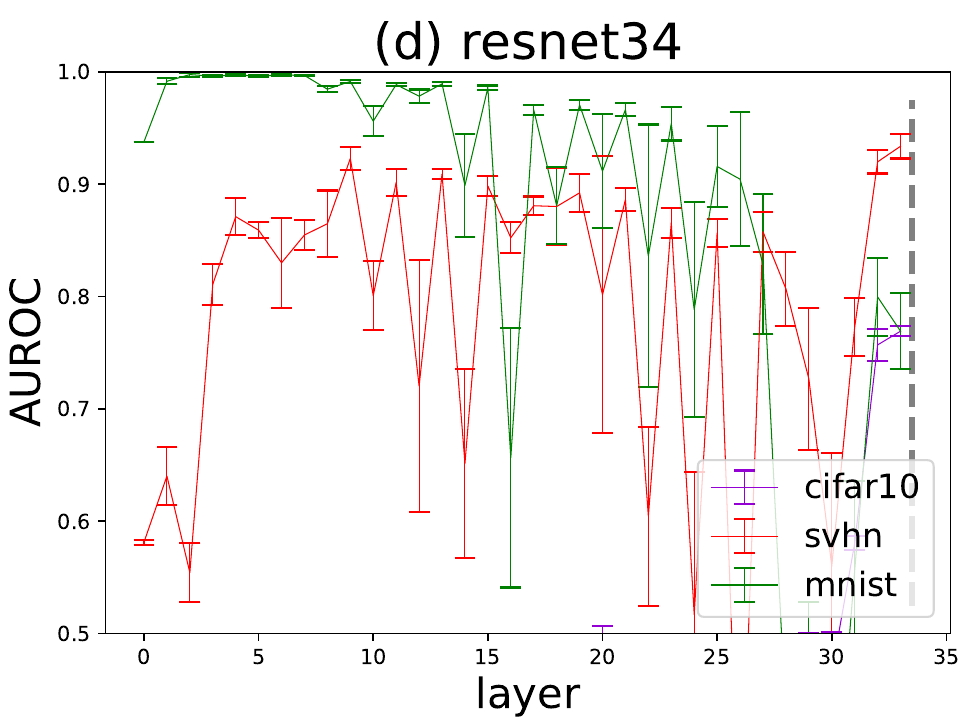}
  \end{tabular}
  \caption{
    Layer dependence of the AUROC detected by the Mahalanobis distance $M(x)$
    for the (a) VGG-13, (b) VGG-16, (c) ResNet-18, and (d) ResNet-34 models.
    The different line colors represent the different OOD datasets evaluated.
  }
  \label{fig:app_auroc-cov_cifar100}
\end{figure}

\begin{figure}[t]
  \centering
  \begin{tabular}{cccc}
    \includegraphics[width=0.22\hsize, bb=0.000000 0.000000 460.800000 345.600000]{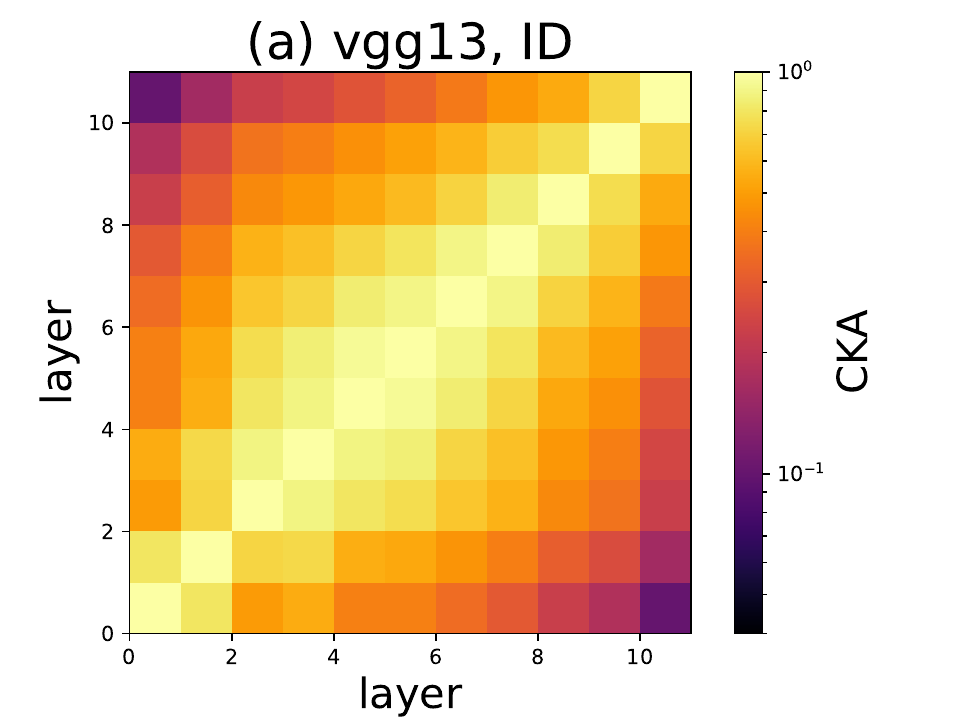}&
    \includegraphics[width=0.22\hsize, bb=0.000000 0.000000 460.800000 345.600000]{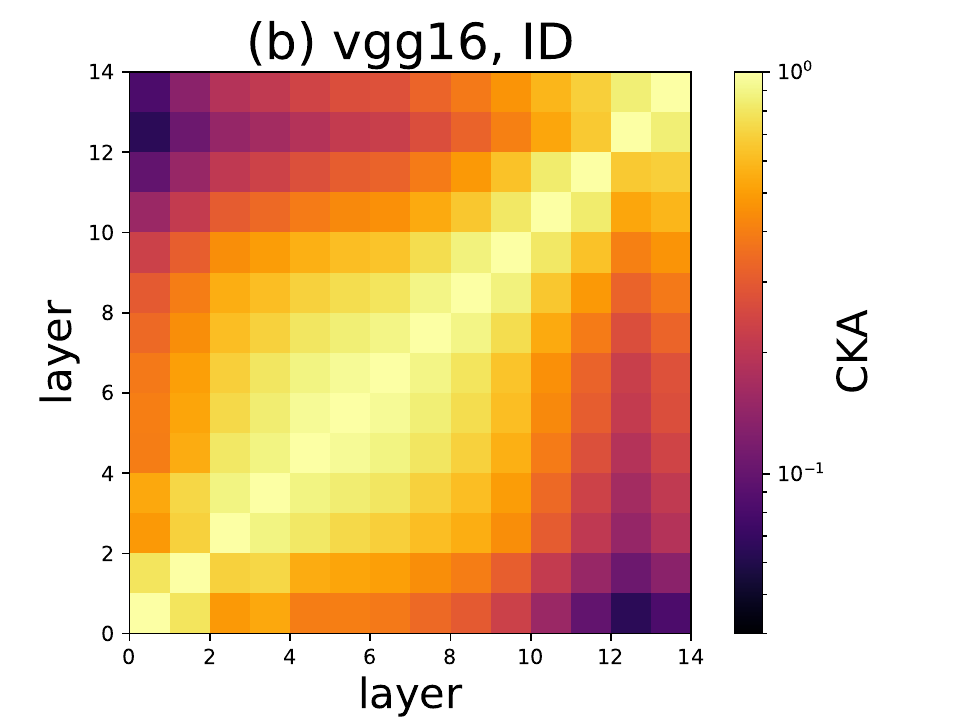}&
    \includegraphics[width=0.22\hsize, bb=0.000000 0.000000 460.800000 345.600000]{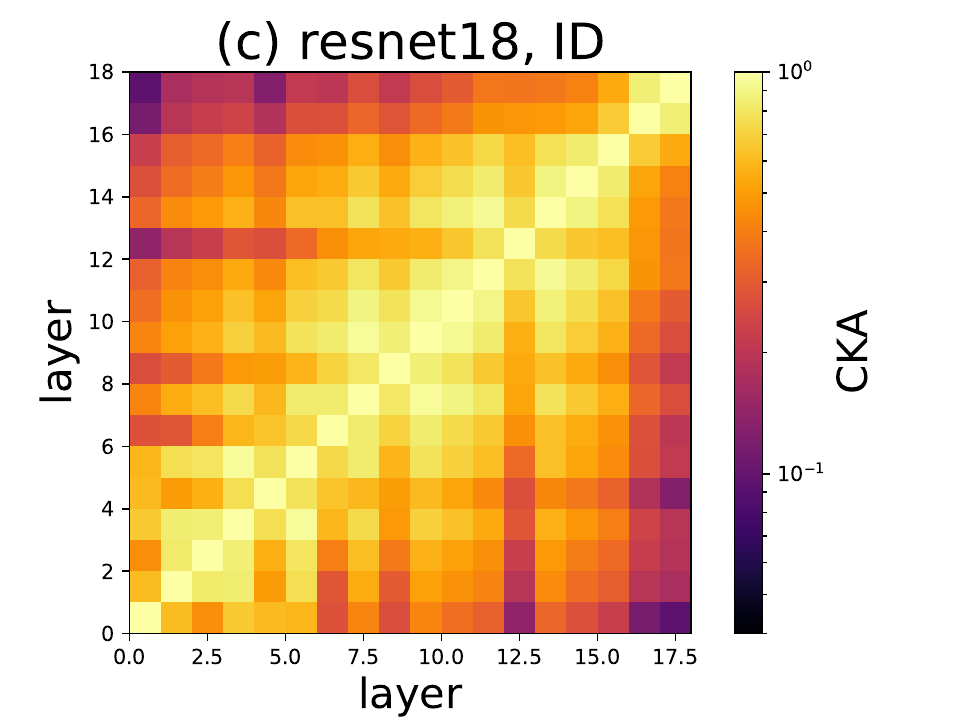}&
    \includegraphics[width=0.22\hsize, bb=0.000000 0.000000 460.800000 345.600000]{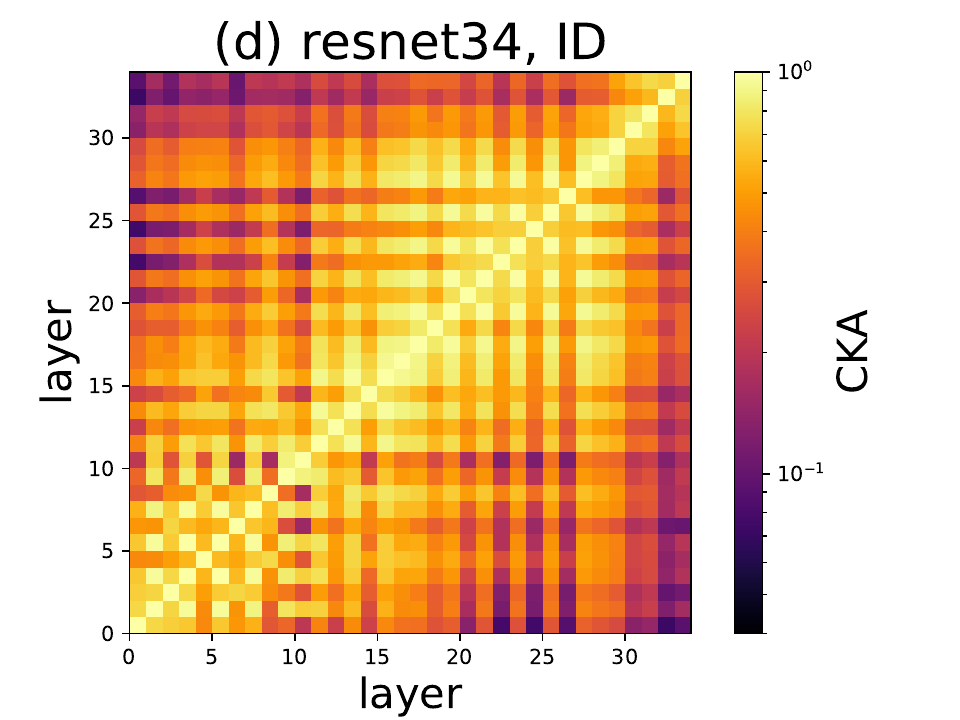}
  \end{tabular}
  \caption{
    CKA of features at various layers
    for the (a) VGG-13, (b) VGG-16, (c) ResNet-18, and (d) ResNet-34 models.
    In each figure, the horizontal and vertical axes represent layers, and the color bar represents CKA.
  }
  \label{fig:app_cka-layers_cifar100}
\end{figure}

\begin{figure}[t]
  \centering
  \begin{tabular}{cccc}
    \includegraphics[width=0.22\hsize, bb=0.000000 0.000000 460.800000 345.600000]{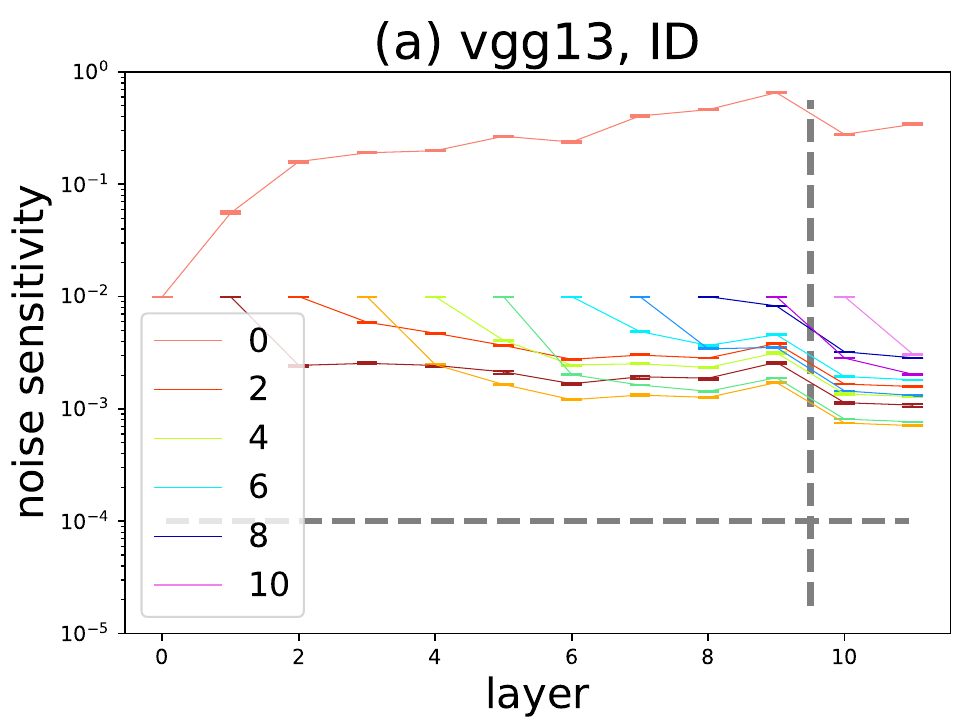}&
    \includegraphics[width=0.22\hsize, bb=0.000000 0.000000 460.800000 345.600000]{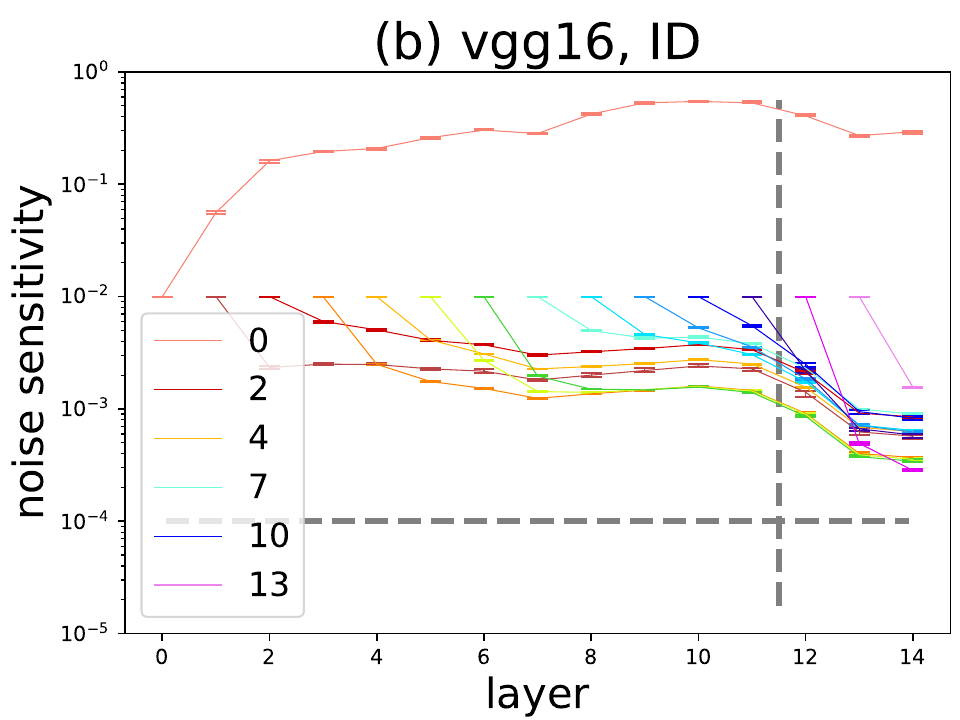}&
    \includegraphics[width=0.22\hsize, bb=0.000000 0.000000 460.800000 345.600000]{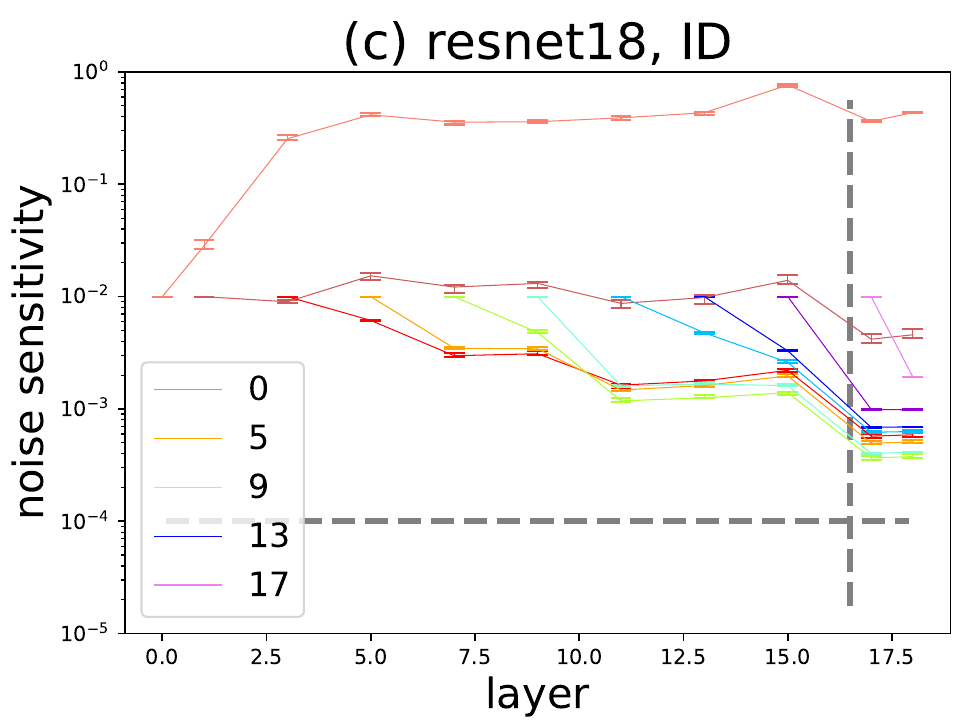}&
    \includegraphics[width=0.22\hsize, bb=0.000000 0.000000 460.800000 345.600000]{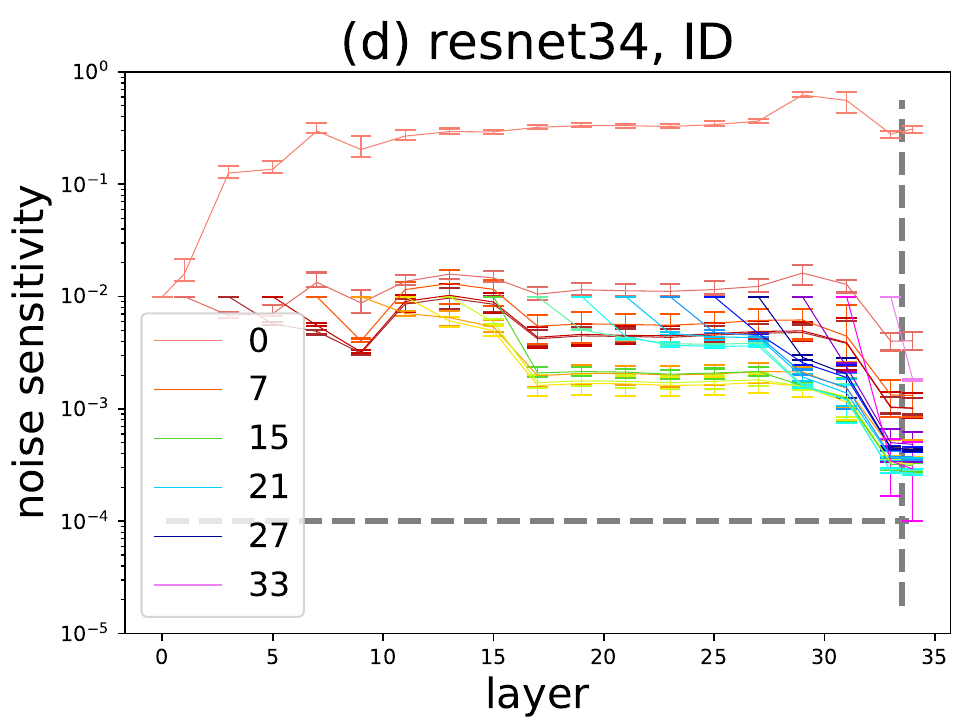}
  \end{tabular}
  \caption{
    Noise sensitivity
    for the (a,e) VGG-13, (b,f) VGG-16, (c,g) ResNet-18, and (d,h) ResNet-34 models.
    In each figure, the horizontal axis is the layer
    and the vertical axis is the corresponding noise sensitivity.
    The different colors represent the different input layers where noise is injected.
  }
  \label{fig:app_sensitivity_cifar100}
\end{figure}

\section{Analysis of CIFAR-100}
\label{sec:app_cifar100}
In this Appendix section, we provide the results for the case where the CIFAR-100 dataset is taken to be ID samples.
CIFAR-100 has much larger number of classes in ID samples, and thus leads to poor low dimensionalization.
The following behaviors suggest the importance of ID optimization for OOD detection.

Poor low dimensionalization can be observed by the stable rank in Fig. \ref{fig:app_stableranks_cifar100}.
In contrast with the case in which CIFAR-10 is ID samples (Fig. \ref{fig:stableranks} in the main text),
the stable rank of the weight matrices in Fig. \ref{fig:app_stableranks_cifar100} does not reach 100, which is expected from the number of classes in CIFAR-100.
This suggests that the DNN should be much deeper and wider in order to obtain a classifier that is good enough to detect OOD samples.

Poor low dimensionalization causes poor OOD detection performance as in Fig. \ref{fig:app_auroc-cov_cifar100}, which shows the layer dependence of AUROC detected by the Mahalanobis distance.
Although detection performance of close-to-ID OOD samples (CIFAR-1) tends to increase with deeper layers,
it does not reach the saturation, suggesting that the number of parameters in the DNN is insufficient.
Related to this, detection performance of other OOD samples is highly layer-dependent in contrast with stable detection performance in Fig. \ref{fig:auroc} in the main text.
These results indicate the difficulty of OOD detection in cases with a large number of classes in ID samples.

Insufficient low dimensionalization can also be checked from CKA in Fig. \ref{fig:app_cka-layers_cifar100} and the noise sensitivity in Fig. \ref{fig:app_sensitivity_cifar100}.
The block-like structure is not visible from Fig. \ref{fig:app_cka-layers_cifar100},
and the comparably large noise sensitivity can be seen in Fig. \ref{fig:app_sensitivity_cifar100}.
These behaviors are in sharp contrast to the case where CIFAR-10 is ID samples as can be seen in Figs. \ref{fig:cka-layers} and \ref{fig:sensitivity}, further supporting the idea that the DNN should be tuned for ID samples.

\section{Further verification in SVHN and MNIST}
\label{sec:app_svhn-mnist}
In this Appendix section, we provide further verification in the case that ID dataset has comparably smaller number of classes as CIFAR-10. 
For this purpose, we show the results for the case where the SVHN and MNIST datasets are taken to be ID samples.
The experimental detail is quite similar to that in the main text, however, the close-to-ID OOD samples are absent from the analysis in this Appendix section.
This is in contrast to the case where the CIFAR-100 dataset corresponds to the close-to-ID OOD samples when the CIFAR-10 dataset is used as ID dataset.
This may provide slightly different visual impressions compared with those in the main text, especially for the layer dependence of AUROCs.
Yet, low dimensionalization is still essential for consistent understanding of results to be presented.
Note that we also check the consistency in different model architectures as in Appendix \ref{sec:app_full-model}.

\begin{figure}[t]
\centering
  \begin{tabular}{cc}
    \includegraphics[width=0.22\hsize, bb=0.000000 0.000000 460.800000 345.600000]{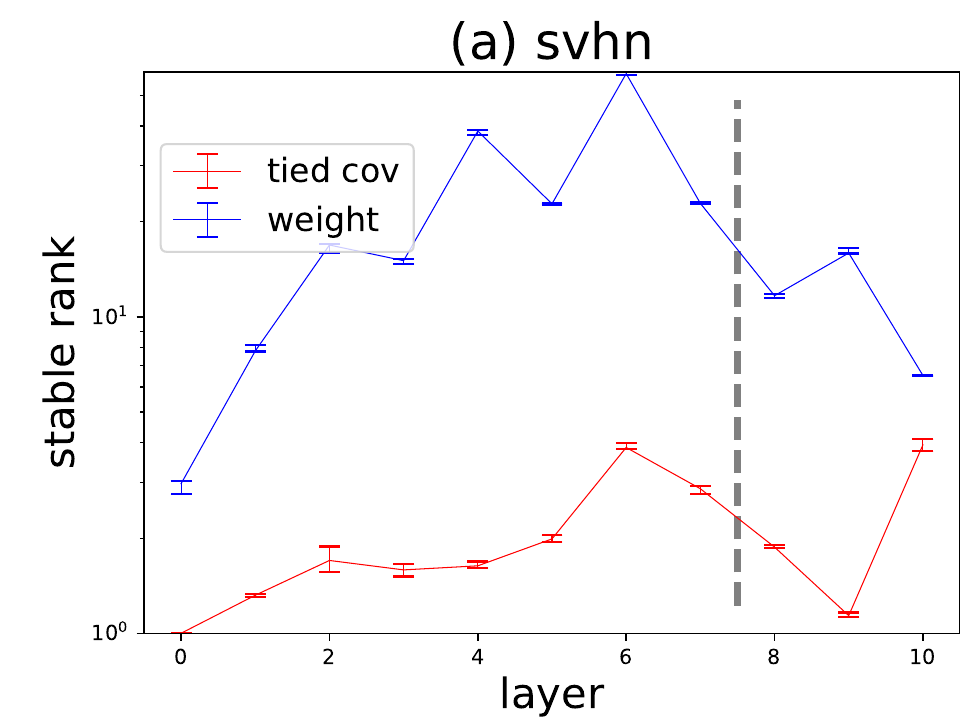}&
    \includegraphics[width=0.22\hsize, bb=0.000000 0.000000 460.800000 345.600000]{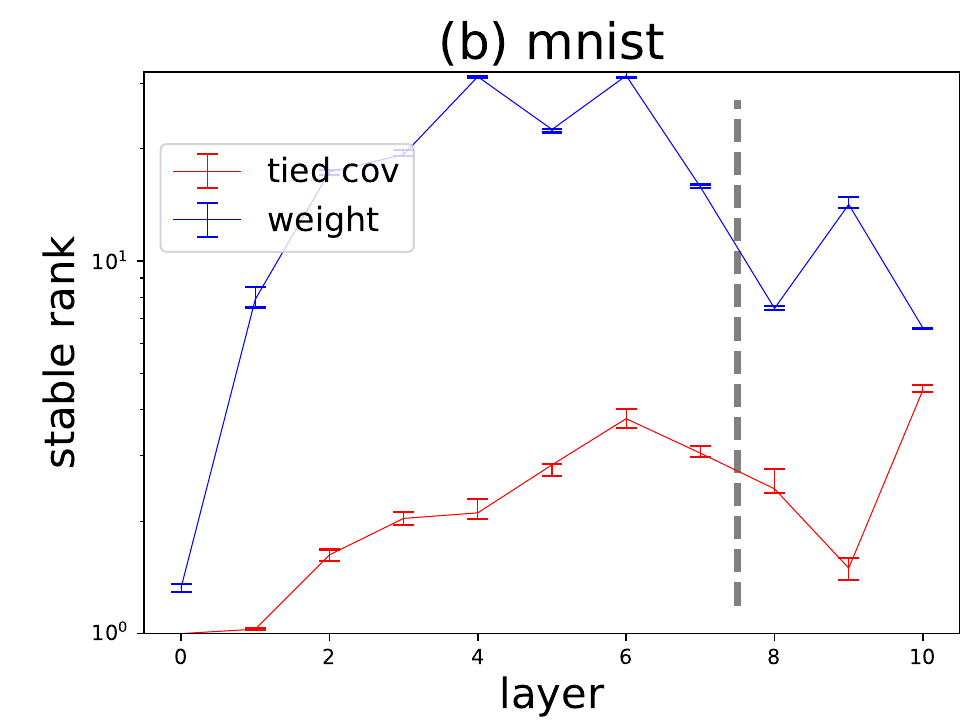}
  \end{tabular}
  \caption{
    Layer dependence of the stable rank of the covariance matrix $\overline{\Sigma}$ and the weight matrix $W$
    for the VGG-13 model with (a) SVHN and (b) MNIST taken to be the ID dataset.
  }
  \label{fig:app_stableranks_svhn-mnist}
\end{figure}

\begin{figure}[t]
\centering
  \begin{tabular}{cc}
    \includegraphics[width=0.22\hsize, bb=0.000000 0.000000 460.800000 345.600000]{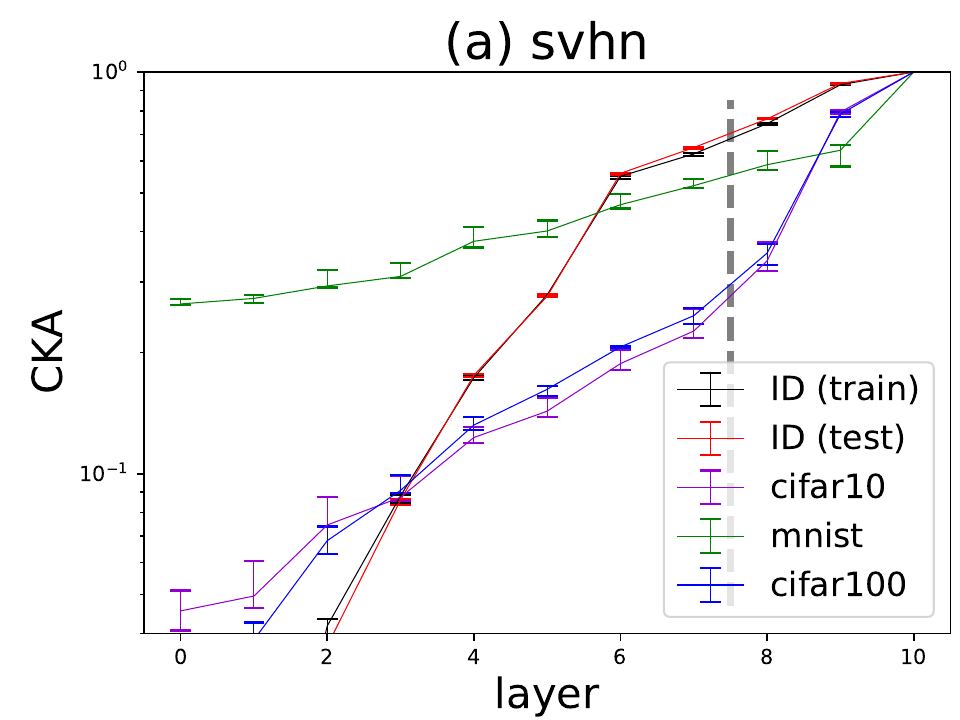}&
    \includegraphics[width=0.22\hsize, bb=0.000000 0.000000 460.800000 345.600000]{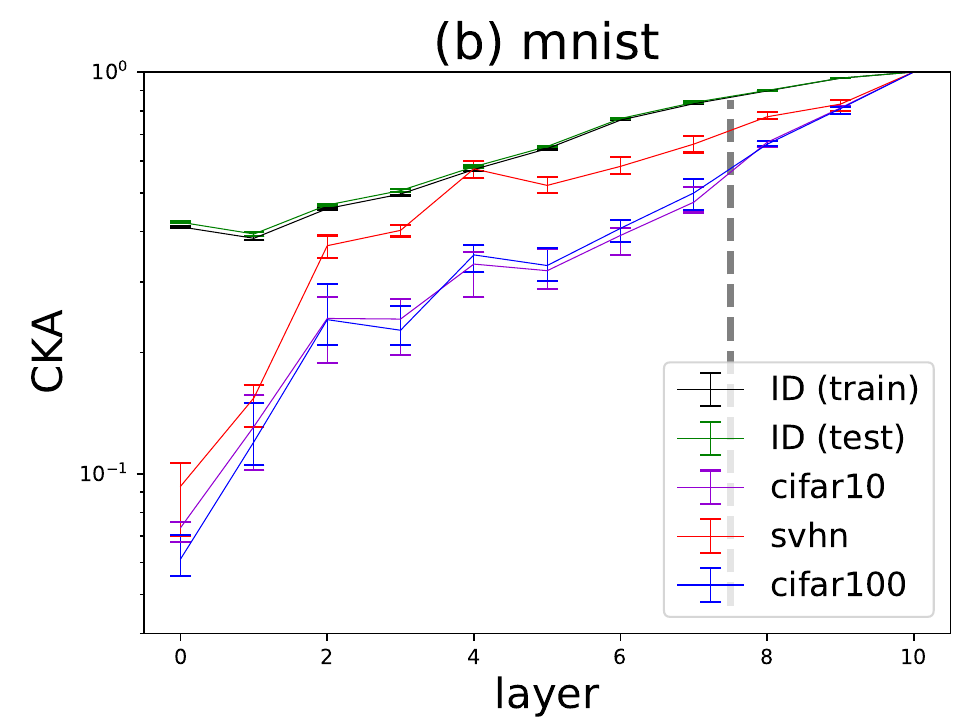}
  \end{tabular}
  \caption{
    CKA with the feature at the penultimate layer
    for the VGG-13 model with (a) SVHN and (b) MNIST taken to be the ID dataset.
    Different line colors represent the different datasets.
  }
  \label{fig:app_cka-penul_svhn-mnist}
\end{figure}

\begin{figure}[t]
\centering
  \begin{tabular}{cccc}
    \includegraphics[width=0.22\hsize, bb=0.000000 0.000000 460.800000 345.600000]{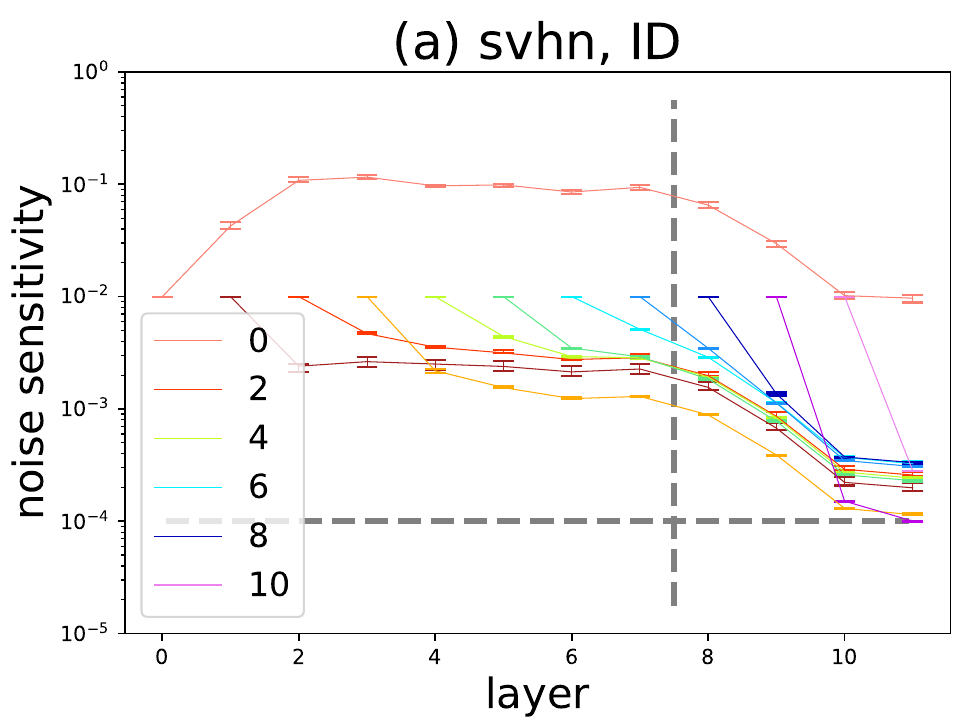}&
    \includegraphics[width=0.22\hsize, bb=0.000000 0.000000 460.800000 345.600000]{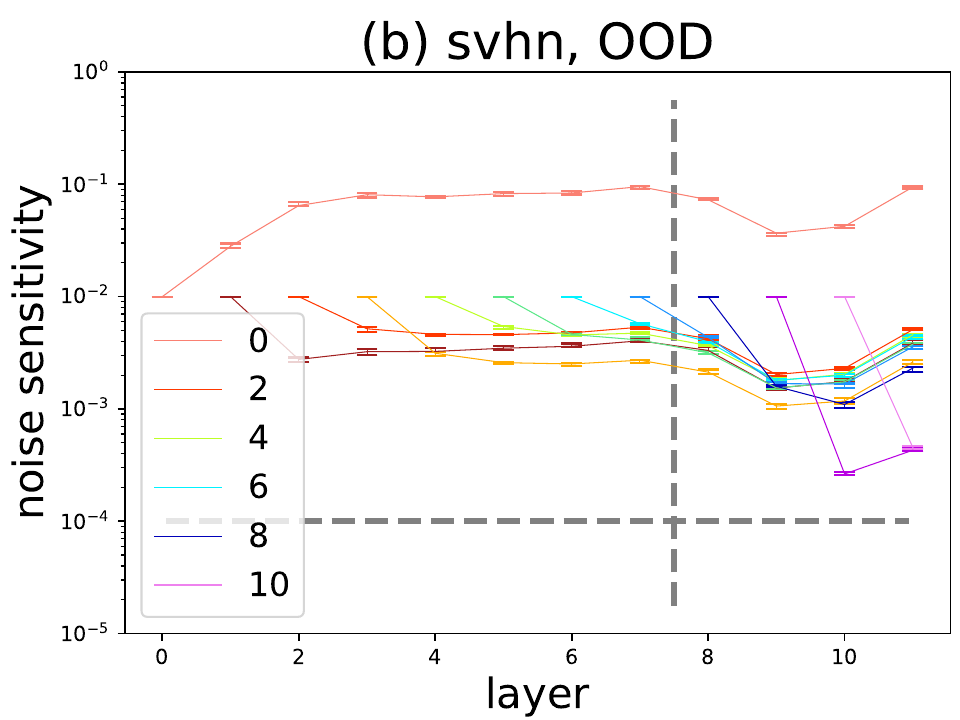}&
    \includegraphics[width=0.22\hsize, bb=0.000000 0.000000 460.800000 345.600000]{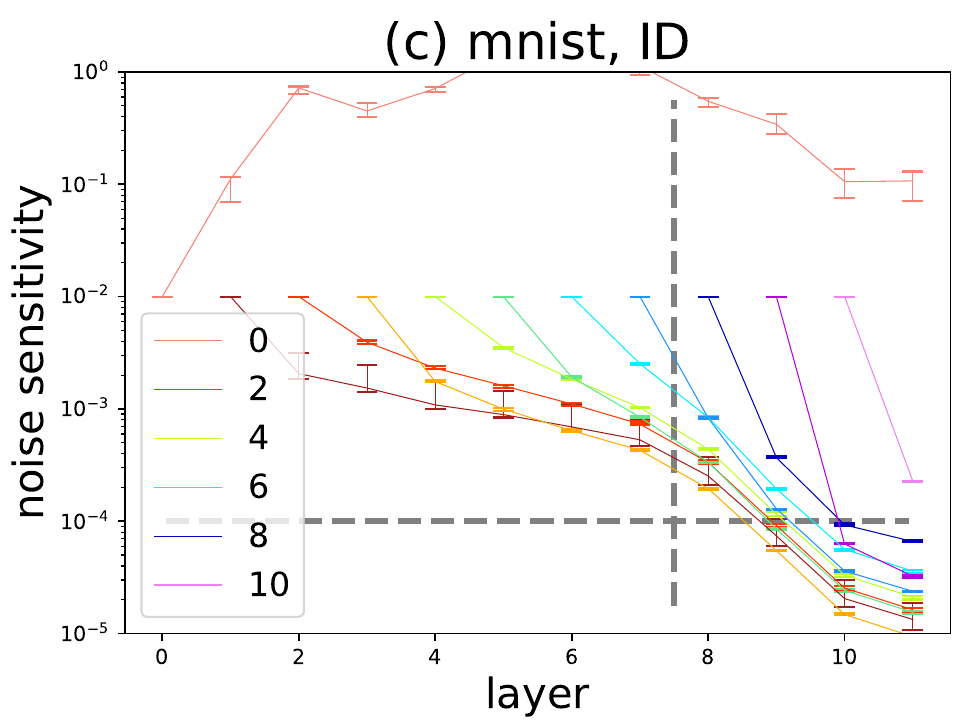}&
    \includegraphics[width=0.22\hsize, bb=0.000000 0.000000 460.800000 345.600000]{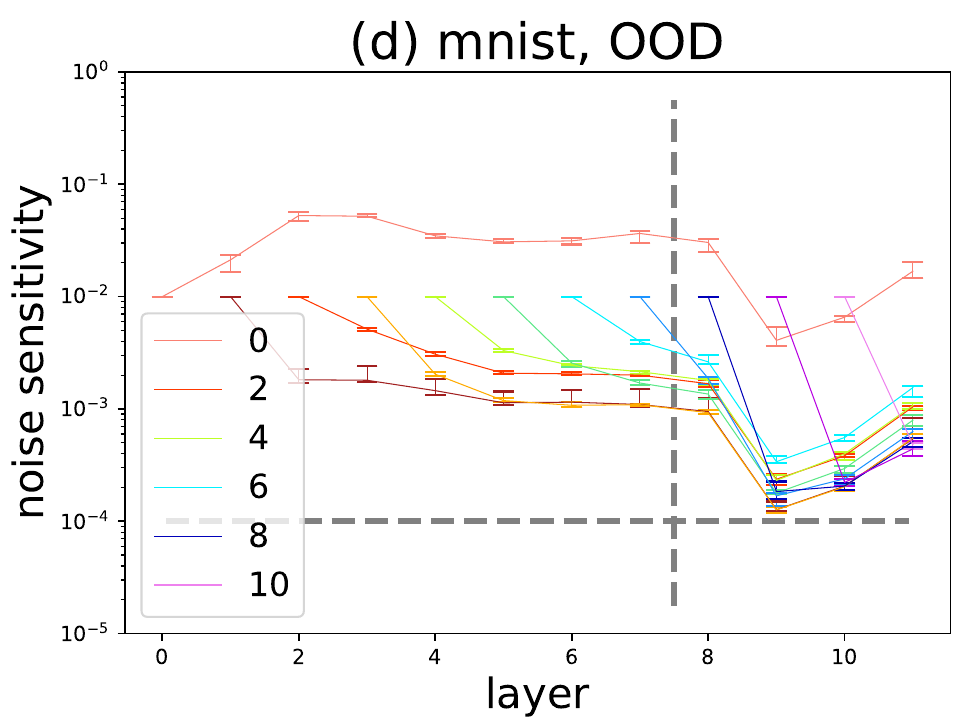}
  \end{tabular}
  \caption{
    Noise sensitivity for the VGG-13 model with (a,b) SVHN and (c,d) MNIST taken to be the ID dataset.
    The (a,c) left figures correspond to the noise sensitivity for ID the dataset, while the (b,d) right figures correspond to that for the OOD (CIFAR-10) dataset.
    In each figure, the horizontal axis is the layer
    and the vertical axis is the corresponding noise sensitivity.
    The different colors represent the different input layers where noise is injected.
  }
  \label{fig:app_sensitivity_svhn-mnist}
\end{figure}

Low dimensionalization can be observed by the stable rank in Fig. \ref{fig:app_stableranks_svhn-mnist} in the same manner as the main text, suggesting the importance of dimensionality.

Consistent behaviors are also visible in CKA and the noise sensitivity.
From CKA in Fig. \ref{fig:app_cka-penul_svhn-mnist}, we can observe the significant decay around the transition layer as in Fig. \ref{fig:app_cka-penul}.
Similarly, the difference of the sensitivity to the noise injection can be observed in Fig. \ref{fig:app_sensitivity_svhn-mnist} as in Fig. \ref{fig:sensitivity} in the main text.

\begin{figure}[t]
\centering
  \begin{tabular}{cccc}
    \includegraphics[width=0.22\hsize, bb=0.000000 0.000000 460.800000 345.600000]{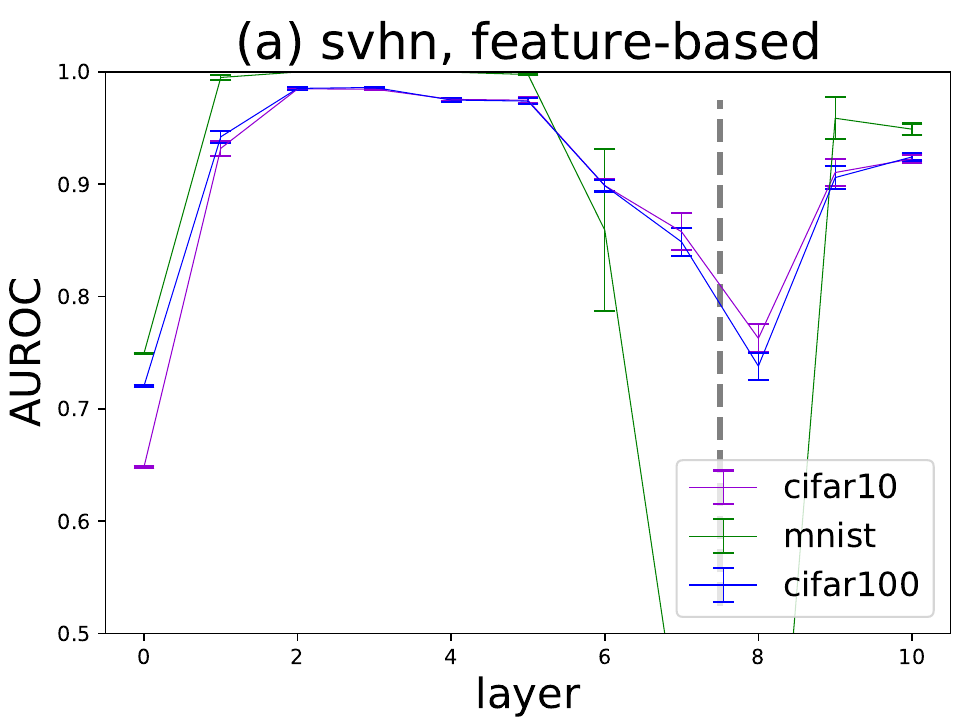}&
    \includegraphics[width=0.22\hsize, bb=0.000000 0.000000 460.800000 345.600000]{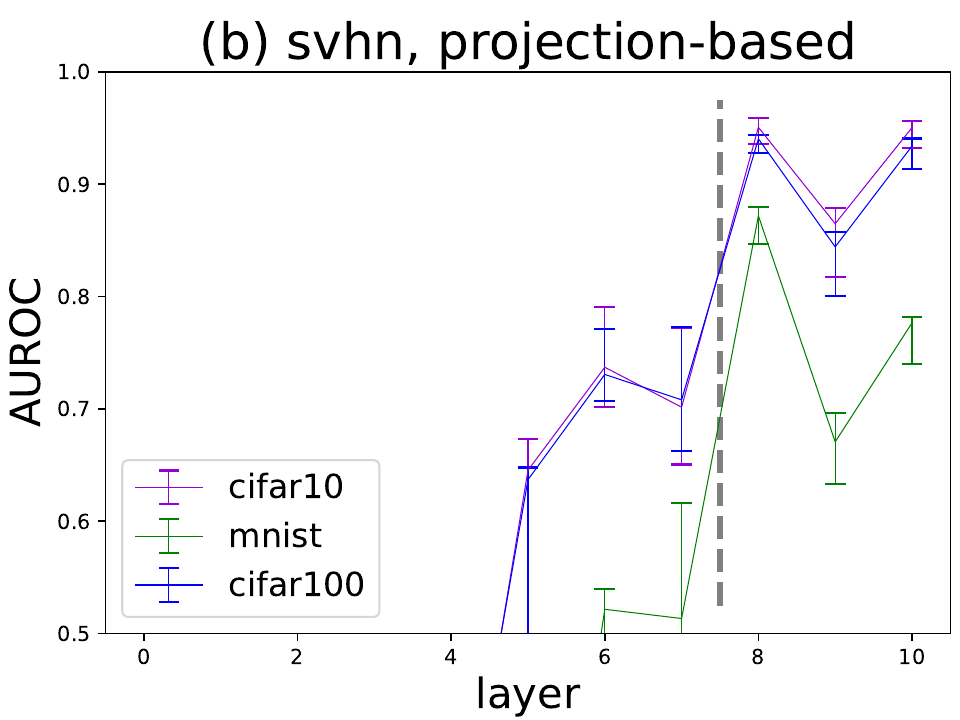}&
    \includegraphics[width=0.22\hsize, bb=0.000000 0.000000 460.800000 345.600000]{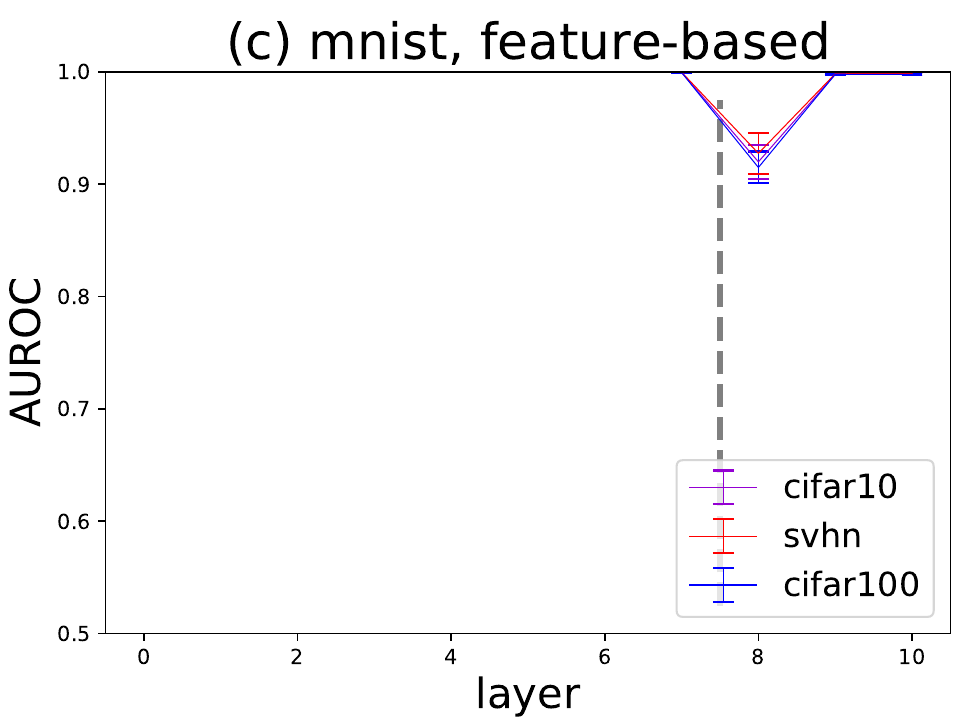}&
    \includegraphics[width=0.22\hsize, bb=0.000000 0.000000 460.800000 345.600000]{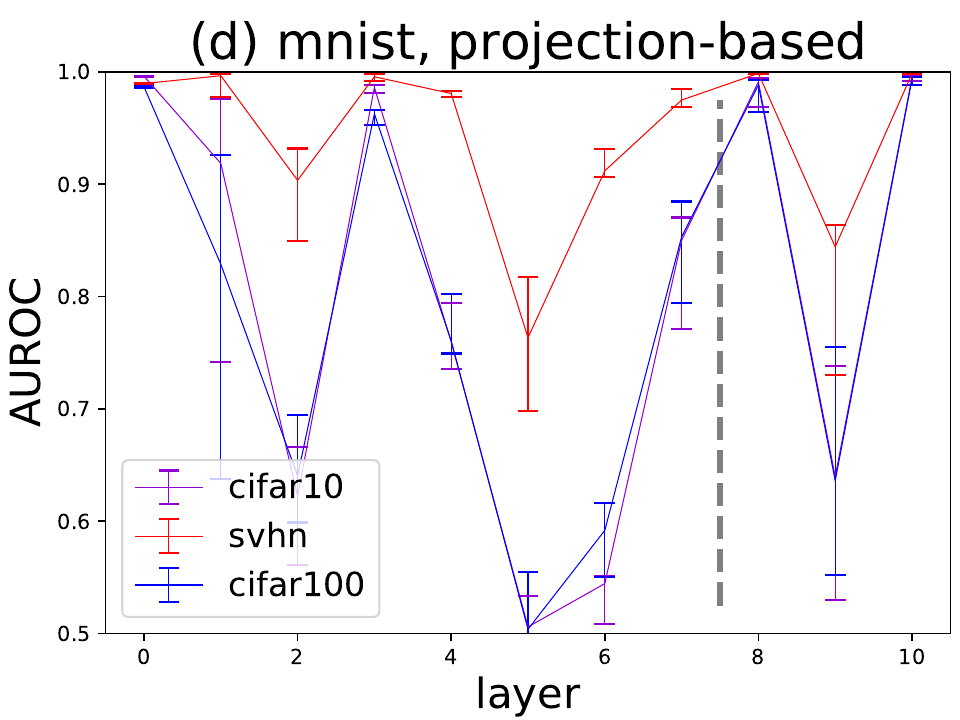}
  \end{tabular}
  \caption{
    Layer dependence of the AUROC 
    for the VGG-13 model with (a,b) SVHN and (c,d) MNIST taken to be the ID dataset.
    The (a,c) left figures are AUROCs of feature-based detection, while the (b,d) right figures are AUROCs of projection-based detection.
    The different line colors represent the different OOD datasets evaluated.
  }
  \label{fig:app_auroc_svhn-mnist}
\end{figure}

The layer dependence of the AUROC in Fig. \ref{fig:app_auroc_svhn-mnist} is also consistent with the main results, 
though visual impressions are slightly different.
That is, although the AUROC values are not the best, feature-based detection performance is stabilized after the transition layer, and projection-based detection performance exhibits high values just around the transition layer.
If we prepared corresponding close-to-ID OOD samples in these cases, the layer dependence of AUROCs would be more similar to the one in Fig. \ref{fig:auroc} in the main text.

\end{document}